\PassOptionsToPackage{unicode}{hyperref}
\PassOptionsToPackage{hyphens}{url}

\documentclass[
]{iopart}

\usepackage{amsmath,amssymb}
\usepackage{iftex}
\ifPDFTeX
  \usepackage[T1]{fontenc}
  \usepackage[utf8]{inputenc}
  \usepackage{textcomp} % provide euro and other symbols
\else % if luatex or xetex
  \usepackage{unicode-math}
  \defaultfontfeatures{Scale=MatchLowercase}
  \defaultfontfeatures[\rmfamily]{Ligatures=TeX,Scale=1}
\fi
\usepackage{lmodern}
\ifPDFTeX\else  
\fi
\IfFileExists{upquote.sty}{\usepackage{upquote}}{}
\IfFileExists{microtype.sty}{% use microtype if available
  \usepackage[]{microtype}
  \UseMicrotypeSet[protrusion]{basicmath} % disable protrusion for tt fonts
}{}
\makeatletter
\@ifundefined{KOMAClassName}{% if non-KOMA class
  \IfFileExists{parskip.sty}{%
    \usepackage{parskip}
  }{% else
    \setlength{\parindent}{0pt}
    \setlength{\parskip}{6pt plus 2pt minus 1pt}}
}{% if KOMA class
  \KOMAoptions{parskip=half}}
\makeatother
\usepackage{xcolor}
\ifx\paragraph\undefined\else
  \let\oldparagraph\paragraph
  \renewcommand{\paragraph}[1]{\oldparagraph{#1}\mbox{}}
\fi
\ifx\subparagraph\undefined\else
  \let\oldsubparagraph\subparagraph
  \renewcommand{\subparagraph}[1]{\oldsubparagraph{#1}\mbox{}}
\fi

\providecommand{\tightlist}{%
  \setlength{\itemsep}{0pt}\setlength{\parskip}{0pt}}\usepackage{longtable,booktabs,array}
\usepackage{calc} % for calculating minipage widths
\usepackage{etoolbox}
\makeatletter
\patchcmd\longtable{\par}{\if@noskipsec\mbox{}\fi\par}{}{}
\makeatother
\IfFileExists{footnotehyper.sty}{\usepackage{footnotehyper}}{\usepackage{footnote}}
\makesavenoteenv{longtable}
\usepackage{graphicx}
\makeatletter
\def\maxwidth{\ifdim\Gin@nat@width>\linewidth\linewidth\else\Gin@nat@width\fi}
\def\maxheight{\ifdim\Gin@nat@height>\textheight\textheight\else\Gin@nat@height\fi}
\makeatother
\setkeys{Gin}{width=\maxwidth,height=\maxheight,keepaspectratio}
\makeatletter
\def\fps@figure{htbp}
\makeatother

\usepackage{orcidlink}
\definecolor{mypink}{RGB}{219, 48, 122}
\usepackage{fvextra}
\DefineVerbatimEnvironment{Highlighting}{Verbatim}{breaklines,commandchars=\\\{\}}
\makeatletter
\@ifpackageloaded{caption}{}{\usepackage{caption}}
\AtBeginDocument{%
\ifdefined\contentsname
  \renewcommand*\contentsname{Table of contents}
\else
  \newcommand\contentsname{Table of contents}
\fi
\ifdefined\listfigurename
  \renewcommand*\listfigurename{List of Figures}
\else
  \newcommand\listfigurename{List of Figures}
\fi
\ifdefined\listtablename
  \renewcommand*\listtablename{List of Tables}
\else
  \newcommand\listtablename{List of Tables}
\fi
\ifdefined\figurename
  \renewcommand*\figurename{Figure}
\else
  \newcommand\figurename{Figure}
\fi
\ifdefined\tablename
  \renewcommand*\tablename{Table}
\else
  \newcommand\tablename{Table}
\fi
}
\@ifpackageloaded{float}{}{\usepackage{float}}
\floatstyle{ruled}
\@ifundefined{c@chapter}{\newfloat{codelisting}{h}{lop}}{\newfloat{codelisting}{h}{lop}[chapter]}
\floatname{codelisting}{Listing}

\usepackage{amsthm}
\theoremstyle{plain}
\newtheorem{lemma}{Lemma}[section]
\theoremstyle{remark}
\AtBeginDocument{}

\makeatother
\makeatletter
\@ifpackageloaded{caption}{}{\usepackage{caption}}
\@ifpackageloaded{subcaption}{}{\usepackage{subcaption}}
\makeatother
\ifLuaTeX
  \usepackage{selnolig}  % disable illegal ligatures
\fi
\usepackage{bookmark}

\IfFileExists{xurl.sty}{\usepackage{xurl}}{} % add URL line breaks if available
\hypersetup{
  pdftitle={ASPIRE: Iterative Amortized Posterior Inference for Bayesian Inverse Problems},
  pdfauthor={Rafael Orozco; Ali Siahkoohi; Mathias Louboutin; Felix J. Herrmann},
  hidelinks,
  pdfcreator={LaTeX via pandoc}}

\usepackage[numbers,square,sort&compress]{natbib}

\begin{document}

\title[ASPIRE]{ASPIRE: Iterative Amortized Posterior Inference for
Bayesian Inverse Problems}

\author{Rafael Orozco$^1$\orcidlink{0000-0003-0917-2442}, Ali
Siahkoohi$^2$\orcidlink{0000-0001-8779-2247}, Mathias
Louboutin$^3$\orcidlink{0000-0002-1255-2107} and Felix J.
Herrmann$^1$\orcidlink{0000-0003-1180-2167}}

    \address{$^1$ Computational Sciences and Engineering, Georgia
Institute of Technology, , Atlanta, USA }
    \address{$^2$ Department of Computational Applied Mathematics and
Operations Research, Rice University, , Houston, USA }
    \address{$^3$ Senior Solution Architect, DevitoCodes
Ltd, , London, UK }

\ead{rorozco@gatech.edu}

\begin{abstract}
Due to their uncertainty quantification, Bayesian solutions to inverse
problems are the framework of choice in applications that are risk
averse. These benefits come at the cost of computations that are in
general, intractable. New advances in machine learning and variational
inference (VI) have lowered the computational barrier by learning from
examples. Two VI paradigms have emerged that represent different
tradeoffs: amortized and non-amortized. Amortized VI can produce fast
results but due to generalizing to many observed datasets it produces
suboptimal inference results. Non-amortized VI is slower at inference
but finds better posterior approximations since it is specialized
towards a single observed dataset. Current amortized VI techniques run
into a sub-optimality wall that can not be improved without more
expressive neural networks or extra training data. We present a solution
that enables iterative improvement of amortized posteriors that uses the
same networks architectures and training data. The benefits of our
method requires extra computations but these remain frugal since they
are based on physics-hybrid methods and summary statistics. Importantly,
these computations remain mostly offline thus our method maintains cheap
and reusable online evaluation while bridging the approximation gap
these two paradigms. We denote our proposed method \textbf{ASPIRE} -
\textbf{A}mortized posteriors with \textbf{S}ummaries that are
\textbf{P}hysics-based and \textbf{I}teratively \textbf{RE}fined. We
first validate our method on a stylized problem with a known posterior
then demonstrate its practical use on a high-dimensional and nonlinear
transcranial medical imaging problem with ultrasound. Compared with the
baseline and previous methods from the literature our method stands out
as an computationally efficient and high-fidelity method for posterior
inference.
\end{abstract}

\section{Introduction}\label{introduction}

\begin{figure}

\centering{

\includegraphics{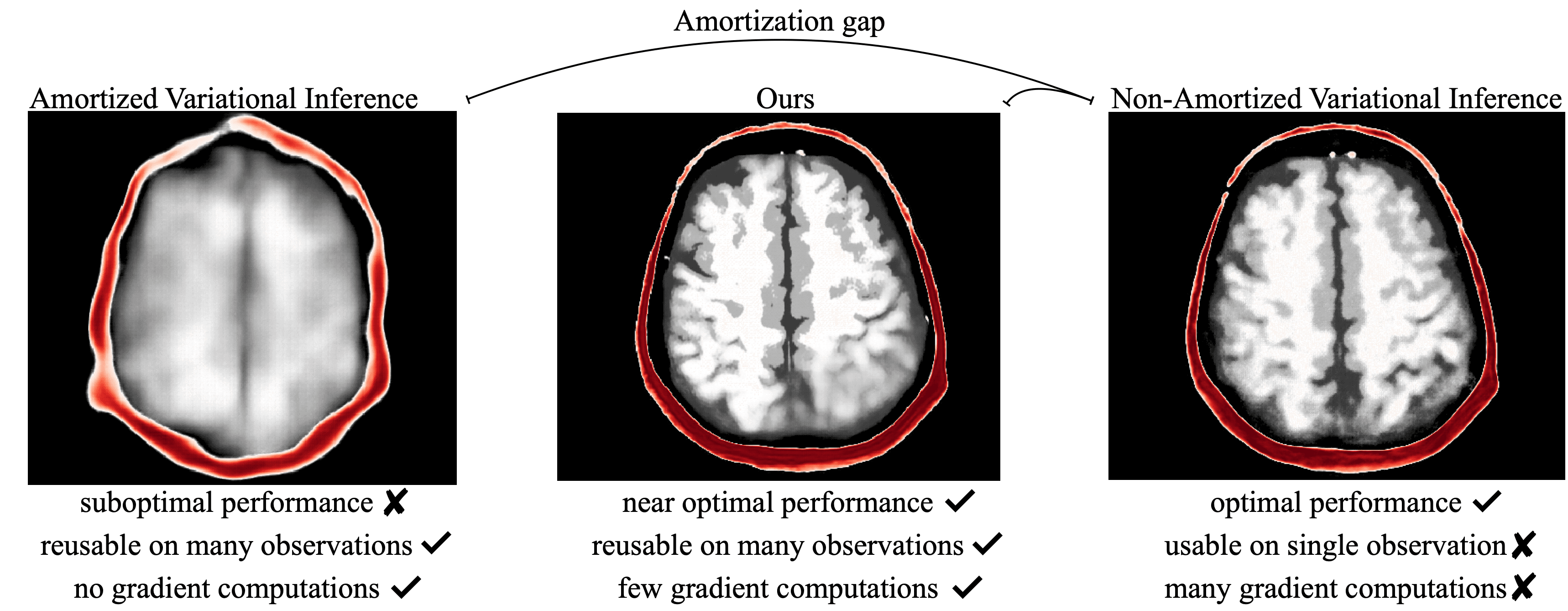}

}

\caption{\label{fig-main-framework}Our algorithm ASPIRE is a middle
ground between amortized and non-amortized variational inference.}

\end{figure}

Inverse problems are fundamental to numerous scientific and engineering
fields, wherein one seeks to infer causative factors from observable
effects. The central challenge in inverse problems is that they are
often ill-posed; the solutions may not exist, be non-unique, or depend
discontinuously on the data. This ill-posedness demands sophisticated
mathematical techniques to ensure stable solutions and to explore the
family of solutions that can explain the data. The example we treat
practically involves challenging medical imaging, where internal
structures of the human body (the brain) are inferred from ultrasound
measurements through the skull seen in Figure~\ref{fig-main-framework}.
The importance of fast and reliable solutions in this application cannot
be overstated as they directly influence possibly life saving diagnostic
decisions.

\subsection{Bayesian inverse problems}\label{bayesian-inverse-problems}

In this work, we tackle inverse problems that are solutions of the
forward problem: \begin{equation}\phantomsection\label{eq-forward}{
\mathbf{y} = \mathcal{G}(\mathcal{F}(\mathbf{x}), \boldsymbol{\varepsilon}) \, \, \, \,\,\,\,\,\, \boldsymbol {\varepsilon}\sim p(\boldsymbol{\varepsilon}).
}\end{equation}

The goal of this inverse problem is to form an image of unknown
parameter \(\mathbf{x}\) by indirectly observing them through a forward
operator \(\mathcal{F}\) and noise operator \(\mathcal{G}\),
parameterized by the noise instance \(\boldsymbol{\varepsilon}\).

In the case of noisy observations and ill-posed forward operators
\citep{hadamard1902problemes}, a single deterministic solution to the
inverse problem fails to characterize the full space of possible
solutions. Bayesian inverse problem solutions
\citep{tarantola2005inverse}, on the other hand, offer a more complete
characterization of the solution space by adhering to a probabilistic
framework. Here the goal is to find a statistical distribution for the
parameters that explains the data. The grail is to sample from the
conditional distribution \(p(\mathbf{x}|\mathbf{y})\), the so-called
posterior distribution. This distribution is given by Bayes' rule

\[
 p(\mathbf{x}\mid\mathbf{y}) \propto p(\mathbf{y}\mid\mathbf{x}) p(\mathbf{x}). 
\]

In words, Bayes' rule states that a Bayesian solution is formed by
updating our prior beliefs of the unknown parameter (prior
\(p(\mathbf{x}))\) with new information given by the observation
\(\mathbf{y}\), expressed by the data likelihood,
\(p(\mathbf{y}\mid\mathbf{x})\). This likelihood,
\(p(\mathbf{y}\mid\mathbf{x})\), encodes our domain knowledge in the
form of the forward operator \(\mathcal{F}\) and the noise process
\(\mathcal{G}\), \(p(\varepsilon)\). Thus, posterior samples are the
``parameters that are likely under the prior and also likely under the
data likelihood---i.e.~they explain the observed data''.

Exact posterior inference---i.e., calculating samples from the posterior
distribution or its statistics (mean, (co)variance or higher order
moments), is in general computationally intractable
\citep{gelman1995bayesian}. The intractable nature of posterior sampling
arises from: the curse of dimensionality when dealing with
high-dimensional parameters, the expense of forward operator evaluation
related to the data likelihood, and multimodality of the distribution,
etc. \citep{curtis2001prior}. For specific cases, such as linear forward
operators and Gaussian or conjugate priors, the posterior distribution
has an analytical form. For example, linear operators, Gaussian noise,
and Gaussian priors lead to Gaussian posteriors with known means and
covariances \citep{bishop2006pattern}. But in many real-world
applications, the forward operator is expensive and/or nonlinear and
there does not exist a known prior. In these cases, more advanced
methods are required for posterior inference. These advanced methods can
be divided into two types: the first type of methods are sample based.
These include Markov-chain Monte Carlo (McMC) and its various
counterparts
\citep{robert1999monte, siahkoohi2022deep, martin2012stochastic}. On the
other hand, there are optimization based methods, such as expectation
maximization \citep{dempster1977maximum}, the Laplace approximation
\citep{tierney1986accurate}, and variational inference (VI)
\citep{jordan1999introduction}. Here, we consider VI because it can
naturally exploit the ability of deep neural networks to learn high
dimensional distributions.

\subsection{Posterior sampling with variational
inference}\label{posterior-sampling-with-variational-inference}

The technique of VI optimizes an approximate distribution,
\(p_{\theta}(\mathbf{x}\mid \mathbf{y}),\,\theta \in \Theta\). The
parameters of these distributions are chosen to match the unknown target
distribution \(p(\mathbf{x}\mid \mathbf{y})\). Due to its connection
with maximum likelihood methods \citep{bishop2006pattern}, and its
relatively easy to optimize objective, the mismatch between the
approximate and target distribution is typically measured by the
Kullback-Leibler (KL) divergence \citep{kullback1951information}.
Because this divergence metric is non-symmetric, it allows for two
complementary VI formulations, namely non-amortized VI, which uses the
backward KL divergence
\(\mathbb{KL}( p_{\theta}(\mathbf{x}\mid \mathbf{y}) \mid \mid p(\mathbf{x}\mid \mathbf{y}) )\)
and amortized VI, which involves the forward KL divergence
\(\mathbb{KL}( p(\mathbf{x}\mid \mathbf{y}) \mid \mid p_{\theta}(\mathbf{x}\mid \mathbf{y}) )\)
\citep{siahkoohi2023reliable}. These two formulations have different
requirements, costs, and benefits, which we will discuss. Firstly, we
will describe the most commonly implemented form: non-amortized VI.

\subsection{Non-amortized variational
inference}\label{non-amortized-variational-inference}

Because the backward KL divergence entails evaluation of the
\(\log\)-likelihood conditioned on a single observation,
\(\mathbf{y}^{\mathrm{obs}}\), its minimization requires knowledge of
the forward operator \(\mathcal{F}\) and its gradient. The inference is
non-amortized since it is carried out with respect to a single
observation. To understand these statements, let us consider the case
where the noise is Gaussian with standard deviation \(\sigma\) for which
the \(\log\)-likelihood can be written out explicitly, yielding
\begin{align*}
   \underset{\mathbf{\theta} }{\operatorname{minimize}} & \, \mathbb{KL}\bigl(  p_{\theta}(\mathbf{x}\mid \mathbf{y}^{\mathrm{obs}} )  \mid \mid  p(\mathbf{x}\mid\mathbf{y}^{\mathrm{obs}} )\bigr)  \\
& =   \mathbb{E}_{ p_{\theta}(\mathbf{x}\mid \mathbf{y}^{\mathrm{obs}} )} \Bigl[-\log p(\mathbf{x}\mid \mathbf{y}^{\mathrm{obs}} )  +  p_{\theta}(\mathbf{x}\mid\mathbf{y}^{\mathrm{obs}} )  \Bigr]   \\ 
& =   \mathbb{E}_{ p_{\theta}(\mathbf{x}\mid \mathbf{y}^{\mathrm{obs}} )} \Bigl[ \frac{1}{2 \sigma^2} \lVert \mathcal{F}(\mathbf{x}) - \mathbf{y}^{\mathrm{obs}} \rVert_2^2   - \log p(\mathbf{x}) +  p_{\theta}(\mathbf{x}\mid \mathbf{y}^{\mathrm{obs}} )  \Bigr].    
\end{align*} From these expressions, we first note that the optimization
is indeed performed for a single observation,
\(\mathbf{y}^{\mathrm{obs}}\). This implies that when inference results
are desired for a different observation, the optimization must be
repeated, which may be an expensive proposition in situations where
\(\mathcal{F}\) and its gradient are expensive to evaluate. For
instance, when \(\mathcal{F}\) and its gradient require the solution of
a partial differential equations (PDE) over a high-dimensional parameter
space, their repeated evaluation as part of gradient descent often
becomes the most expensive computation when minimizing the backward KL
divergence. Finally, minimization of the backward KL divergence also
requires evaluation of the prior, \(p(\mathbf{x})\), and its gradient.
This imposes a difficulty because, in many cases, this prior is not
known analytically and must be approximated
\citep{siahkoohi2021preconditioned, feng2023score}.

There are a variety of implementations of non-amortized posterior
inference, including those based on Langevin dynamics
\citep{welling2011bayesian, siahkoohi2022deep} and those that make use
of normalizing flows \citep{sun2021deep}. Other examples include methods
based on the Stein discrepancy \citep{liu2016stein, zhang20233} and
randomize-then-optimization methods
\citep{blatter2022uncertainty, bardsley2014randomize}. While these
non-amortized inference techniques have shown promise, their online
application can be rendered ineffective when applications call for a
rapid time-to-solution as may be the case in medical imaging. We will
address this situation by presenting an inference technique where most
of the computational costs are incurred off-line, so the inference is
fast for different observations.

\subsection{Amortized variational
inference}\label{amortized-variational-inference}

Guiding distributional optimization with the forward KL divergence as a
mismatch metric between the target distribution and the approximate
distribution leads to a formulation called amortized VI. To arrive at
this formulation, let us first write out the expression for the forward
KL divergence:

\[
   \underset{\mathbf{\theta} }{\operatorname{minimize}}  \, \mathbb{KL} \bigl( \,  p(\mathbf{x}\mid \mathbf{y}) \, \,||\,\, p_{\theta}(\mathbf{x}\mid \mathbf{y}) \bigr) \nonumber  
 =  \mathbb{E}_{p(\mathbf{x}\mid \mathbf{y})} \Bigl[-\log p_{\theta}(\mathbf{x}\mid \mathbf{y})  +  p(\mathbf{x}\mid \mathbf{y})  \Bigr] \nonumber 
\]

and quickly note that evaluating this expression depends on having
access to samples from the ground-truth posterior distribution
\(p(\mathbf{x}\mid \mathbf{y})\). As these posterior samples are
typically not available, we marginalize over the distribution of
observations instead---i.e., we have

\begin{align}
   \underset{\mathbf{\theta} }{\operatorname{minimize}} 
 \,\mathbb{E}_{p(\mathbf{y})} \Bigl[ \mathbb{KL} \bigl( \,  p(\mathbf{x}\mid \mathbf{y}) \, \,||\,\, p_{\theta}(\mathbf{x}\mid \mathbf{y}) \bigr) \Bigr]  \nonumber  &= \mathbb{E}_{p(\mathbf{y})} \Biggl[ \mathbb{E}_{p(\mathbf{x}\mid \mathbf{y})} \Bigl[-\log p_{\theta}(\mathbf{x}\mid \mathbf{y})  +  p(\mathbf{x}\mid \mathbf{y})  \Bigr]\Biggr] \nonumber \\ 
 & = \mathbb{E}_{p(\mathbf{x}, \mathbf{y})} \Bigl[-\log p_{\theta}(\mathbf{x}\mid \mathbf{y})  +  p(\mathbf{x}\mid \mathbf{y})  \Bigr] \nonumber \\
  & = \mathbb{E}_{p(\mathbf{x}, \mathbf{y})} \Bigl[-\log p_{\theta}(\mathbf{x}\mid \mathbf{y})   \Bigr].   \label{eq-amort-objective-joint}   
\end{align}

To arrive at the final expression, we made use of the law of total
probability and the fact that the optimization is only over parameters
\(\theta\). See also \citep{siahkoohi2023reliable}. From this final
expression , the requirements of training amortized VI become clear: we
need samples of the joint distribution
\(\mathbf{x},\mathbf{y} \sim p(\mathbf{x},\mathbf{y})\) and a parametric
conditional density estimator. In this work, we obtain samples of the
joint distribution using a simulation-based inference framework
\citep{cremer2018inference}, and train a generative neural network as a
conditional density estimator \citep{radev2020bayesflow} for the
posterior.

Amortized VI is so-called ``amortized'' because its strength lies in its
reusability. Once optimized during off-line training, the approximation
is not exclusive to a single observation but rather can be applied
across numerous observations. This means that the computational expenses
involved in the initial optimization phase are effectively ``spread
out'' over multiple inference tasks, making the inference for any new
observation significantly cheaper. As we will see in the methods
section, this ``spread out'' practically refers to learning inference
tasks over a set of training examples. By learning from examples, the
method can remember important features that can be reused at inference
time for many unseen observations. Due to this phenomenon, this
formulation is sometimes called \emph{inference with memory} while
non-amortized inference is called \emph{memory-less inference}
\citep{putzky2023amortized}. Table~\ref{tbl-compare} summarizes the main
requirements and benefits of amortized approaches compared to
non-amortized ones.

\begin{longtable}[]{@{}
  >{\raggedright\arraybackslash}p{(\columnwidth - 4\tabcolsep) * \real{0.3247}}
  >{\raggedright\arraybackslash}p{(\columnwidth - 4\tabcolsep) * \real{0.2857}}
  >{\raggedright\arraybackslash}p{(\columnwidth - 4\tabcolsep) * \real{0.3896}}@{}}
\caption{Comparison requirements and benefits of Amortized posterior
versus Non-amortized posterior
inference}\label{tbl-compare}\tabularnewline
\toprule\noalign{}
\begin{minipage}[b]{\linewidth}\raggedright
\end{minipage} & \begin{minipage}[b]{\linewidth}\raggedright
Amortized posterior inference
\end{minipage} & \begin{minipage}[b]{\linewidth}\raggedright
Non-amortized posterior inference
\end{minipage} \\
\midrule\noalign{}
\endfirsthead
\toprule\noalign{}
\begin{minipage}[b]{\linewidth}\raggedright
\end{minipage} & \begin{minipage}[b]{\linewidth}\raggedright
Amortized posterior inference
\end{minipage} & \begin{minipage}[b]{\linewidth}\raggedright
Non-amortized posterior inference
\end{minipage} \\
\midrule\noalign{}
\endhead
\bottomrule\noalign{}
\endlastfoot
Reusable on many observations & Yes & No \\
Forward operator & Only evaluations & Evaluations and gradients \\
Needs prior & Only samples & Density calculations \\
\end{longtable}

Because non-amortized inference focuses on one single observation, its
inference typically outperforms amortized inference
\citep{siahkoohi2023reliable}. Unfortunately, this improvement often
comes at the expense of prohibitively high computational costs at
inference time, rendering non-amortized inference impractical in
situations where fast turn-around times are needed, such as in many
medical imaging fields \citep{bauer2013real}. Amortized VI methods, on
the other hand, while fast \citep{cremer2018inference} at inference may
suffer from the so-called amortization gap, a phenomenon that has been
studied theoretically \citep{marino2018iterative} and confirmed
empirically from comparisons between amortized and non-amortized VI
\citep{siahkoohi2023reliable, whang2021composing, orozco2021photoacoustic}.
In that sense, there is a trade-off between runtime and quality at
inference time. One either spends more on computations at inference
time, or one accepts inferior inference quality in situations where fast
turn-around times are essential. While trading quality for speed may be
acceptable in some situations, it becomes problematic in circumstances
where amortized VI produces unacceptable results, e.g., in cases where
the inference problem is high dimensional and complicated by nonlinear
forward operators. Currently, the following remedies exist: \emph{(1)}
increase the expressiveness of the parametric family used to approximate
the posterior \citep{grcic2021densely} or \emph{(2)} add more samples to
the training set \citep{shorten2019survey}. In this work, we will
explore a third complementary option to narrow the amortization gap. To
this end, we propose an iterative amortized inference approach during
which physics-based summary statistics are refined in tandem with neural
posterior estimators thus bootstrapping the quality of the approximated
posterior. We call this approach: \textbf{ASPIRE} - \textbf{A}mortized
posteriors with \textbf{S}ummaries that are \textbf{P}hysics-based and
\textbf{I}teratively \textbf{RE}fined. To motivate this approach, we
will first explore the implementation of amortized posterior inference
via neural density estimation, followed by physics-based gradient
summary statistics, and their iterative refinement.

\section{Contributions}\label{contributions}

\begin{enumerate}
\def\labelenumi{\arabic{enumi}.}
\tightlist
\item
  Motivated by score-based, maximally informative summary statistics we
  introduce the ASPIRE algorithm, which iteratively refines amortized
  posterior inference while maintaining low online costs.
\item
  Evaluation of our method on a stylized problem to ensure that the
  approximate posterior accurately samples the ground-truth posterior.
\item
  Evaluation of our method's performance on a realistic and challenging
  transcranial medical imaging inverse problem with ultrasound, focusing
  on the accuracy of the posterior mean and the effectiveness of our
  uncertainty in predicting reconstruction errors.
\item
  Introduction of a novel non-amortized inference method to serve as a
  ``gold standard''.
\item
  Qualitative and quantitative comparisons against a current literature
  baseline and our ``gold standard'' non-amortized inference method. To
  accelerate the development of inference techniques for transcranial
  medical imaging \citep{donoho2023data}, we introduce benchmarks with
  accompanying datasets and code.
\item
  Cost-benefit analysis of the computational costs associated with
  offline training versus the rapid online capabilities of the amortized
  method.
\end{enumerate}

\section{Method}\label{method}

To close the amortization gap, we describe an iterative approach to
posterior inference where learned physics-based summary statistics are
refined with Conditional Normalizing Flows (CNFs). VI with CNFs is
reviewed first. Its improvement with learned physics-based summary
statistics is discussed next, including addition of the crucial
refinement step.

\subsection{Amortized variational inference with conditional normalizing
flows}\label{amortized-variational-inference-with-conditional-normalizing-flows}

While our method can, in principle, be applied to any generative model
(GAN, diffusion, VAE), we focus on normalizing flows
\citep{dinh2016density}. Thanks to their simple maximum-likelihood
training objective, low training memory requirements
\citep{orozco2023invertiblenetworks}, and fast sampling, CNFs have
become one of the generative methods of choice when inverse problems are
concerned. CNFs learn to sample from a target density by learning
invertible transformations from the data distribution to the standard
Normal distribution. By taking advantage of the change of variables
formula \citep{kovachki2020conditional}, CNFs can be trained with a
relatively simple objective:
\begin{equation}\phantomsection\label{eq-train-cond}{
\widehat{\boldsymbol{\theta}} = \underset{\boldsymbol{\theta}}{\operatorname{arg min}}
 \,  \frac{1}{N} \sum_{n=0}^{N}  \left( \frac{1}{2} \lVert  f_{\boldsymbol{\theta}}(\mathbf{x}^{(n)};\mathbf{y}^{(n)}) \rVert_{2}^2 - \log \bigr|  \det{  \mathbf{J}_{f_{\boldsymbol{\theta}}}} \bigr| \right),
}\end{equation}

where the neural network's Jacobian,
\(\mathbf{J}_{f_{\boldsymbol{\theta}}}\), is calculated with respect to
the target variable \(\mathbf{x}\). Please refer to
\citep{siahkoohi2023reliable}, for a derivation on the above expression
from the generic amortized posterior objective in Equation
\ref{eq-amort-objective-joint}. The training pairs, collected in
\(\mathcal{D}=\{(\mathbf{x}^{(n)},\, \mathbf{y}^{(n)})\}_{n=0}^N\), are
generated by sampling from the prior,
\(\{\mathbf{x}^{(n)}\}_{n=0}^N \sim p(\mathbf{x})\), followed by a
forward simulation applying Equation~\ref{eq-forward}. After the
optimization is completed, the CNF with optimized weights,
\(\widehat{\boldsymbol{\theta}}\), can be used to sample from the
approximate posterior
\(p_{\widehat{\boldsymbol{\theta}}}(\mathbf{x}\mid\mathbf{y}=\mathbf{y}^{\mathrm{obs}})\)
by sampling Gaussian noise and passing it through the inverse network
that is conditioned on an observation \(\mathbf{y}^{\mathrm{obs}}\).

\begin{equation}\phantomsection\label{eq-sampling}{
 \mathbf{x}=f^{-1}_{\widehat{\boldsymbol{\theta}}}(\mathbf{z};\mathbf{ y}^{\mathrm{obs}}) \,\, \text{where}\,\,\mathbf{z}\sim \mathcal{N}(0,\,I).
}\end{equation}

Contrary to non-amortized VI, the above posterior sample generation
holds for any observation, \(\mathbf{y}^{\mathrm{obs}}\), as long as the
observations remain close---i.e., the \(\mathbf{y}^{\mathrm{obs}}\) are
produced by applying Equation~\ref{eq-forward} to samples of the prior,
\(\mathbf{x}\sim p(\mathbf{x})\). However, the quality of the posterior
approximation, and therefore the quality of its samples---
\(\mathbf{x}\sim p_{\widehat{\boldsymbol{\theta}}}(\mathbf{x}\mid\mathbf{y}=\mathbf{y}^{\mathrm{obs}})\),
depends on the complexity of the posterior that is being approximated.
This in turn depends on the complexity of prior samples and the
likelihood. To account for realistic situations where both the prior and
likelihood are complicated, CNFs demand increases in the size of the
training set and the expressive power of the density estimator
\(f_{\theta}(\cdot)\), a requirement we like to avoid. For this reason,
we will introduce the concept of \emph{summary statistics} that allows
us to improve the quality of the posterior approximation in these
situations.

\subsection{Score-based summary
statistics}\label{score-based-summary-statistics}

While CNFs are in principle capable of capturing complex data-to-image
space mappings, amortization can be challenging to achieve in situations
where the mapping is complex, or the observed data are
heterogeneous---i.e., the observed data differ in dimension and/or
source-receiver acquisition geometry. To overcome these challenges,
statisticians introduced so-called \emph{summary statistic}. These often
hand-derived summary statistics are designed to capture the main
features in the data, reduce and homogenize its dimensionality, while
posterior distributions remain informed \citep{radev2020bayesflow}. For
compressed posterior distributions to remain informed, it is essential
in all approaches that the conditioning by the summary statistic,
denoted by \(\mathbf{\overline y}\), minimally changes the original
conditional distribution---i.e., we have \[
p(\mathbf{x}|\mathbf{\overline y}) \approx p(\mathbf{x}|\mathbf{y}). 
\]

When this approximate equality holds exactly, the summary statistic is
known to be a sufficient summary statistic. Since equality can not
always be met, there also exists the notion of being close to
sufficient. To be more specific, a summary statistic remains maximally
informative \citep{deans2002maximally} with respect to a set of summary
statistics \(\mathbf{y'} \in \mathbf{Y}\) when some distance measure
between the summarized and original posterior distributions is
minimized. For a distance measure given by the KL divergence, this
amounts to finding the minimum of the following objective: \[
 \mathbf{\overline y} = \underset{\mathbf{y'\in Y}}{\operatorname{arg min}}
 \,  \mathbb{KL} \left( \,  p(\mathbf{x}\mid \mathbf{y'}) \, \,||\,\, p(\mathbf{x}\mid \mathbf{y}) \right). \nonumber 
\]

Alternatively, one can also measure the informativeness of a certain
summary statistic by its Fischer information \citep{heavens2000massive},
which corresponds to the expected variance of the score function given
by the gradient of the natural logarithm of the likelihood. Based on
this argument, \citep{alsing2018generalized} proposed the gradient of
the \(\log\)-likelihood
\(\nabla_\mathbf{x} \log p(\mathbf{y}\mid\mathbf{x})\) as a maximally
informative summary statistic. It can be shown that under certain
assumptions the information inequality is saturated by the gradient of
the \(\log\)-likelihood, in other words that no other summary statistic
can have more Fischer information \citep{alsing2018generalized}. The
gradient of the \(\log\)-likelihood is an attractive summary statistic
because it (the score function) allows for the inclusion of knowledge on
the forward operator, \(\mathcal{F}\), and its Jacobian,
\(\nabla \mathcal{F}\). For instance, if the noise is Gaussian and the
noise operator \(\mathcal{G}\) is additive then the summary statistic is
given by the action of the adjoint of the Jacobian,
\(\nabla\mathcal{F}^\top[\mathbf{x}_0]\) evaluated at \(\mathbf{x}_0\),
on the residual. In that case, we can write
\begin{equation}\phantomsection\label{eq-score}{ 
  \mathbf{\overline y} = \nabla_\mathbf{x} \log p(\mathbf{y}\mid\mathbf{x}) \Bigr|_{\mathbf{x}_0} = \nabla \mathcal{F}^\top[\mathbf{x}_0](\mathcal{F}(\mathbf{x}_0)-\mathbf{y}). 
}\end{equation}

To train a CNF with this gradient summary statistic, the objective of
Equation~\ref{eq-train-cond} is minimized on a new training set obtained
by applying Equation~\ref{eq-score} to the observed datasets collected
in \(\mathcal{D}\), yielding
\(\mathcal{D}_0 = \{\mathbf{x}^{(n)}, \mathbf{\overline y}_0^{(n)}\}_{n=0}^{N}\).
The subscript \(0\) was added in our notation to indicate that each
single summary statistic, \(\mathbf{\overline y}_0^{(n)}\), derives from
the evaluation of the gradient at \(\mathbf{x}_0^{(n)}\), a point which
is henceforth referred to as a fiducial point. This fiducial point
represents a trusted guess of the unknown parameter.

The quality of gradient summary statistics is contingent on two key
factors, namely the quality of the assumed likelihood and the quality of
the fiducial points. The quality of the former depends on choices for
the noise distribution and the forward operator, \(\mathcal{F}\). When
misspecified, or poorly calibrated, these choices may affect the quality
of the summary statistic. We assume in this work that the forward
problem and noise model are well specified. The second factor that
determines the quality concerns choices for the fiducial points
themselves. Because the fiducial point is used to calculate the score
function then its choice correlates with the information content of the
resulting score. Thus, the choice for these fiducial points determines
the quality of the summary statistic, which in turn determines the
quality of the posterior inference itself. As shown by
\citep{alsing2018generalized}, fiducial points that stay close to the
maximum likelihood: \[
 \mathbf{x}_{ML} = \underset{\mathbf{x}}{\operatorname{argmax}} \,  p(\mathbf{y}\mid \mathbf{x})
\]

lead to score-based summary statistics that are maximally informative
with respect to the Fisher information. However, in the common situation
where the fiducial points are not close to the maximum likelihood this
property may potentially no longer hold, rendering score-based summary
statistics less informative. As in many practical situations, we
unfortunately do not always have access to high-quality fiducial points,
a situation we will remedy in the next section where data-driven
learning will be combined with score-based summary.

Before introducing a novel iterative approach to render fiducial points
more informative, note that the assumed likelihood in
Equation~\ref{eq-score} need not be the actual likelihood
\(p(\mathbf{y}\mid\mathbf{x})\). Indeed, as discussed by
\citep{alsing2018massive}, this summary statistic is still applicable in
methods that are likelihood-free since you only need an assumed
likelihood to be reasonably close to the true likelihood, or learned
likelihoods \citep{brehmer2020mining} derived from samples of the
likelihood.

\subsection{Refining summary
statistics}\label{refining-summary-statistics}

\begin{figure}

\centering{

\includegraphics{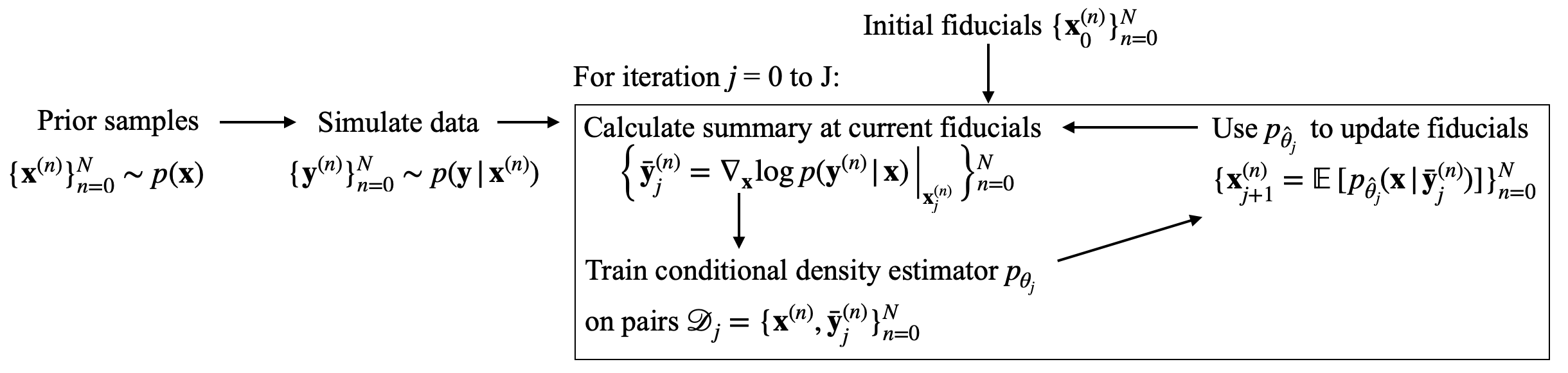}

}

\caption{\label{fig-algo-train}Offline training phase: During training,
our proposed algorithm refines the fiducial point used to calculate the
summary statistics. Importantly, this refinement is done in an amortized
manner over a training dataset.}

\end{figure}

Notwithstanding the fact that score-based summary statistics represent a
natural approach to inference problems that involve well-understood
physics, reliance on good fiducial points---i.e., points that are close
to their respective maximum likelihoods, remains problematic especially
in situations where inference problems are ill-posed or reliant on good
starting parameters to succeed. Instead of performing expensive, and
potentially local-minima prone \citep{van2013mitigating}, Gauss-Newton
updates to bring the fiducial points closer to the maximum likelihood
points, as suggested by \citep{alsing2018massive}, we propose an
iterative scheme during which CNFs are trained then sampled to estimate
improved fiducial points. The iterations, outlined in
Figure~\ref{fig-algo-train} , proceed as follows: given the current
iterate for the fiducial points, this would be
\(\{\mathbf{x}_0^{(n)}\}_{n=0}^{N}\) at the first iteration, score-based
summary statistics are computed with Equation~\ref{eq-score} and a CNF
is trained on \(\mathcal{D}_j\) (\(j=0\), for the first iteration) by
minimizing Equation~\ref{eq-train-cond}. This minimization produces
optimized CNF weights, \(\boldsymbol{\widehat{\theta}}_j\), which are
used to draw multiple samples from the posterior, via
Equation~\ref{eq-sampling}, for each simulated observation in
\(\{\mathbf{\overline{y}}^{(n)}_j\}_{n=0}^{N}\). Next, these posterior
samples are averaged to approximate the expectation,
\(\{\mathbf{x}^{(n)}_{j+1}= \mathbb{E}\bigl[p_{\widehat{\boldsymbol{\theta}}_j}(\mathbf{x}\mid \overline{\mathbf{y}}^{(n)}_j)\bigr]\}_{n=0}^{N}\),
computed for each score-based summary statistic, separately. This
averaging, which corresponds to approximating the posterior mean for
each \(\{\overline{\mathbf{y}}^{(n)}\}_{n=0}^{N}\), produces the next,
and arguably improved set of fiducial points,
\(\{\mathbf{x}^{(n)}_{j+1}\}_{n=0}^{N}\). This process is repeated \(J\)
times. As long as the new set of fiducials points moves closer to the
maximum likelihood points, the quality of the score-based summary
statistic can be expected to improve. These improvements in turn
produces better posterior inferences by the CNFs.

The above iterative refinement scheme hinges on the assumption that the
fiducial points improve---i.e., they are indeed closer to the maximum
likelihood points. We argue that improvements are to be expected for a
few reasons. First, the gradient of the \(\log\)-likelihood that defines
our summary statistic engenders the first iteration of gradient descent,
which under certain conditions is known to move initial fiducial points
closer to the maximum-likelihood points \citep{casella2002statistical}.
\citep{alsing2018massive} used the same argument by suggesting that a
limited number of Gauss-Newton updates improves the informativeness of
these refined fiducial points. Second, we propose to update the
fiducials with the average samples from the posterior for each
observation. This average, known as the posterior mean, is unbiased and
contains information on the prior. To add some rigor, we summarize our
claim with the following lemma:

\begin{lemma}[]\protect\hypertarget{lem-line}{}\label{lem-line}

If the score \(\mathbf{\overline y}_0\) is calculated at a fiducial
\(\mathbf{x}_0\) that is inside the basin of attraction of the maximum
likelihood \(\mathbf{x}_{ML}\) and the conditional density estimator
\(p_{\widehat{\boldsymbol{\theta}}_0}\) is trained on dataset
\(\mathcal{D}_0\) to convergence then \(\mathbf{x}_1\) defined by the
mean of the conditional density estimator samples will be closer to the
maximum-likelihood \(\mathbf{x}_{ML}\): \[
\|\mathbf{x}_1 - \mathbf{x}_{ML}  \|_2 < \| \mathbf{x}_0 - \mathbf{x}_{ML} \|_2 \,\,\,\, \text{where} \, \, \mathbf{x}_1 = \mathbb{E}_{\mathbf{x}}[ p_{\widehat{\boldsymbol{\theta}}_0}(\mathbf{x} \mid \mathbf{\overline y}_0)]. 
\]

\end{lemma}

Informally, Lemma~\ref{lem-line} derives from the observation that the
conditional mean of the posterior distribution is the estimator that
minimizes the mean-squared error with respect to the ground truth,
\(\mathbf{x}^{\ast}\) \citep{adler2018deep}. Since the
\(\mathbf{x}^{\ast}\)'s are also maximally likely, then CNF updates that
bring the \(\mathbf{x}_0\)'s closer to \(\mathbf{x}^{\ast}\) will also
make them closer to maximum-likelihood. If Lemma~\ref{lem-line} holds,
initial (potentially poor) summary statistics computed for the fiducials
points, \(\{\mathbf{x}_0^{(n)}\}_{n=0}^{N}\), can be improved by
training a CNF to generate new fiducial points,
\(\{\mathbf{x}_1^{(n)}\}_{n=0}^{N}\). Since these new fiducial points
are closer to their corresponding maximum likelihood then the new
summary statistics will be more informative, improving inference during
the next iteration where the CNF is trained on these improved summary
statistics.

\begin{figure}

\centering{

\includegraphics{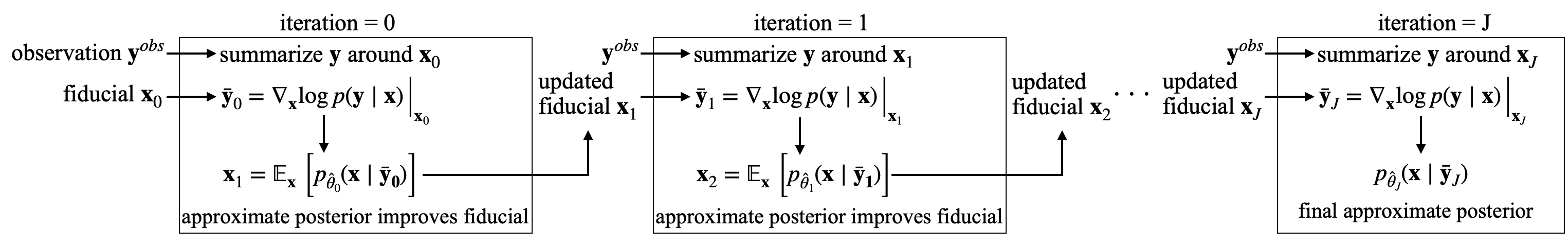}

}

\caption{\label{fig-algo-test}Online inference phase: our algorithm
refines the fiducial used to calculate the summary statistics. During
inference, it maintains the ability to amortize to many observations and
has low cost since \(J\) is typically a low number (3-4).}

\end{figure}

After training is completed, we obtain \(J\) trained CNFs, each with
their own set of optimized weights, \(\widehat \theta_j=0\cdots J-1\).
Because these networks are trained on
\(\{(\mathbf{x}_j^{(n)},\, \mathbf{\overline{y}}_j^{(n)})\}_{n=0}^{N}\, \text{for}\, j=0\cdots J-1\)
, these networks have been amortized to perform \(J\) refinements, given
a new unseen observation, \(\mathbf{y}^{\mathrm{obs}}\). A schematic for
the online inference phase is included in Figure~\ref{fig-algo-test}.
Note that each refinement incurs the cost of a gradient calculation. In
practice, \(J=3\) to \(4\) refinements are often adequate resulting in a
total online computational cost that is significantly lower than
non-amortized inference, which can result in \(10000\)'s of gradients
\citep{zhao2022bayesian}.

In summary, by pairing the theory of \citep{alsing2018generalized} with
Lemma~\ref{lem-line}, we arrived at a formulation where the refined
fiducial points yield improved summary statistics and refine amortized
VI at limited additional online computational costs. To verify this
claim, we will first evaluate our method on a stylized example for which
the analytical posterior is known. This example will show that the
improvements thanks to refined summary statistics indeed lead to
converge to the correct posterior distribution. To demonstrate our
amortized VI in a more practical setting, we will also evaluate its
performance on a realistic ultrasound transcranial medical imaging
problem. Finally, although the above lemma assumes that the fiducial
points reside in the correct basin of attraction, we believe that the
prior information contained in the paired datasets
\(\mathcal{D}_j,\,j=0\cdots J-1\) may mitigate the effect of local
minima. Indeed, our initial results in the nonlinear medical imaging
example start from a poor fiducial suggesting this might be possible.

\subsection{Stylized example}\label{stylized-example}

To build trust in our method, we first demonstrate it improves the
quality of the posterior approximation by testing on an inverse problem
with a known posterior distribution. One such inverse problem is the
linear Gaussian inverse problem where: the forward operator
\(\mathcal{F}\) is a known matrix \(\mathbb{R}^{m\times n}\), the prior
and noise comes from Gaussian distributions with known means and
covariances. We chose the unknown parameter vector to have size \(n=16\)
and the data as size \(m=80\). Given these settings, it is possible to
calculate samples from the analytical posterior distribution
\citep{bishop2006pattern}. We use samples from the analytical posterior
distribution to compare against our posterior sampling results with
ASPIRE.

\begin{figure}

\centering{

\includegraphics[width=0.9\textwidth,height=\textheight]{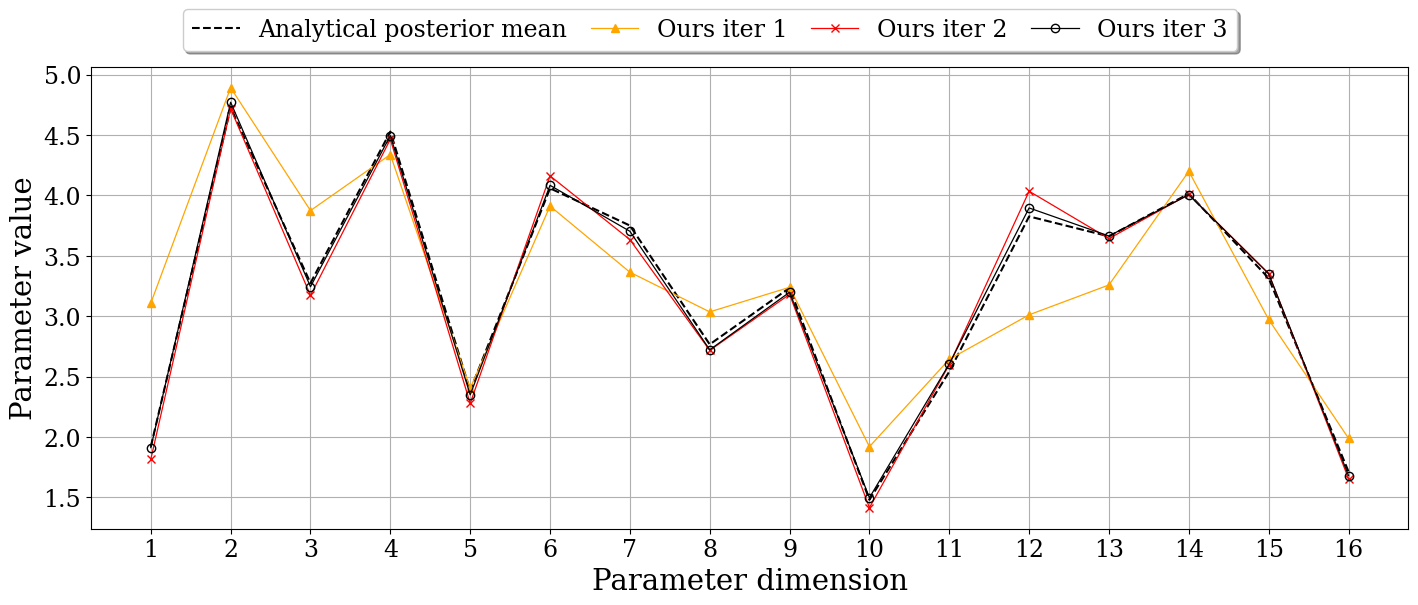}

}

\caption{\label{fig-means}The quality of the proposed amortized
posterior approximation improves at each iteration as measured by the
estimated posterior mean with respect to the analytically known ground
truth posterior mean.}

\end{figure}

We train our method using \(N=1000\) samples from the Gaussian prior and
use the forward operator to form training pairs. Since the forward
operator is linear, the score-based summary statistic is calculated from
the transposed operator. After training, we evaluate our method on an
unseen observation, \(\mathbf{y}^{\mathrm{obs}}\), simulated from a
known ground-truth parameter \(\mathbf{x}^{\ast}\). We first observe
that after each ASPIRE iteration the posterior mean
\(\mathbb{E} \, p_{{\widehat{\boldsymbol{\theta}}}_j}\) (using short
hand described in Equation~\ref{eq-pm}) becomes a better reconstruction
of the ground truth \(\mathbf{x}^{\ast}\) as seen in
Figure~\ref{fig-means}.

\begin{figure}

\centering{

\includegraphics{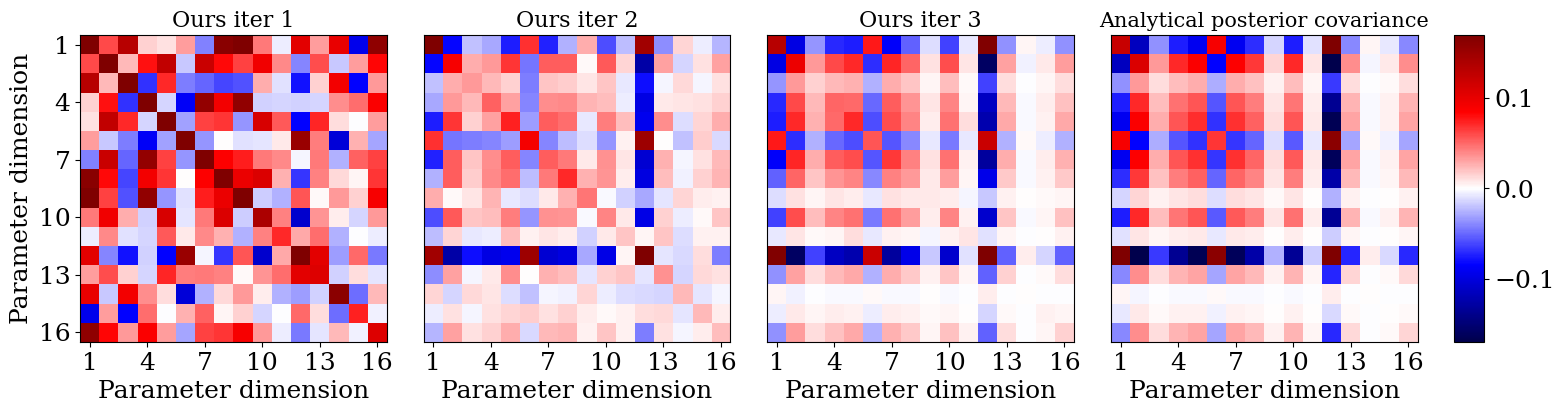}

}

\caption{\label{fig-covs}Comparison of full covariance matrix from our
method as compared to the analytical ground truth posterior covariance.
After three iterations of our method, the estimated posterior covariance
is close to the ground truth covariance.}

\end{figure}

Furthermore, Figure~\ref{fig-covs} shows the empirical covariance
derived from our method's posterior samples, compared to the
analytically calculated covariance matrix. It is clear that each
iteration improves the approximation of the estimated covariance and it
is almost exactly correct at the third iteration, \(J=3\). While this
stylized example confirms that amortized inference with ASPIRE in
principle be feasible, the real challenge is to apply this concept to
medical ultrasound where problems are high dimensional and forward
modeling computationally expensive to evaluate.

\section{Medical wave-based imaging}\label{sec-wavebased}

Transcranial Ultrasound Computed Tomography (TUCT) is a non-ionizing,
non-radiative imaging modality that creates images of brain tissue from
measurements of impinging ultrasound waves due to contrast in tissue
acoustic properties. Unlike other ultrasound imaging targets, like
breast imaging \citep{robins2021deep, wang2015waveform}, TUCT faces the
challenge of high acoustic contrast in the cranial bone, leading to
scattering unsuitable for traditional traveltime tomography methods
\citep{williamson1991guide}. Tomographic methods require high
frequencies for higher resolution imaging, but attenuation through the
skull is exacerbated at higher frequencies thus preventing
high-resolution imaging when relying on traveltime methods. This
challenge has hindered ultrasound's application to brain imaging until
recent developments when \citep{guasch2020full} identified that similar
challenges exist in transcranial ultrasound imaging as with sub-salt
imaging used by exploration seismology. The main reason seismic
techniques are capable of imaging through high-acoustic contrast salt is
because these sophisticated inversion methods model the full physics of
the wave equation to make sense of the scattered waves. Whereas
traditional ultrasound only uses arrival times, seismic imaging
techniques model all waveforms allowing for higher effective resolutions
at lower frequencies that experience less attenuation through the skull.
These methods are denoted Full-Waveform Inversion (FWI) since they model
and match the full observed waveform, see Figure~\ref{fig-dobs} for an
example of the full waveform.

\subsection{Medical ultrasound with full-waveform
inversion}\label{medical-ultrasound-with-full-waveform-inversion}

\begin{figure}

\begin{minipage}{0.33\linewidth}

\centering{

\includegraphics[width=0.9\textwidth,height=\textheight]{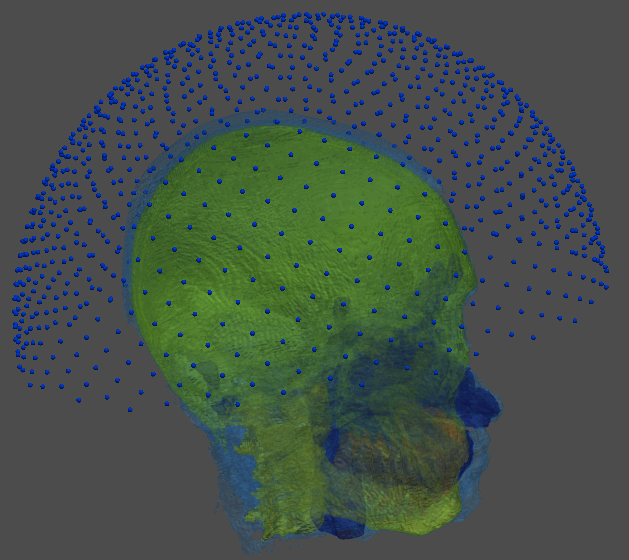}

}

\subcaption{\label{fig-helmet}3D setup.}

\end{minipage}%
\begin{minipage}{0.33\linewidth}

\centering{

\includegraphics{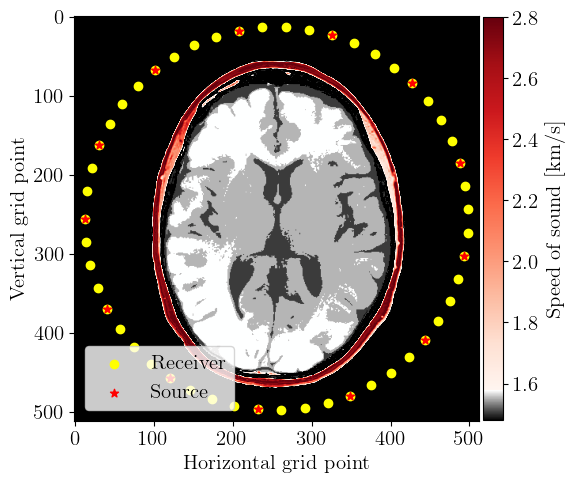}

}

\subcaption{\label{fig-setup}Acoustic velocity \(\mathbf{x}\).}

\end{minipage}%
\begin{minipage}{0.33\linewidth}

\centering{

\includegraphics{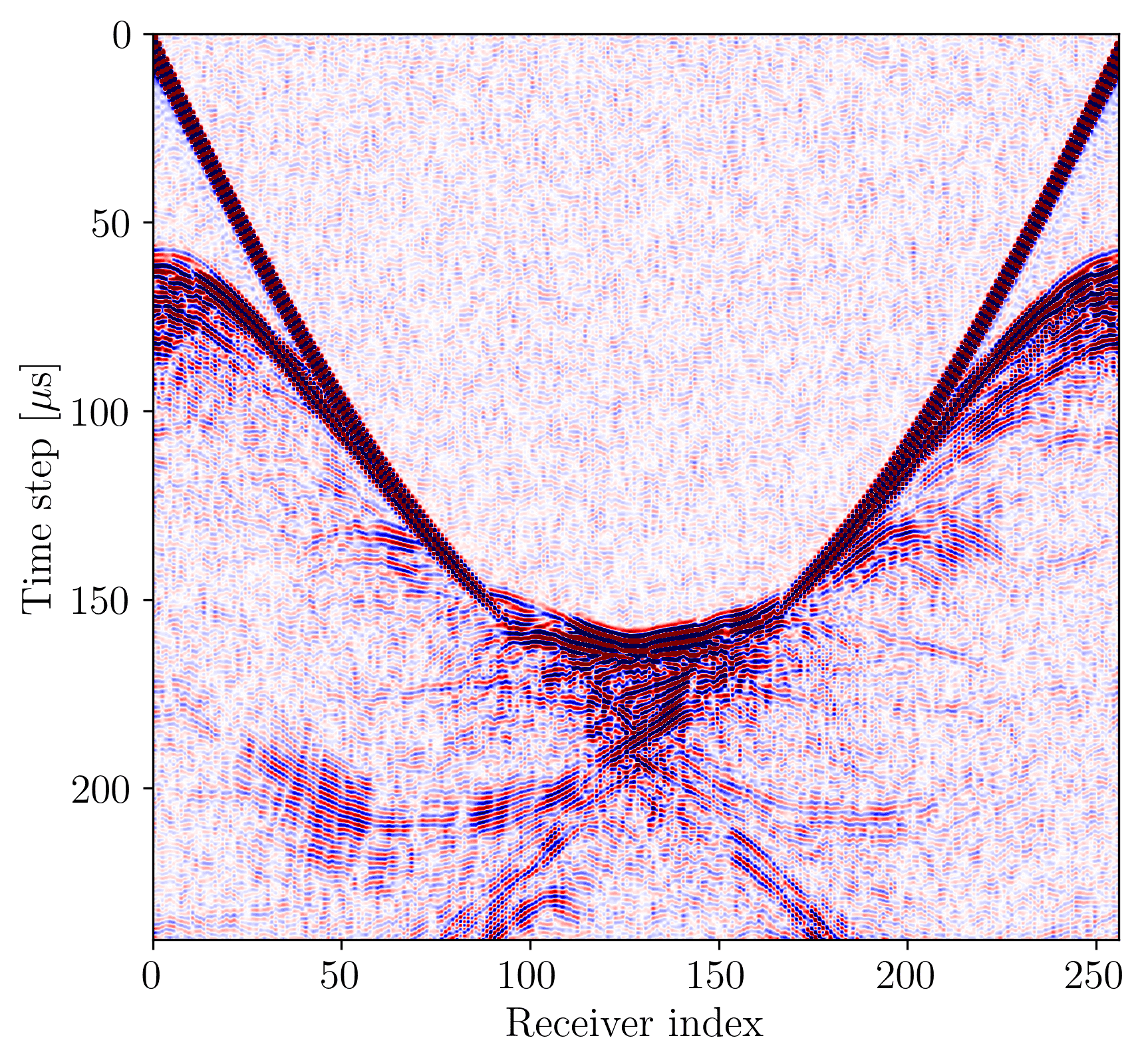}

}

\subcaption{\label{fig-dobs}Observation \(\mathbf{y}\)}

\end{minipage}%

\caption{\label{fig-simulation-setup}Experimental setup (a) Transcranial
ultrasound 3D setup as used in field, blue dots indicate transducers.
(b) Transcranial ultrasound 2D synthetic experimental setup used in this
work. (c) Simulated waveform from a single source synthetic experiment
\(\mathbf{y}\). Each column corresponds to acoustic-pressure amplitudes
measured at one transducer for a single experiment.}

\end{figure}

Since the groundbreaking work of \citep{guasch2020full}, FWI techniques
for TUCT are showing promise as a high-resolution imaging modality with
potential clinical applications ranging from early hemorrhage diagnosis
to tumor imaging \citep{robins2023dual, thomson2023ultrasonic}. The TUCT
inverse problem involves reconstruction of the acoustic velocity,
\(\mathbf{x}\), of brain tissue from acoustic data, \(\mathbf{y}\),
collected as shown in Figure~\ref{fig-helmet}. In this setup, ultrasound
transducers placed around the patient's head perform multiple
experiments, with each involving a tone-burst transmission by one
transducer and recording by all others, as simulated in
Figure~\ref{fig-dobs}. Experiments proceed by transducers transmitting
from different positions until all transducers have transmitted,
providing a full coverage from many angles. The forward operator
\(\mathcal{F}\) that maps the acoustic parameters to the observed data
is simulated with the numerical solution of the second-order wave
equation with varying acoustic velocity in two space dimensions:

\begin{equation}\phantomsection\label{eq-pde-wave}{
\frac{1}{c(x,y)^2}\frac{\partial^2}{\partial t^2}u(x,y,t) - \nabla^2 u(x,y,t) = q_s(x,y,t).
}\end{equation}

where the acoustic velocity \(c(x,y)\) is parameterized by a gridded
array of values in the unknown vector \(\mathbf{x}\) and the transducers
are modeled by \(N_s\) different source terms
\(\{q_s(x,y,t)\}_{s=0}^{s=N_s}\). Although the forward operator
\(\mathcal{F}\) is technically defined for each source as
\(\mathcal{F}(\mathbf{x}; q_s)\), for simplicity, we denote it as
\(\mathcal{F}(\mathbf{x})\), representing the collection of PDE
solutions for all transducer sources that contribute to all the
observations collected in \(\mathbf{y}\) and the restriction of the
solution wavefields to receiver positions. Given the set of
observations, traditional FWI workflows setup the variational problem

\[
  \underset{\mathbf{x} }{\operatorname{minimize}} \, \,
 \lVert \mathcal{F}(\mathbf{x}) - \mathbf{y} \rVert_2^2  
   \nonumber \\ 
\]

and minimize this data-misfit objective with stochastic gradient descent
by using randomized subsets of the sources to calculate the gradient of
each descent step beginning from the starting parameter vector,
\(\mathbf{x_0}\). This, of course, assumes access to an efficient
routine for calculating the gradient of \(\mathcal{F}\), which we will
discuss further in Section~\ref{sec-jacobian}. Under controlled
assumptions, such as a good starting parameter \(\mathbf{x_0}\) and
calibrated transducers \citep{cueto2021spatial}, FWI is known to produce
high-resolution images \citep{espin2023fwi, guasch2020full}. However,
clinical adoption of FWI is hampered by the prohibitively long runtime
of full physics modeling and the parasitic local minima related to the
non-convex optimization \citep{marty2021acoustoelastic}. Previous
literature has explored the regularization of the FWI optimization with
handcrafted priors such as the Total Variation norm
\citep{esser2016constrained, esser2018total} and model extensions
\citep{guasch2019adaptive, van2013mitigating}. Our approach, ASPIRE,
addresses these issues by reducing physics computations, giving
data-driven regularization of the non-convex optimization, and providing
uncertainty-aware solutions crucial for clinical applications by
sampling the Bayesian posterior.

\subsection{Transcranial Ultrasound Computed Tomography with
ASPIRE}\label{transcranial-ultrasound-computed-tomography-with-aspire}

To abridge, we will use ASPIRE to solve TUCT by generating samples from
a realistic brain velocity prior \(\mathbf{x} \sim p(\mathbf{x})\), use
a wave PDE solver \(\mathcal{F}\) to simulate acoustic data
\(\mathbf{y}\) and use the gradient of the simulator at a fiducial point
\(\mathbf{x}_0\) as the score-based summary statistic for training a CNF
with Equation~\ref{eq-train-cond}. Our test results demonstrate that
iterative refinement of this summary statistic through ASPIRE
significantly enhances the accuracy of the posterior approximations.
Now, we detail how each of these experiments and tests are implemented.

\subsubsection{Brain prior samples:}\label{brain-prior-samples}

The first step of implementing ASPIRE concerns obtaining samples from a
realistic prior for the target parameter vector
\(\mathbf{x}\sim p(\mathbf{x})\), in this case, gridded velocity
parameters of human brains and skulls. The parameters collected in the
MIDA dataset \citep{iacono2015mida} correspond to a single 3D volume for
the acoustic velocity collected from a single subject and will
unfortunately not be appropriate to train a neural model that will
generalize to other human patients. As far as we know, there is no
dataset that includes acoustic velocity collected from many patients, so
we made our own dataset based off the multi-subject FASTMRI dataset
\citep{zbontar2018fastmri}. This custom dataset, detailed in
Section~\ref{sec-prior-generation}, comprises N=1000 diverse acoustic
velocity parameters collected from different human patients,
\(\{\mathbf{x}^{(n)} \}_{n=0}^{1000}\). This size of datasets
facilitates generalization of the amortized posterior sampler across
different datasets collected from unseen patients. The dataset is
accessible via the repo
\href{https://github.com/slimgroup/ASPIRE.jl}{ASPIRE.jl}.

\subsubsection{Wave simulations:}\label{wave-simulations}

Our synthetic TUCT experiment, based on the configuration from
\citep{guasch2020full}, models the unknown parameter as discretized
acoustic velocity on a \(512 \times 512\) grid, with a
\(0.5\left[\mathrm{mm}\right]\) discretization. We modeled transducer
sources as point sources with a three-cycle tone-burst signature with
central frequency of \(400\mathrm{Khz}\) and \(240\) {[}microseconds{]}
recording time. The transducers are placed in a circular arrangement
around the skull, the setup, with \(16\) sources and \(256\) receivers,
mimics a 2D slice of the 3D experiment shown in Figure~\ref{fig-setup}.
The forward operator, \(\mathcal{F}(\mathbf{x})\), corresponds
simulating the forward waveforms and their restriction to the receiver
locations. The wave equation and its Jacobian were solved using the
open-source software packages Devito and JUDI
\citep{devito-api, devito-compiler, witte2019large}, which automatically
generate optimized C code and leverage GPU accelerators, thereby
facilitating scalability to realistic problem sizes. To simulate noise
corruption, we used additive Gaussian noise, \(\varepsilon\), with a
\(35\mathrm{dB}\) Signal-Noise-Ratio, matching lab values
\citep{guasch2020full}. A synthetic observation,
\(\mathbf{y}= \mathcal{F}(\mathbf{x}) + \varepsilon\), is displayed in
Figure~\ref{fig-dobs}.

\subsubsection{TUCT summary statistic:}\label{sec-jacobian}

The score-based summary statistic \(\mathbf{\overline y}\), is
calculated as in Equation~\ref{eq-score}, which requires the action of
the Jacobian adjoint on the data residual at the fiducial point,
\(\mathbf{x}_0\). For computational efficiency, the adjoint-state method
\citep{plessix2006review, virieux2009overview} is used. To avoid the
inverse crime, the observed data is simulated with finer time
discretization and a higher-order spatial finite-difference stencil than
those used in the residual calculation and adjoint simulation. Each
transducer defines a source term in Equation~\ref{eq-pde-wave}, so we
sum the gradient over all \(16\) sources into the final summary
statistic.

\subsection{Traditional amortized
inference}\label{traditional-amortized-inference}

\begin{figure}

\begin{minipage}{0.25\linewidth}

\centering{

\includegraphics{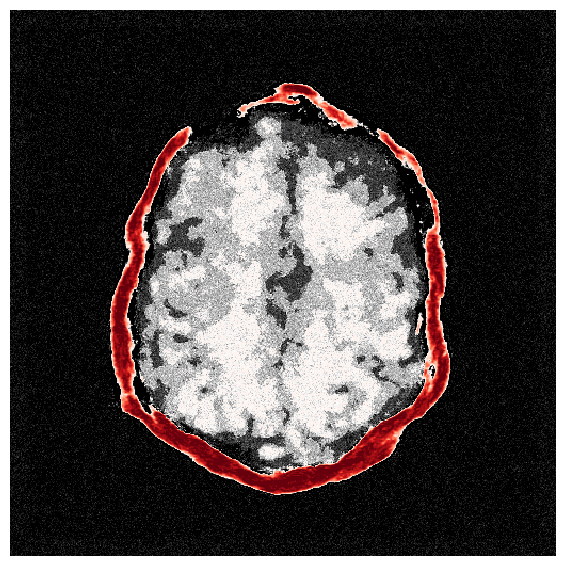}

}

\subcaption{\label{fig-post1-end2end}\(\mathbf{x}\sim p_{\widehat{\boldsymbol{\theta}}}\)}

\end{minipage}%
\begin{minipage}{0.25\linewidth}

\centering{

\includegraphics{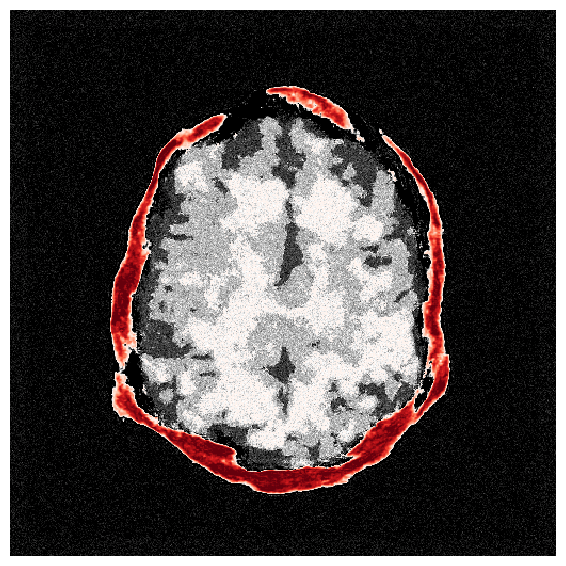}

}

\subcaption{\label{fig-post2-end2end}\(\mathbf{x}\sim p_{\widehat{\boldsymbol{\theta}}}\)}

\end{minipage}%
\begin{minipage}{0.25\linewidth}

\centering{

\includegraphics{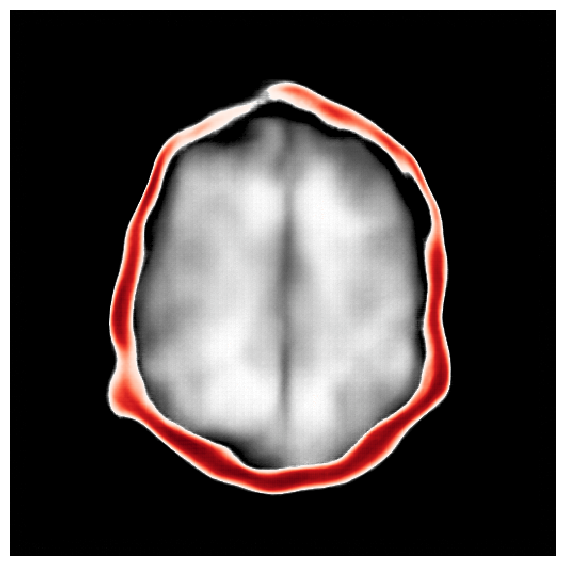}

}

\subcaption{\label{fig-mean-end2end}\(\mathbb{E} \, p_{\widehat{\boldsymbol{\theta}}}\)}

\end{minipage}%
\begin{minipage}{0.25\linewidth}

\centering{

\includegraphics{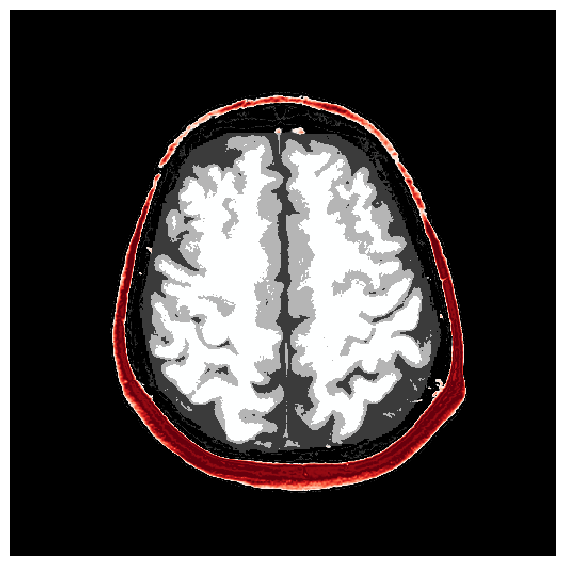}

}

\subcaption{\label{fig-gt-1}\(\mathbf{x}^{\ast}\)}

\end{minipage}%

\caption{\label{fig-fromacoustic}Baseline amortized inference. (a),(b)
Posterior samples. (c) Posterior mean. (d* Ground-truth velocity
parameters paired to test observation
\(\mathbf{y}^{\mathrm{obs}} = \mathcal{F}(\mathbf{x}^{\ast})+\varepsilon\).
The samples have poor quality since it is difficult to learn the direct
mapping from acoustic waveforms to the unknown parameter.}

\end{figure}

To illustrate the limitations of amortized VI, we train a CNF on pairs
\(\{(\mathbf{x}^{(n)},\mathbf{y}^{(n)}) \}_{n=0}^{N}\) without evoking
iterative improvements by ASPIRE. We emphasize that the observations
\(\mathbf{y}\) are the raw unsummarized waveforms similar to that shown
in Figure~\ref{fig-dobs}. Training details are included in
Section~\ref{sec-training}. After training by minimizing
Equation~\ref{eq-train-cond}, the CNF with weights
\(\widehat{\boldsymbol{\theta}}\), provides an amortized approximation
of the posterior,
\(p_{\widehat{\boldsymbol{\theta}}} \approx p(\mathbf{x}\mid\mathbf{y})\),
from which we can sample (cf. Equation~\ref{eq-sampling}). The results,
shown in Figure~\ref{fig-fromacoustic}, demonstrate that the samples
from \(p_{\widehat{\boldsymbol{\theta}}}\), for an unseen test
observation, \(\mathbf{y}^{\mathrm{obs}}\), lack distinct features
beyond an unrealistic skull and unresolved internal tissue structure. A
comparison of these samples and the posterior mean,
\(\mathbb{E} \, p_{\widehat{\boldsymbol{\theta}}}\), in
Figure~\ref{fig-mean-end2end} with the ground truth,
Figure~\ref{fig-gt-1}, highlights the poor quality of this
approximation. Note, throughout this exposition we calculate the
posterior statistics (i.e.~mean and standard deviation) over \(512\)
samples, please consult the Section~\ref{sec-num-post-samples} for a
discussion on this quantity. This experiment underscores the challenge
of directly learning the probabilistic inverse mapping from the acoustic
data \(\mathbf{y}\) to the velocity parameters, a difficulty previously
noted in the literature \citep{orozco2023adjoint}. This challenge is
often referred to as the `end-to-end' problem \citep{mukherjee2021end}.
We address this problem with the score-based summary statistic employed
by ASPIRE.

\subsection{Amortized inference with score-based summary
statistics}\label{amortized-inference-with-score-based-summary-statistics}

\begin{figure}[H]

\begin{minipage}{0.50\linewidth}

\centering{

\includegraphics[width=0.75\textwidth,height=\textheight]{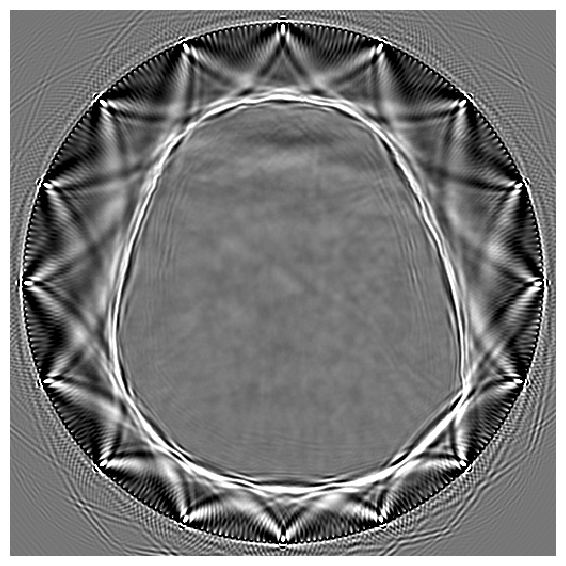}

}

\subcaption{\label{fig-grad-iter1}Initial summary statistic
\(\mathbf{\overline y}_0\).}

\end{minipage}%
\begin{minipage}{0.50\linewidth}

\centering{

\includegraphics[width=0.75\textwidth,height=\textheight]{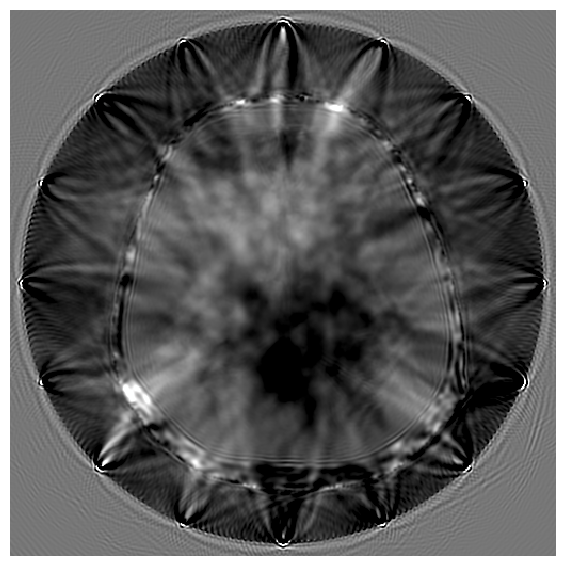}

}

\subcaption{\label{fig-grad-iter2}Refined summary statistic
\(\mathbf{\overline y}_1\).}

\end{minipage}%

\caption{\label{fig-gradients}Score-based summary statistics (a) First
summary statistic calculated at the \(\mathbf{x}_0\) fiducial consisting
of constant water velocity. (b) Second summary statistic calculated at
the fiducial point, \(\mathbf{x}_1\), derived from the first CNF
posterior mean. Thanks to improved fiducial point, we now can
``illuminate'' the inside of the skull.}

\end{figure}

\begin{figure}[H]

\begin{minipage}{0.25\linewidth}

\centering{

\includegraphics{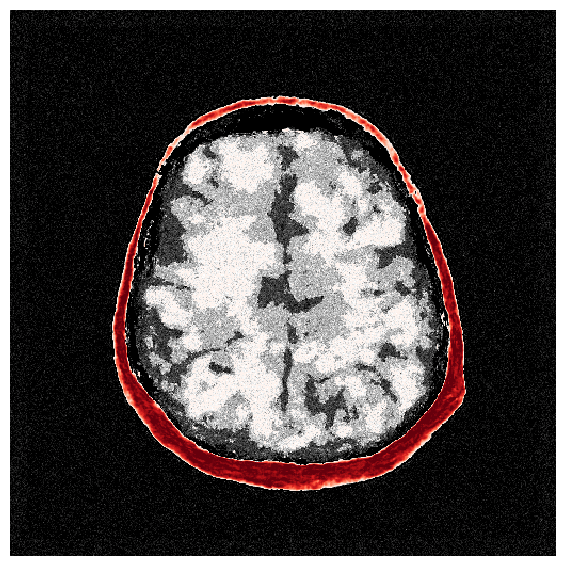}

}

\subcaption{\label{fig-post1-iter-1}\(\mathbf{x}\sim p_{\widehat{{\theta}}_1}\)}

\end{minipage}%
\begin{minipage}{0.25\linewidth}

\centering{

\includegraphics{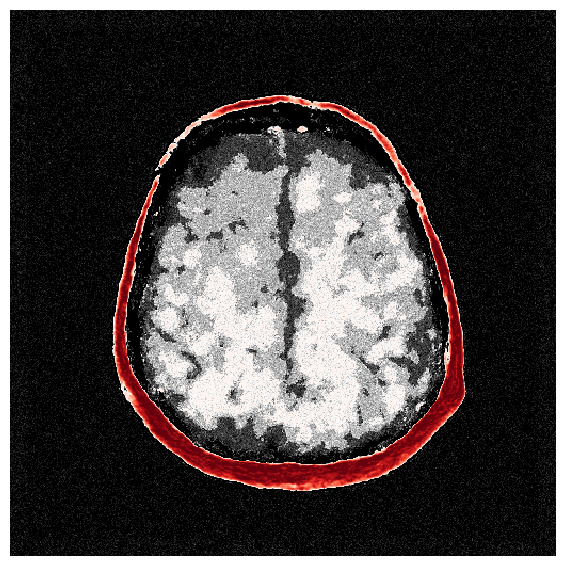}

}

\subcaption{\label{fig-post2-iter-1}\(\mathbf{x}\sim p_{\widehat{{\theta}}_1}\)}

\end{minipage}%
\begin{minipage}{0.25\linewidth}

\centering{

\includegraphics{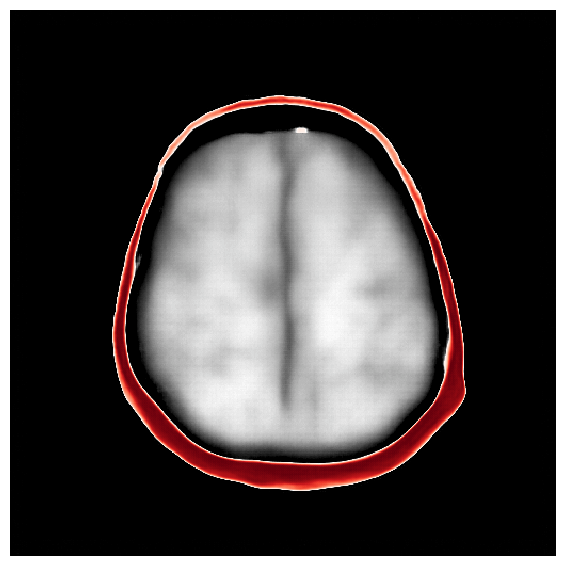}

}

\subcaption{\label{fig-mean-iter-1-1}\(\mathbb{E} \, p_{\widehat{{\theta}}_1}\)}

\end{minipage}%
\begin{minipage}{0.25\linewidth}

\centering{

\includegraphics{./figs/indx=24_gt.png}

}

\subcaption{\label{fig-gt-2}\(\mathbf{x}^{\ast}\)}

\end{minipage}%

\caption{\label{fig-aspire-1}The first iteration of our method learns
the mapping from the summarized data \(\mathbf{\overline y}\) to unknown
parameter \(\mathbf{x}\) (a),(b) Posterior samples. (c) Posterior mean.
(d) Ground-truth velocity parameters. Our method has learned to
reconstruct a reasonable estimate of the skull outline by making use of
the summary statistic.}

\end{figure}

To overcome the end-to-end inference problem, we apply one iteration of
ASPIRE by training a CNF on pairs,
\(\mathcal{D}_0 = \{(\mathbf{x}^{(n)}, \mathbf{\overline y}_0^{(n)}) \}_{n=0}^{N}\),
where the \(\mathbf{\overline y}_0^{(n)}\)'s represent the score-based
summary statistics at the fiducial points, \(\mathbf{x}_0^{(n)}\), taken
to be the uniform water velocity for all samples. An example of this
initial summary statistic is shown in Figure~\ref{fig-grad-iter1}. While
the outer edge of the skull is reasonably well delineated, the inner
edge of the skull is still poorly resolved and details inside the skull
are mostly absent. However, the inference based on these initial summary
statistics, shown in Figure~\ref{fig-mean-iter-1-1}, present a
significant improvement over the baseline (cf.
Figure~\ref{fig-mean-end2end}), despite the presence of strong imaging
artifacts in the summary statistics. The improvements concern the
skull's structure in particular, although details within the skull
remain elusive due to the summary statistic's limited information. To
enhance fidelity further, ASPIRE 2 (shorthand for ASPIRE at iteration
\(J=2\)) is applied by recalculating the score at the new posterior mean
estimate for each training sample. Given these new training pairs, the
next CNF is trained. While posterior sampling is efficient with CNFs
(using Equation~\ref{eq-sampling}), recalculation of the score for each
sample is computationally intensive, a topic we address in
Section~\ref{sec-computation}.

\subsection{Amortized inference with iterative
refinements}\label{amortized-inference-with-iterative-refinements}

\begin{figure}[H]

\begin{minipage}{0.25\linewidth}

\centering{

\includegraphics[width=1\textwidth,height=\textheight]{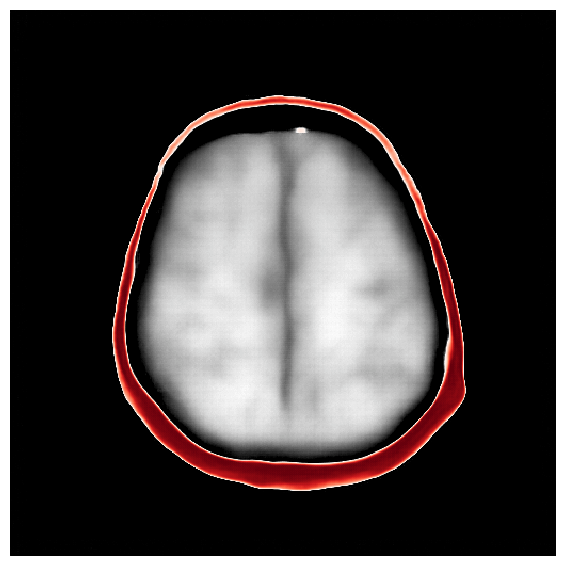}

}

\subcaption{\label{fig-mean-iter1}ASPIRE 1
\(\mathbb{E}\, p_{\widehat{\boldsymbol{\theta}}_1}\)}

\end{minipage}%
\begin{minipage}{0.25\linewidth}

\centering{

\includegraphics[width=1\textwidth,height=\textheight]{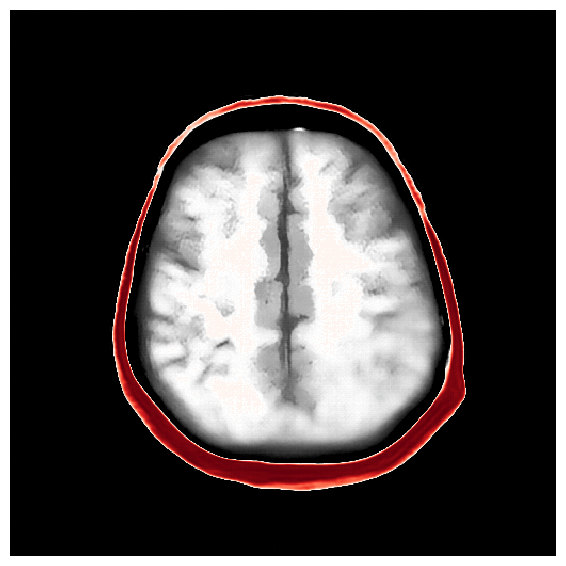}

}

\subcaption{\label{fig-mean-iter2}ASPIRE 2
\(\mathbb{E}\,p_{\widehat{\boldsymbol{\theta}}_2}\)}

\end{minipage}%
\begin{minipage}{0.25\linewidth}

\centering{

\includegraphics[width=1\textwidth,height=\textheight]{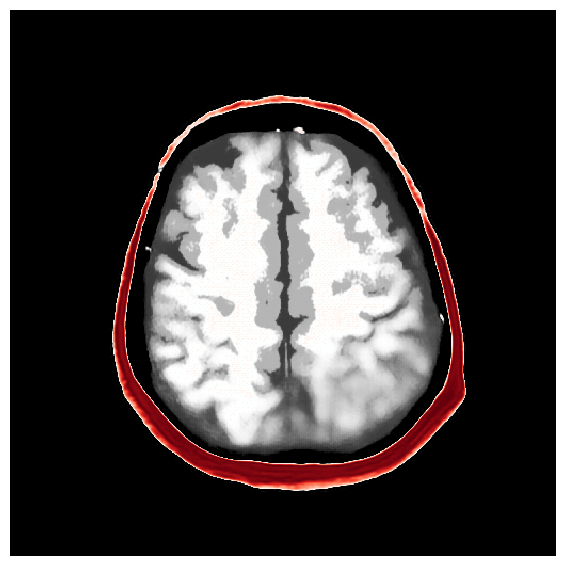}

}

\subcaption{\label{fig-mean-iter4}ASPIRE 4
\(\mathbb{E} \,p_{\widehat{\boldsymbol{\theta}}_4}\)}

\end{minipage}%
\begin{minipage}{0.25\linewidth}

\centering{

\includegraphics[width=1\textwidth,height=\textheight]{./figs/indx=24_gt.png}

}

\subcaption{\label{fig-gt-3}Ground truth \(\mathbf{x}^{\ast}\)}

\end{minipage}%

\caption{\label{fig-reconstruction}The posterior approximation improves
as measured by the posterior mean quality. (a) Posterior mean from
ASPIRE 1 where the observation is preprocessed with the gradient as
summary statistic. (b) Posterior mean from ASPIRE 2 where the summary
statistic has been refined using the posterior mean from the first
iteration. (c) Posterior mean from ASPIRE 4. (d) Ground-truth.}

\end{figure}

After the refinements of ASPIRE 2, significant improvements are evident
in the posterior samples, particularly in capturing the structures
within the brain tissue itself. The mean of these posterior samples,
displayed in Figure~\ref{fig-mean-iter2}, is clearly enhanced in
resolution and details. We attribute these enhancements to the increased
informativeness of the summary statistic in the second iteration
compared to the information yielded by the initial iteration. A detailed
inspection of the second summary statistic (shown in
Figure~\ref{fig-grad-iter2}) reveals more detail on the internal brain
structures. Unlike the first summary statistic (cf.
Figure~\ref{fig-grad-iter1}), which primarily delineated the skull, the
second iteration's summary statistic better `illuminates' the softer
tissues within the brain, offering a more informative image for the
posterior network. Thanks to accounting for the scattering at the skull,
the acoustic illumination of the brain is improved significantly.
Accurately resolving the skull structure is an important consideration
as noted by \citep{marty2023shape}.

As one can observe from Figure~\ref{fig-reconstruction}, the
reconstruction quality improves for increasing number of refinements of
ASPIRE. By virtue of the iterative recalculation of the score-based
summary statistic, the method is progressively able to discern finer
details within the brain albeit the updates become less pronounced as
the number of refinements increases. We further illustrate this
refinement by plotting posterior samples from all four ASPIRE iterations
in Figure~\ref{fig-more-post}. Practically, a user of ASPIRE can decide
on the number of refinements based on the amount of compute available or
by refining until there are diminished returns on enhancements.

\subsection{Reconstruction quality}\label{reconstruction-quality}

\begin{figure}[H]

\centering{

\includegraphics[width=0.5\textwidth,height=\textheight]{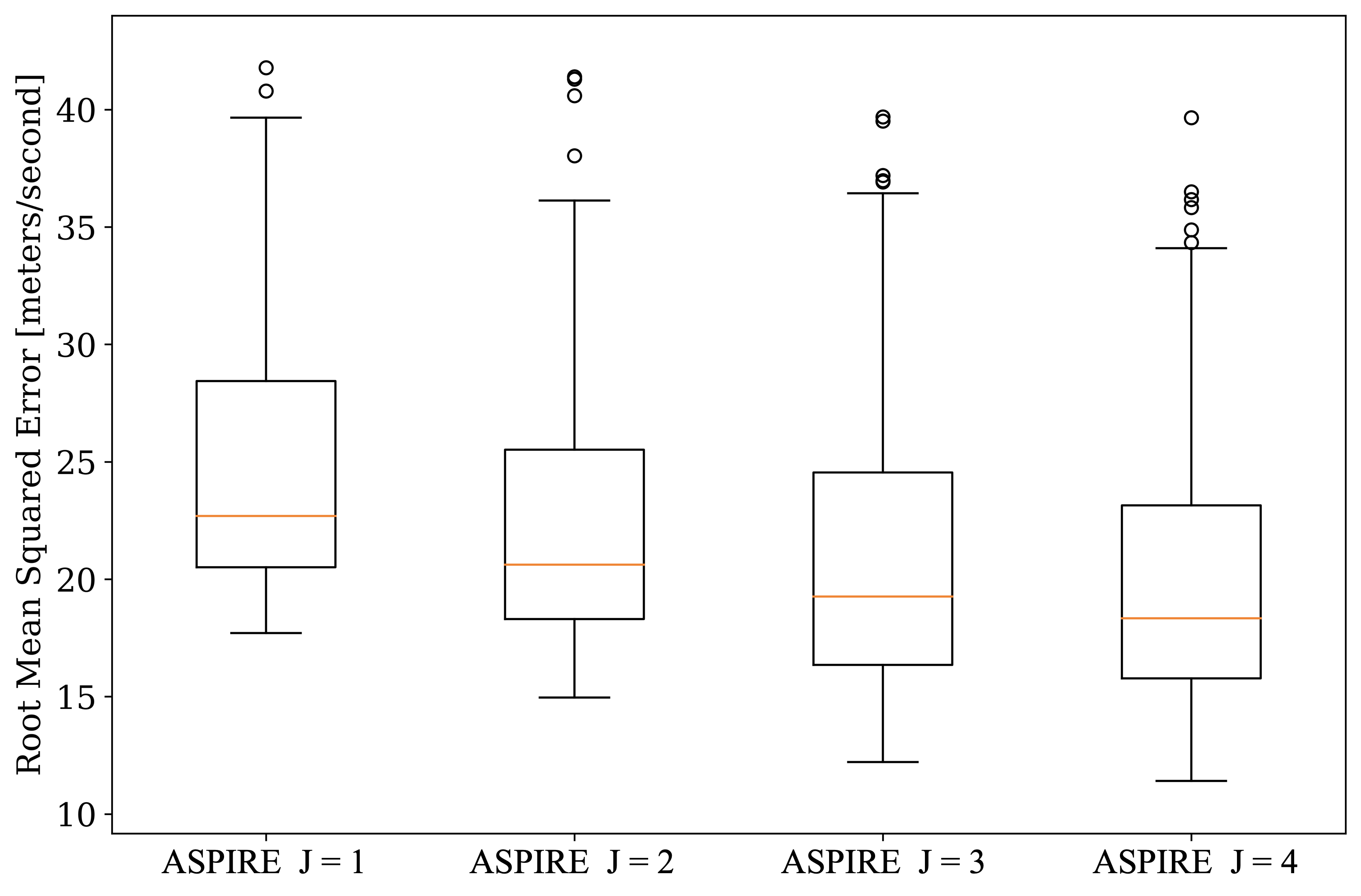}

}

\caption{\label{fig-improvement-mean}Image quality metric measured on
average over 50 sample leave-out test set. The quality of the posterior
mean improves after each ASPIRE refinement.}

\end{figure}

To quantitatively assess enhancements of each ASPIRE iteration, we
compare the posterior means with their corresponding ground truths,
using a test set comprising \(50\) unseen observations. By calculating
the Root Mean Squared Error (RMSE) for these comparisons, we establish a
metric to quantify the improvements across iterations. At each iteration
we plot the RMSE for all examples in the test set and plot a box plot.
We emphasize that testing on this many samples was only tractable since
the posterior sampler is amortized. The trend, as showcased in
Figure~\ref{fig-improvement-mean}, confirms that on average each
iteration reduces the RMSE, indicating an increasingly precise
approximation of the true posterior means. We hypothesize that we are
observing the effect known as Bayesian contraction
\citep{ghosal2017fundamentals} since each summary statistic is
extracting more information from the observations.

While the reconstructions in Figure~\ref{fig-reconstruction} clearly
improve as the number refinements increases, certain areas remain
smooth, especially near the top and lower-right corner. This smoothing
effect of the posterior mean is well-known and arises from relative
strong variations among the samples in certain areas. Variations between
samples are a reflection of inconsistent reconstructions by the
posterior samples and correspond to areas of increased uncertainty. This
phenomenon is a direct result of treating ultrasound medical imaging as
an inference problem that produces posterior distributions instead of a
single answer.

\section{Uncertainty quantification}\label{uncertainty-quantification}

Due to the risk of hallucinations, generative AI for imaging inverse
problems benefits from uncertainty awareness. Furthermore, an
uncertainty-aware approach becomes crucial in medical applications, as
underscored by \citep{barbano2022uncertainty}. Fortunately, Bayesian
posterior sampling provides a natural sense of uncertainty, reflected in
the spread of the samples. ASPIRE is designed with amortized posterior
sampling in mind to quickly deliver crucial uncertainty quantification.
By providing both the mean and a uncertainty information, our method
offers a dual perspective, namely a robust reconstruction of the
tissues, complemented by an insight into the statistical variations of
each pixel's value. This dual analysis is particularly valuable in
medical diagnostics, where understanding both the image and the
associated uncertainty is crucial for decision making.

\subsection{Amortized uncertainty
quantification}\label{amortized-uncertainty-quantification}

\begin{figure}

\begin{minipage}{\linewidth}

\centering{

\includegraphics{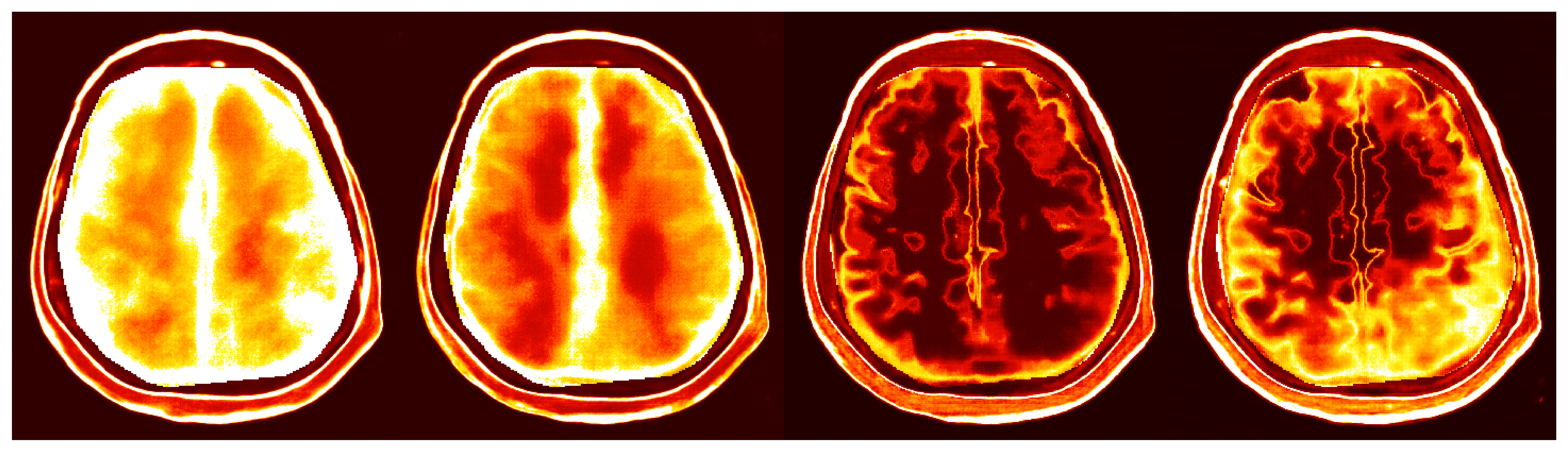}

}

\subcaption{\label{fig-vars-iter}\(\sqrt{\mathbb{V}} \, \, p_{\widehat{\boldsymbol{\theta}}_j}\)}

\end{minipage}%
\newline
\begin{minipage}{\linewidth}

\centering{

\includegraphics{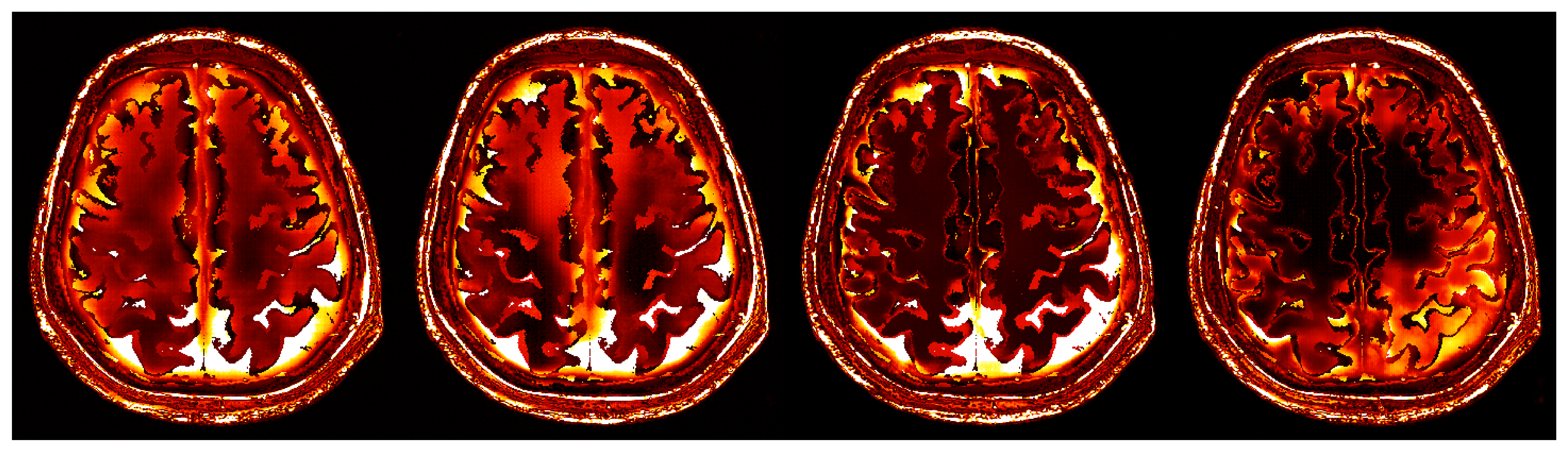}

}

\subcaption{\label{fig-errors-iter}Error
\(|\mathbf{x}^{\ast} - \mathbb{E}_{} \, p_{\widehat{\boldsymbol{\theta}}_j}|\)}

\end{minipage}%

\caption{\label{fig-variance-mask}Posterior standard deviation compared
to predictive error. (a) Posterior standard deviations of ASPIRE with
increasing iterations 1 through 4 from left to right. (b) Same but with
error of the posterior mean. Each ASPIRE refinement uncovers higher
resolution details; furthermore the apparent correlation between the
uncertainty and error increases. All plots are shown on the same
colorbar from \(0\mathrm{[m/s]}\) to \(50\mathrm{[m/s]}\)}

\end{figure}

To visualize uncertainty, we calculate uncertainty images by taking the
pixel-wise standard deviations \(\sqrt{\mathbb{V}}\) as defined in
Equation~\ref{eq-stdev} with \(512\) posterior samples.
Figure~\ref{fig-variance-mask} shows uncertainty images for four
iterations of ASPIRE alongside the error of the posterior mean from the
ground truth. From these figures, we make the following qualitative
observations: \emph{(1)} uncertainty images increase in resolution with
each ASPIRE refinement \emph{(2)} refinements increase correlations
between the uncertainty and the error. Specifically, the errors
concentrate near the top and lower-right of the internal brain tissue.
The reason being that high acoustic contrast in these areas is creating
multiple reverberations of the wavefield inside the brain impeding
accurate imaging, importantly these are areas that are highlighted by
the uncertainty. Correlations between the uncertainty and the error
constitute important empirical evidence of the trustworthiness of the
uncertainty. To more rigorously quantify this correlation, and
quantitatively validate the uncertainty quantification, we study the
calibration of our uncertainty in the following section.

\subsection{Calibration of the uncertainty}\label{sec-calibration}

To assess the calibration of our method's uncertainty quantification
against errors, we employ the calibration test described by
\citep{guo2017calibration, laves2020well}. This test involves
comparisons between errors --- defined by the Euclidean distance between
the posterior mean estimates,
\(\widehat{\mathbf{x}}=\mathbb{E} \, p_{\widehat{\theta}}\), derived
from samples of the posterior conditioned on the observations,
\(\mathbf{y}\), and the ground-truth parameters, \(\mathbf{x}^\ast\) ---
and the inferred uncertainty in terms of the square-root of the
posterior variance,
\(\widehat{\sigma}=\sqrt{\mathbb{V}}p_{\widehat{\boldsymbol{\theta}}}\).
Given predictions, \(\widehat{\mathbf{x}}\), derived from observations,
\(\mathbf{y}\), and a measure of uncertainty, \(\widehat \sigma\), the
calibration test seeks to verify the relationship: \[
\mathbb{E}_{\mathbf{x}^\ast,\mathbf{y}}\Bigl[ \|\mathbf{x}^\ast - \widehat{\mathbf{x}}\|_2^2 \mid \widehat{\sigma} = \sigma\Bigr] = \sigma \quad  \{\forall \sigma \in \mathbb{R} \mid \sigma \geq 0\}.
\]

This expression implies that uncertainty is properly calibrated when the
uncertainty is proportional to the error. For instance, if a set of
gridpoints has an uncertainty of \(10\mathrm{[m/s]}\), their expected
error should be \(10\mathrm{[m/s]}\). The calibration benchmark follows
as such, first the set of uncertainty values for each pixel in
\(\widehat{\sigma}=\sqrt{\mathbb{V}}p_{\widehat{\boldsymbol{\theta}}}\)
is categorized into \(K\) bins of equal width, the uncertainty at each
bin \(B_k\) is calculated as: \[
UQ(B_k) := \frac{1}{|B_k|} \sum_{i\in B_k}\widehat{\sigma}_i
\] the average error (with
\(\widehat{\mathbf{x}}=\mathbb{E} \, p_{\widehat{\boldsymbol{\theta}}}\))
is also calculated at the same bins: \[
Error(B_k) := \frac{1}{|B_k|} \sum_{i\in B_k}(\mathbf{x}_i^{\ast} - \widehat{\mathbf{x}}_i)^2.
\] The uncertainty \(UQ(B_k)\) and error \(Error(B_k)\) at each bin is
then plotted against each other. If there is a high correlation between
these values we expect the plot to match the \(45^\circ\)-degree angle.
For details on this test see \citep{laves2020well}.

\begin{figure}[H]

\centering{

\includegraphics[width=0.5\textwidth,height=\textheight]{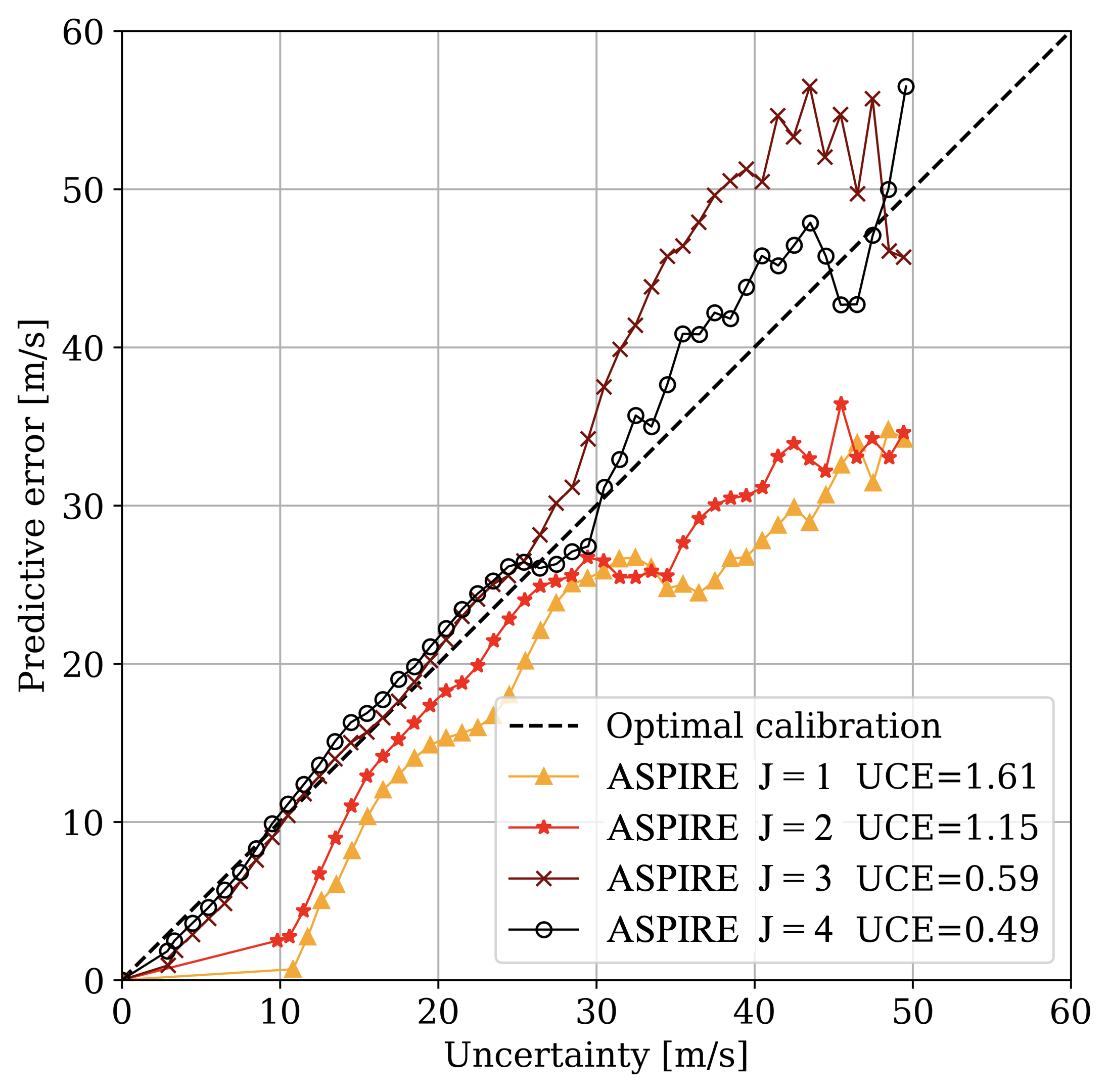}

}

\caption{\label{fig-improvement-uq}Calibration plot of four refinements
from ASPIRE. The quality of uncertainty quantification of ASPIRE
improves as measured by the calibration with respect to the error.}

\end{figure}%

The resulting calibration curve from ASPIRE 4, included in
Figure~\ref{fig-improvement-uq}, exhibits the expected behavior by
closely following the expected line. This means that our method is well
calibrated for error magnitudes up to \(30 \mathrm{[m/s]}\). More
importantly, each iterative refinement of ASPIRE improves the
calibration. To quantify these improvements, we calculate the
Uncertainty Calibration Error (UCE), which represents the average
absolute difference between the predicted error and actual uncertainty
across all bins. A lower UCE indicates a more precise calibration, and
according to our experiment, ASPIRE manages to reduce the UCE by a
factor of three from a value of \(1.61\) to \(0.49\). This significant
improvement underscores the ability of our iterative refinement approach
to produce reliable uncertainty estimates in complex imaging scenarios.
By confirming the calibration of our uncertainty, we build trust in our
method to properly inform downstream tasks that require access to the
Bayesian posterior.

\section{Discussion}\label{discussion}

Our main goal is to convince the reader of ASPIRE's ability to close the
amortization gap with iterative refinements. We were able to show that
significant improvements were achieved compared to the baseline
(e.g.~juxtapose Figure~\ref{fig-mean-iter1} and
Figure~\ref{fig-mean-iter4}). To further substantiate our claims, we
place our amortized method's performance in a broader context that
includes comparisons to non-amortized inference. These comparisons
illustrate our strides towards narrowing the gap between these two
paradigms. Although a wide range of non-amortized VI techniques are
available, our application in medical ultrasound presents unique
requirements and challenges --- absence of an analytical prior and the
need for short time-to-solution challenged by the computational
intensity of the forward/adjoint operators ---- limit our options for
comparison. For this reason, we selected two methods: the mean-field
approximation and a non-amortized normalizing flow. Through this
comparative analysis, we aim to highlight our method as an efficient
alternative to scenarios where non-amortized methods may be impeded by
computational demands.

\subsection{Mean-field approximation}\label{mean-field-approximation}

Perhaps, the method with the most comparable aims in the current
literature for uncertainty quantification in TUCT is the recent work by
\citep{bates2022probabilistic}. This approach employs a mean-field
approximation of the posterior, and assumes the posterior covariance
matrix to be diagonal for computational reasons. While understandable
from a computational perspective, the established FWI literature
cautions against this assumption because of the non-diagonal nature of
the wave-equation Hessian \citep{virieux2009overview}, which introduces
correlations in the errors. Our method does not make statistical
assumptions on the prior or likelihood and is designed to capture the
complete statistics of the posterior distribution, including long-range
correlations in the covariance as evidenced in Figure~\ref{fig-covs}.

Given a single set of observations, \(\mathbf{y}^{\mathrm{obs}}\) the
mean-field algorithm by \citep{bates2022probabilistic} directly outputs
estimates for the posterior mean, \(\boldsymbol{\mu}_{m}\), and
point-wise posterior standard deviation, \(\boldsymbol{\sigma}_{m}\).
This approach hinges on a relative simple modification of traditional
FWI --- it multiplies updates of gradient-descent with a Gaussian field.
Because the computational cost of the mean-field approximation roughly
correspond to that of traditional FWI that includes 100s of forward and
adjoint calls, we argue that our method offers a distinct computational
advantage at inference time. ---i.e., ASPIRE achieves online inference
at approximately 1/100th the online computational cost of the mean-field
approximation. For a detailed discussion on computational costs, please
refer to Section~\ref{sec-computation} below. We also observed that the
solution from the mean-field is similar to traditional FWI thus to avoid
repetition we only include mean-field results.

\subsection{Non-amortized inference with normalizing
flows}\label{non-amortized-inference-with-normalizing-flows}

We also compare with a non-amortized method deriving from the same prior
knowledge in the form of training samples. Given the problem size and
expensive to evaluate wave-physics based forward operators and
gradients, these comparisons are made with respect to a novel
computationally efficient inference method inspired by recent work of
\citep{siahkoohi2023reliable, detommaso2019hint}. Instead of starting
from scratch with a non-informative prior \citep{zhang20233}, which
proves to be computationally prohibitively expensive, the proposed
approach optimizes network weights, \(\boldsymbol{\phi}\), of a
non-amortized normalizing flow (NF), \(g_{\boldsymbol{\phi}}(\cdot)\)
that acts in the latent space of a pre-trained amortized CNF,
\(f_{\widehat{\boldsymbol{\theta}}}(\cdot)\). Given a single set of
observations, \(\mathbf{y}^{\mathrm{obs}}\), the objective reads:
\begin{align}
\underset{\boldsymbol{\phi}}{\operatorname{minimize}} \, \, \, & \mathbb{KL}\left(p\left(h_{\boldsymbol{\phi}}(\mathbf{z})\right)\Vert\, p(\mathbf{z} \mid \mathbf{y}^{\mathrm{obs}})\right) \nonumber \\
= \, \,& \mathbb{E}_{\mathbf{z} \sim \mathcal{N}(\mathbf{0}, \mathbf{I})} \left[ \frac{1}{2\sigma^2} \left\| \mathcal{F} \circ f_{\boldsymbol{\widehat{\theta}}}^{-1}\left(g_{\boldsymbol{\phi}}(\mathbf{z}); \mathbf{\overline{y}}^{\mathrm{obs}}\right) - \mathbf{y}^{\mathrm{obs}} \right\|_2^2 \right. \nonumber \\
& \quad \quad \quad \quad  \, \, \, \, + \left. \frac{1}{2} \left\| g_{\boldsymbol{\phi}}(\mathbf{z}) \right\|_2^2 - \log \left| \det \mathbf{J}_{g_{\boldsymbol{\phi}}}(\mathbf{z}) \right| \right]. \label{eq-train-composition}
\end{align}

The \(f_{\widehat{\boldsymbol{\theta}}}(\cdot)\) denotes the pre-trained
CNF, optimized as per Equation~\ref{eq-train-cond} and
\(\left|\det\mathbf{J}_{g_{\boldsymbol{\phi}}}\right|\) is the
determinant of the second network's Jacobian. By minimizing this
objective, the network \(g_{\boldsymbol{\phi}}(\cdot)\) is trained to
generate latent codes that further minimize residuals in data-misfit
objective, which involves the nonlinear forward operator, and a
\(\ell_2\)-norm penalty term, which ensures that the network output
stays close to Gaussian distributed, therefore respecting the prior
defined by the pre-trained network
\(f_{\widehat{\boldsymbol{\theta}}}(\cdot)\). To avoid having to
calculate the forward map and its gradient at each iteration, we follow
\citep{siahkoohi2020weak} and replace the strong constraint in
\ref{eq-train-composition} by a weak constraint that allows for an
outer-inner-loop optimization algorithm. The optimization alternates
between an expensive outer loop with \(L\) iterations inside
\{++during++\} which \(L_{\textrm{inner}}\gg L\) iterations of an
inexpensive inner loop are performed. The forward operator and its
gradients are only calculated once during each outer loop iteration
while the inner loop contains several updates to the networks. Through
Monte-Carlo approximation of the above expectation, we arrive at this
weak formulation by introducing \(N_p\) slack variables,
\(\mathbf{x}_{1:N_p}\), that alongside the network weights are minimized
in the following objective: \begin{align}
 \underset{\mathbf{x}_{1:N_p}, \boldsymbol{\phi}}{\operatorname{minimize}} \,  \frac{1}{N_p} \sum_{n=0}^{N_p}& \left[  \frac{1}{2\sigma^2} \left\| \mathcal{F}(\mathbf{x}^{(n)}) - \mathbf{y}^{\mathrm{obs}} \right\|_2^2 \right.  + \frac{1}{2\gamma^2} \left\| \mathbf{x}^{(n)} - f_{\boldsymbol{\widehat{\theta}}}^{-1}\left(g_{\boldsymbol{\phi}}(\mathbf{z}^{(n)}); \mathbf{\overline y}^{\mathrm{obs}}\right) \right\|_2^2 \nonumber \\
 & \left. + \frac{1}{2} \left\| g_{\boldsymbol{\phi}}(\mathbf{z}^{(n)}) \right\|_2^2 - \log \left| \det \mathbf{J}_{h_{\boldsymbol{\phi}}}(\mathbf{z}^{(n)}) \right| \right] \label{eq-train-composition-weak}
\end{align}

where \(\gamma\) is the slack factor and if \(\gamma \rightarrow 0\) the
strong formulation is recovered. We treat the result of this
optimization as the ``gold standard'' since it produces the best results
but needs access to prior samples and to a large amount of non-amortized
compute in the form of forward and adjoint PDE solves.

To assess the efficacy of posterior inference between our amortized
methods and the non-amortized methods, we devised benchmarks inspired by
the prescriptions in \citep{donoho2023data} to accelerate the
incremental development of this class of algorithms. Firstly, we
evaluate the image reconstruction quality of the point estimate
generated by each method. Secondly, we conduct a qualitative review of
the uncertainty images they produce. Lastly, we quantitatively analyze
their uncertainty calibration using the same calibration test outlined
in Section~\ref{sec-calibration}. Due to the expensive nature of the
non-amortized methods, we are only able to compare results on a single
unseen observation, but we expect these results to generalize to other
observations.

\subsection{Benchmark I: Comparing reconstruction
quality}\label{benchmark-i-comparing-reconstruction-quality}

\begin{figure}[H]

\begin{minipage}{0.25\linewidth}

\centering{

\includegraphics[width=1\textwidth,height=\textheight]{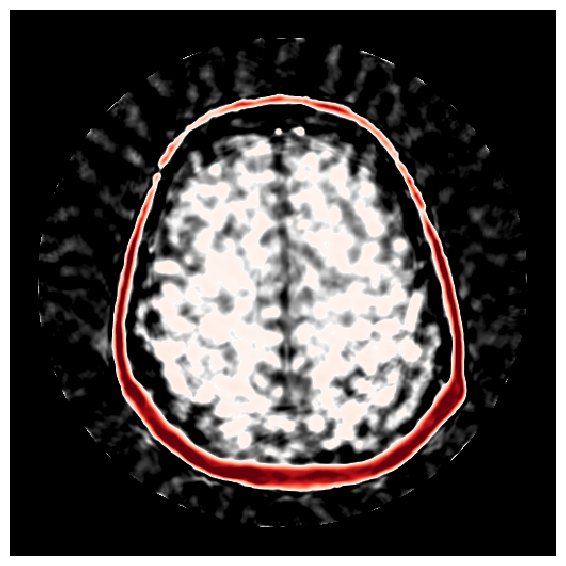}

}

\subcaption{\label{fig-mean-gaush}Mean-field}

\end{minipage}%
\begin{minipage}{0.25\linewidth}

\centering{

\includegraphics[width=1\textwidth,height=\textheight]{./figs/indx=24_num_post_samples=512_cm_iter_4.png}

}

\subcaption{\label{fig-mean-iter4-comp}Our amortized}

\end{minipage}%
\begin{minipage}{0.25\linewidth}

\centering{

\includegraphics[width=1\textwidth,height=\textheight]{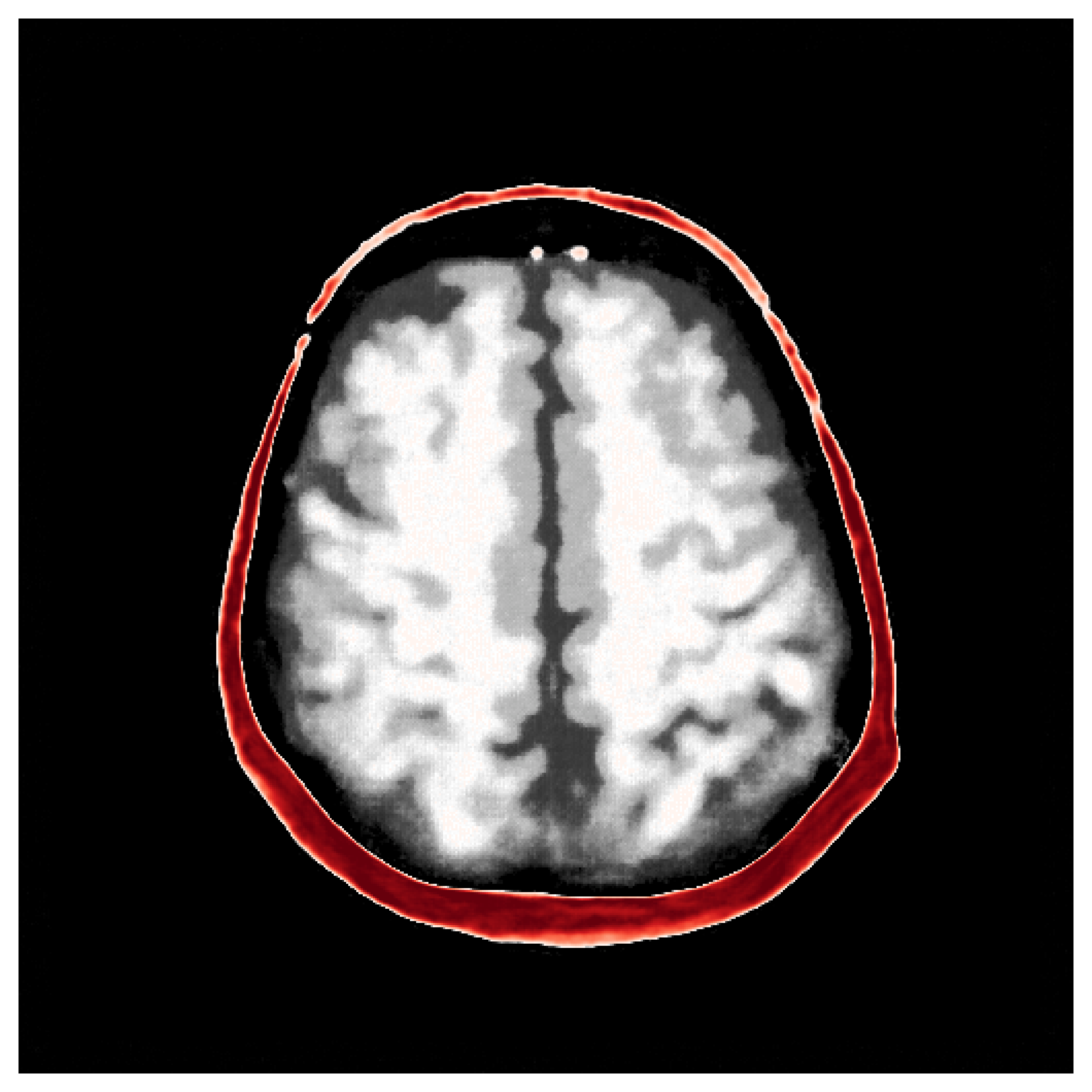}

}

\subcaption{\label{fig-mean-hint2}Our non-amortized}

\end{minipage}%
\begin{minipage}{0.25\linewidth}

\centering{

\includegraphics[width=1\textwidth,height=\textheight]{./figs/indx=24_gt.png}

}

\subcaption{\label{fig-gt-4}Ground truth}

\end{minipage}%

\caption{\label{fig-compare-mean}Reconstruction from benchmarked
methods. (a) Mean-field approximation \(\boldsymbol{\mu}_{m}\). (b) Our
amortized ASPIRE 4
\(\mathbb{E} \, p_{\widehat{\boldsymbol{\theta}}_4}\). (c) Our
non-amortized gold standard \(\mathbb{E} \, p_{\widehat \phi}\). (d)
Ground truth \(\mathbf{x}^{\ast}\). As expected, the non-amortized
method shows the highest quality, while our method shows similar quality
albeit missing some details on the lower right.}

\end{figure}

Our first comparison evaluates the posterior means from each method
against the ground truth. Note, the mean-field approximation method
yields estimates for posterior mean, \(\boldsymbol{\mu}_{m}\), directly,
whereas the normalizing flow-based methods estimate the posterior mean
from samples. From Figure~\ref{fig-compare-mean}, we observe that the
mean-field estimate contains strong artifacts in the soft tissue due to
overfitting the noise as it lacks prior knowledge. As expected, our
non-amortized normalizing flow method's posterior mean exhibits a
superior point estimate compared to our amortized approach. However,
this improvement comes with significantly higher computational costs
since this result requires \(800\) online evaluations of the forward
operator and its adjoint when solving the optimization in Equation
\eqref{eq-train-composition-weak}, compared to only four online
evaluations for amortized ASPIRE 4. This difference in computational
expense highlights the trade-off between efficiency and point-estimate
quality. This example shows we can achieve results that are close to
those by the non-amortized method while using only a small fraction of
the online compute.

\subsection{Comparing uncertainty
estimates}\label{comparing-uncertainty-estimates}

\begin{figure}

\begin{minipage}{0.33\linewidth}

\centering{

\includegraphics{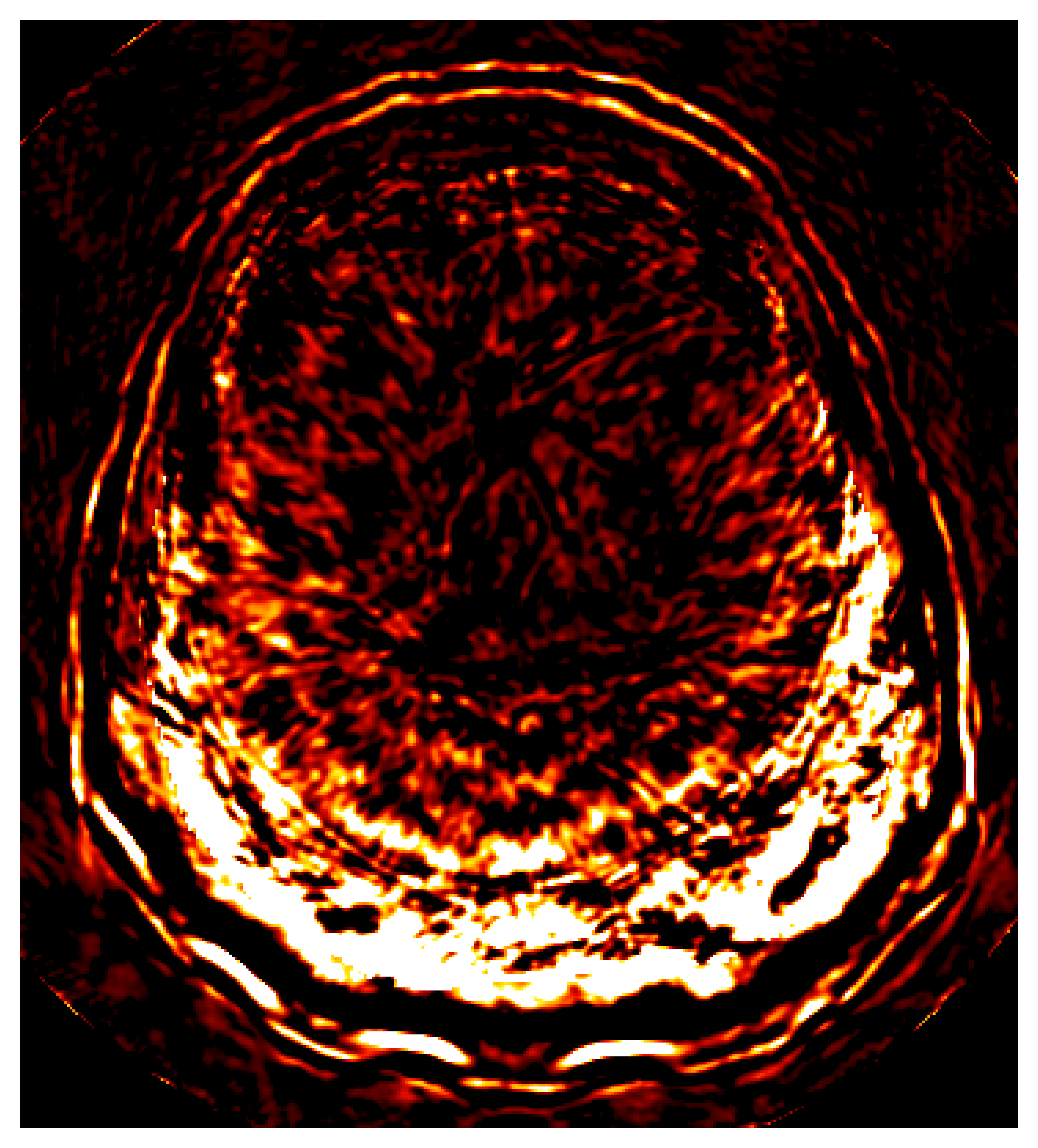}

}

\subcaption{\label{fig-var-gaush}Mean-field \(\boldsymbol{\sigma}_m\)}

\end{minipage}%
\begin{minipage}{0.33\linewidth}

\centering{

\includegraphics{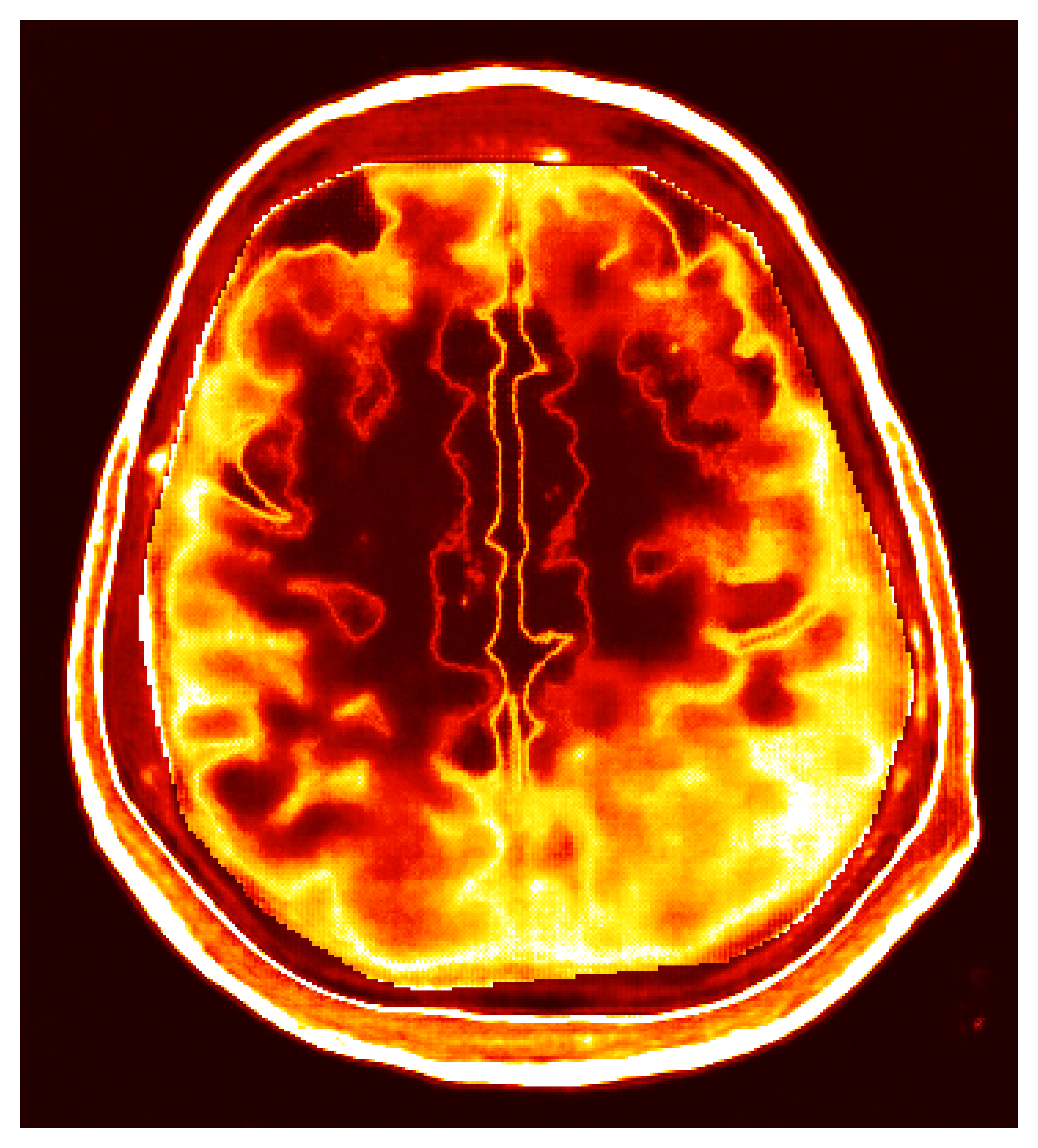}

}

\subcaption{\label{fig-var-iter4}\(\sqrt{\mathbb{V}}\, \, p_{\widehat{\boldsymbol{\theta}}_4}\)}

\end{minipage}%
\begin{minipage}{0.33\linewidth}

\centering{

\includegraphics{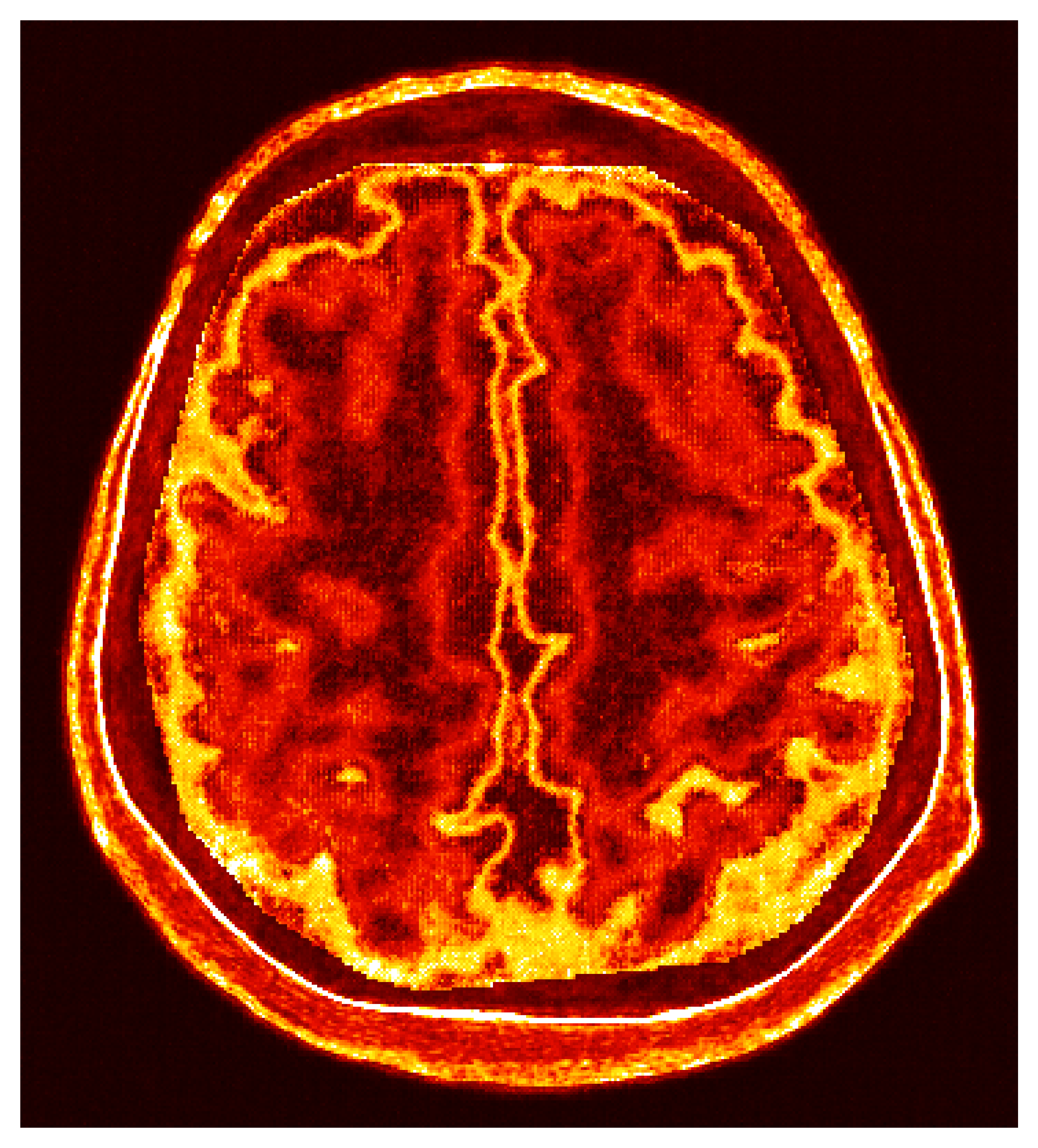}

}

\subcaption{\label{fig-var-non-amort}\(\sqrt{\mathbb{V}} \, \, p_{\widehat \phi}\)}

\end{minipage}%
\newline
\begin{minipage}{0.33\linewidth}

\centering{

\includegraphics{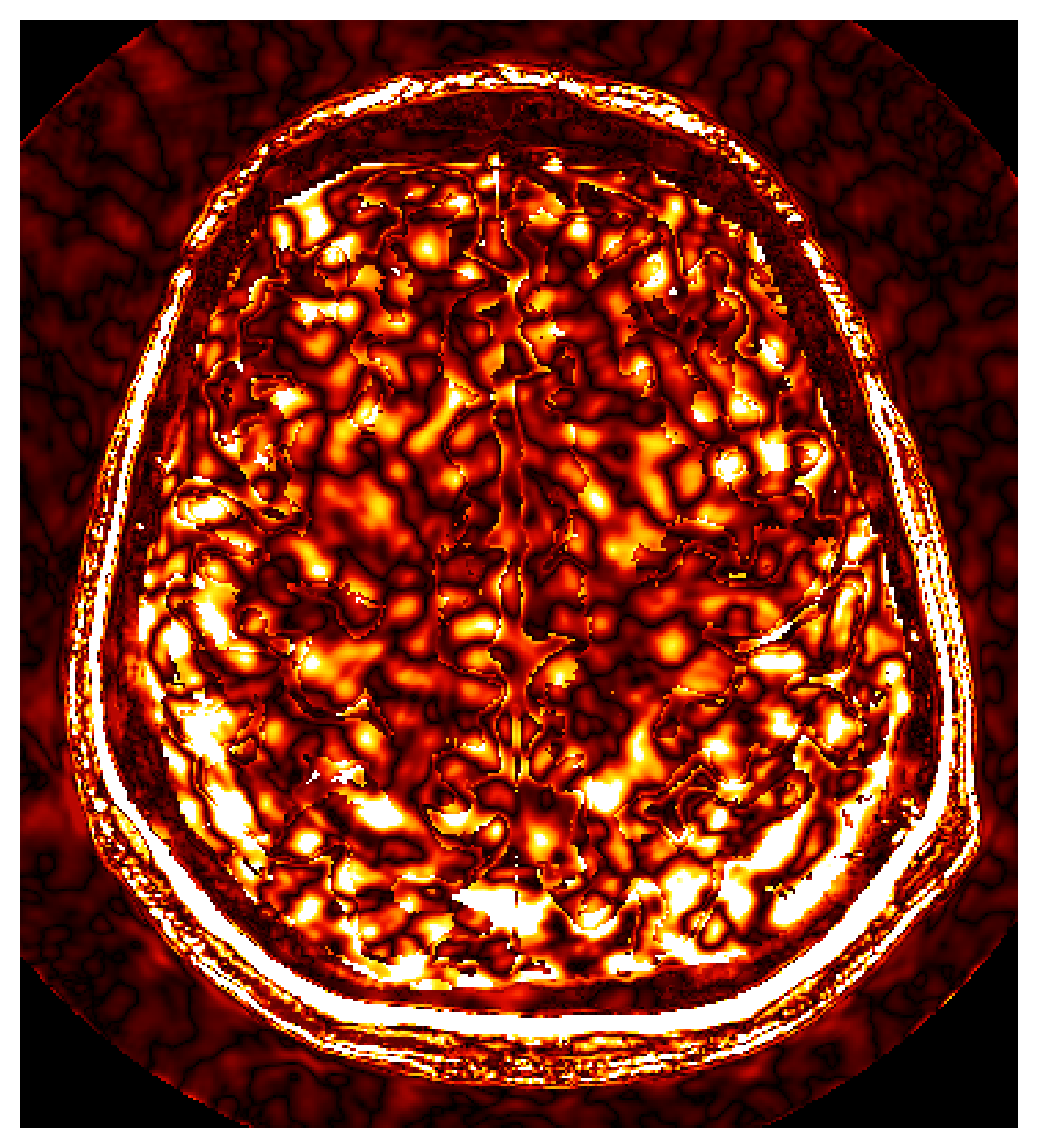}

}

\subcaption{\label{fig-error-guash}Error
\(|\mathbf{x}^{\ast} - \boldsymbol{\mu}_m|\)}

\end{minipage}%
\begin{minipage}{0.33\linewidth}

\centering{

\includegraphics{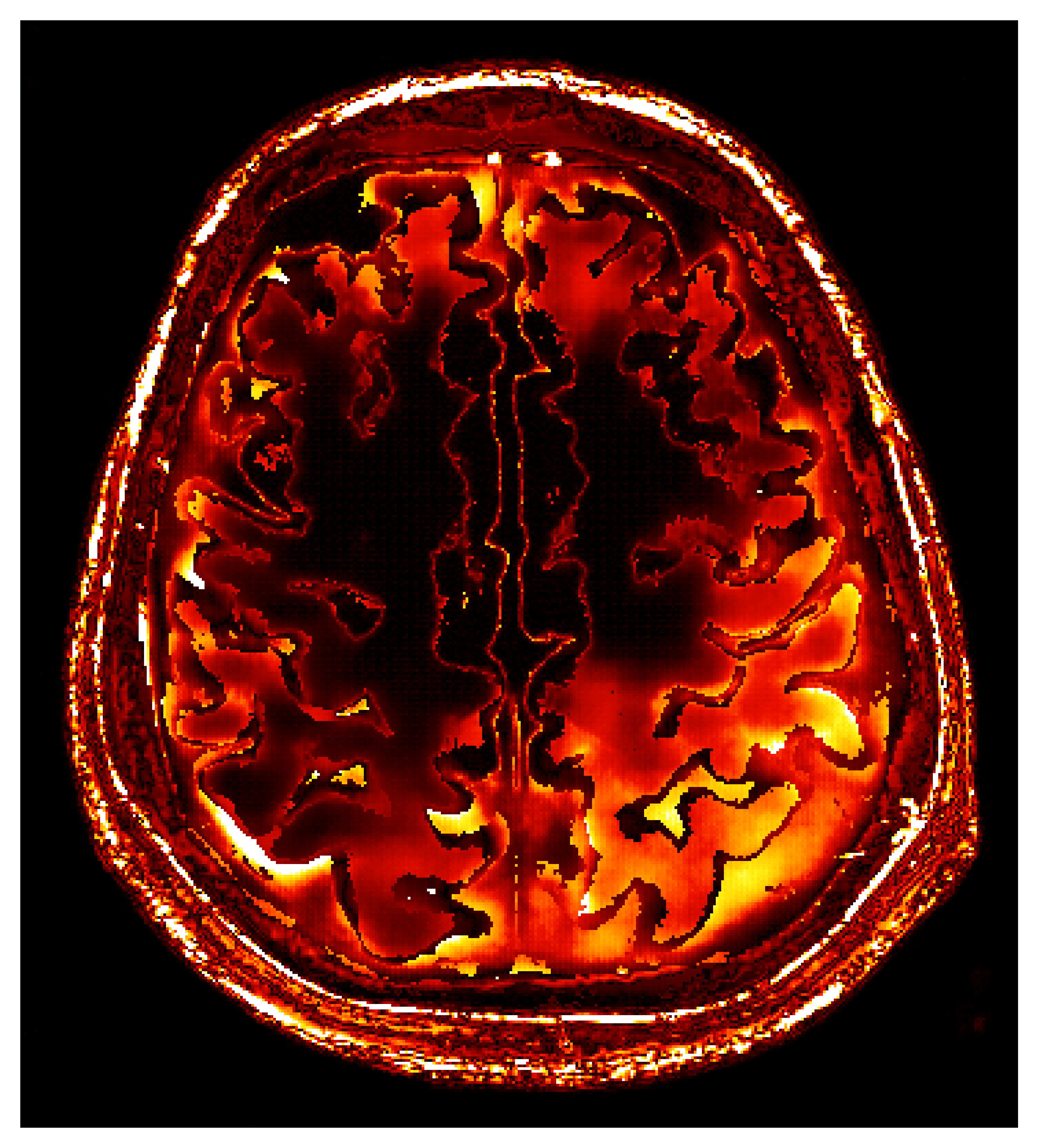}

}

\subcaption{\label{fig-error-iter4}Error
\(|\mathbf{x}^{\ast} - \mathbb{E} \, p_{\widehat{\boldsymbol{\theta}}_4}|\)}

\end{minipage}%
\begin{minipage}{0.33\linewidth}

\centering{

\includegraphics{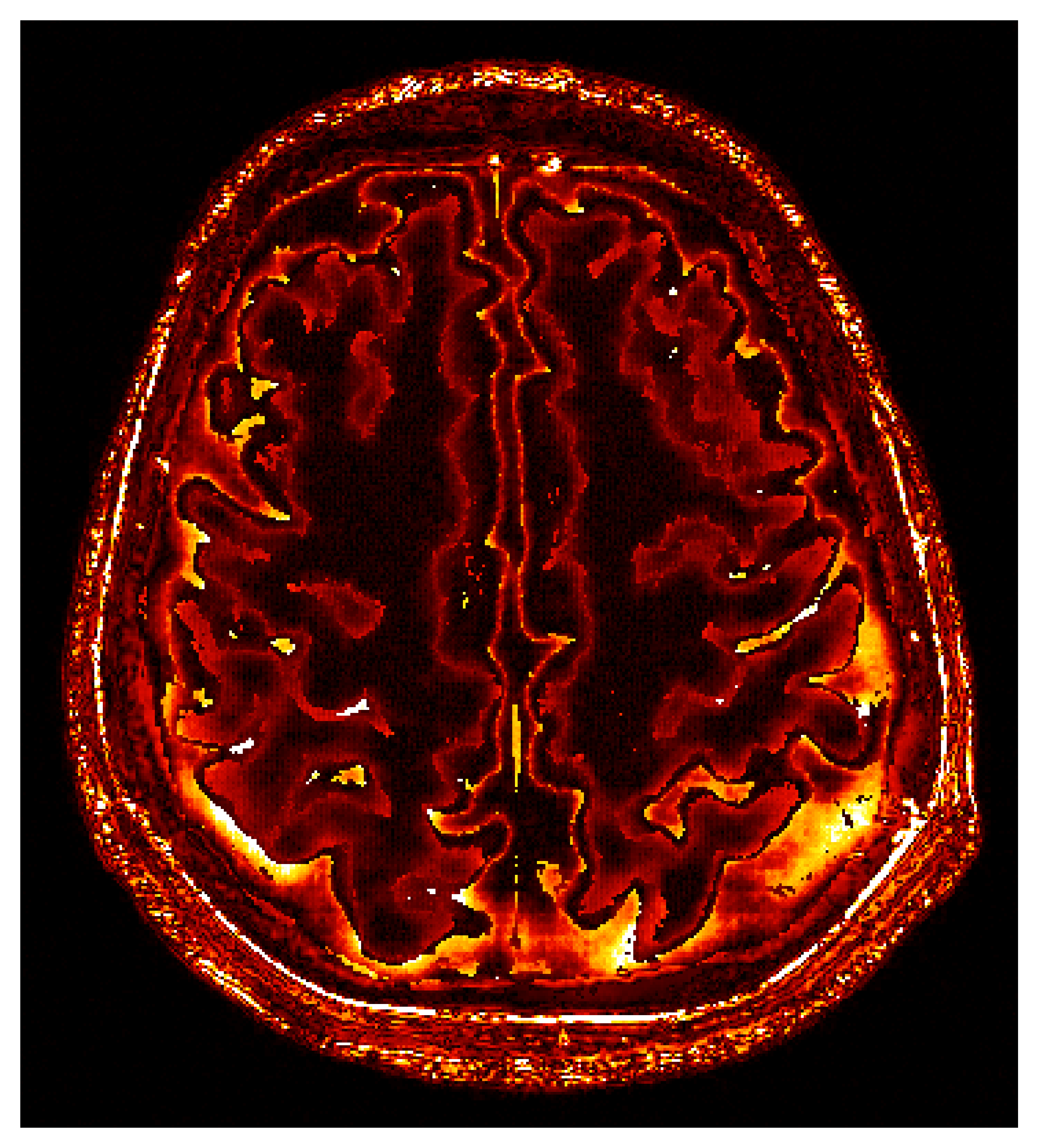}

}

\subcaption{\label{fig-error-non-amort}Error
\(|\mathbf{x}^{\ast} - \mathbb{E} \, p_{\widehat \phi}|\)}

\end{minipage}%

\caption{\label{fig-variance-compare}Comparing uncertainty of methods.
The first row shows the posterior standard deviation from: (a)
Non-amortized mean-field approximation sigma value. (b) Our amortized
method. (c) Our non-amortized gold-standard method. The second row shows
the corresponding errors. All plots have same colorbar from \(0\) to
\(50\mathrm{[m/s]}\).}

\end{figure}

The posterior standard deviation of the methods tell a similar story to
the posterior means. The structure of the standard deviation from the
mean-field approximation Figure~\ref{fig-var-gaush} comes mainly from
the physics of the problem therefore correctly concentrates in the lower
parts of the parameters where high-contrast has created complicated
wavefield reverberations, but the method has failed to warn of errors
due to noise artifacts throughout the reconstruction shown in
Figure~\ref{fig-error-guash}.

\begin{figure}

\centering{

\includegraphics[width=0.7\textwidth,height=\textheight]{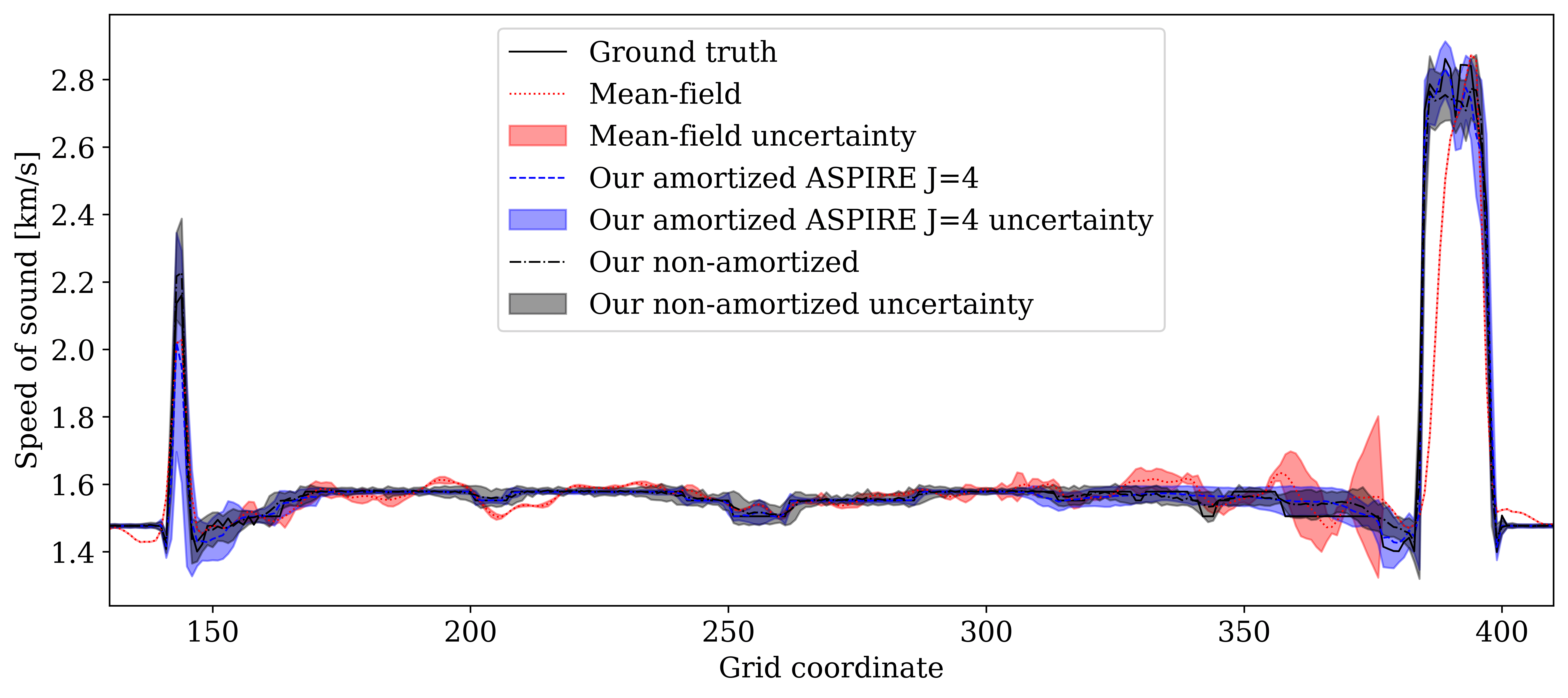}

\includegraphics[width=0.7\textwidth,height=\textheight]{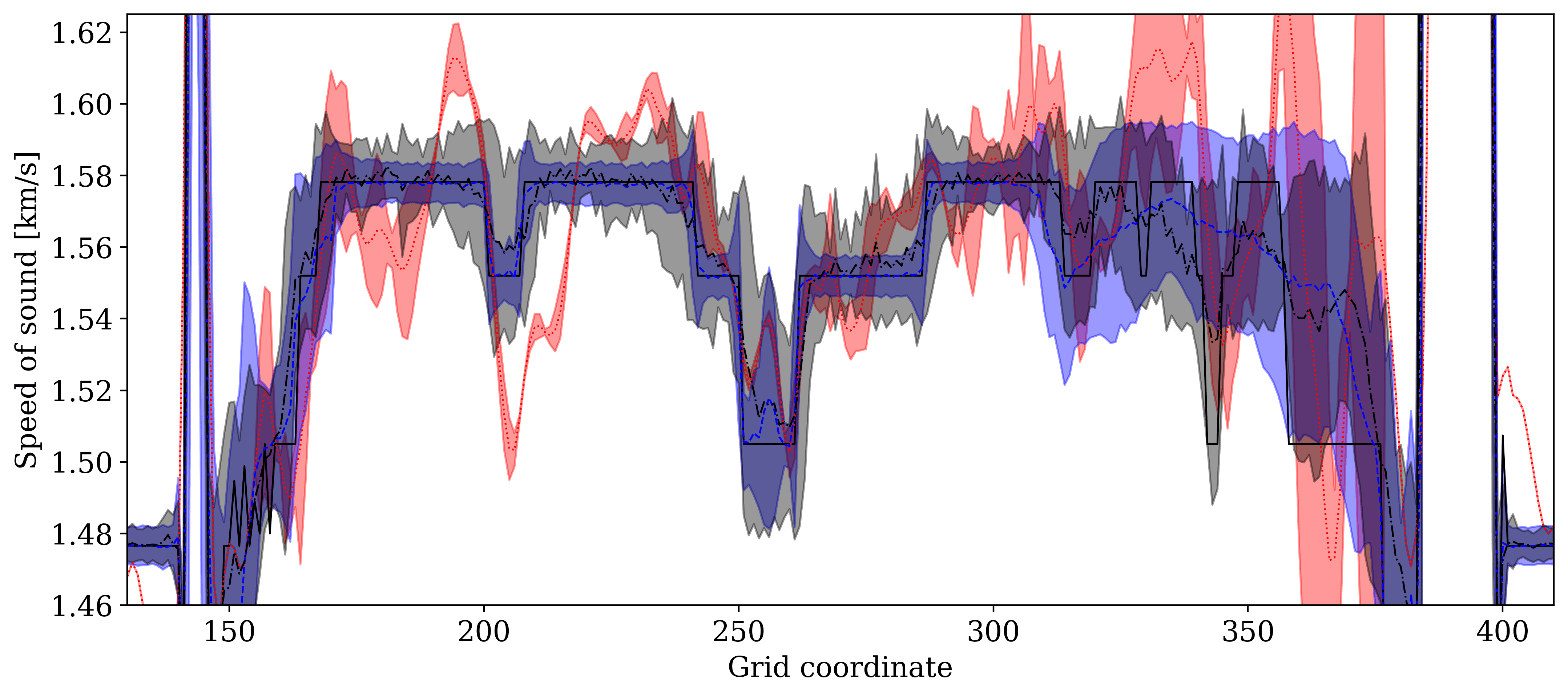}

}

\caption{\label{fig-traces}Comparison for a single trace. (Top figure)
Estimated parameters juxtaposed with the ground truth. (Bottom figure)
Same plot with zoomed vertical axis. While our posterior estimates have
relatively high error in the area with coordinates \(300\) to \(350\)
the uncertainty increases, suggesting the uncertainty is well
calibrated.}

\end{figure}

To perform a close inspection on the uncertainty of the VI methods, we
take a single trace through the posterior means (the diagonal trace
going from the top left to the bottom right). Figure~\ref{fig-traces}
shows that the mean-field method has large errors compared to the ground
truth and that its uncertainty band does not contain the ground truth.
Here we have chosen a \(2\sigma\) band around the mean. On the other
hand, both ASPIRE and the non-amortized method produce high-fidelity
estimates of the ground truth. Furthermore, when our methods have high
error, the uncertainty bands expand such that they contain the ground
truth with high fidelity.

\subsection{Benchmark II: Comparing uncertainty
calibration}\label{benchmark-ii-comparing-uncertainty-calibration}

\begin{figure}[H]

\centering{

\includegraphics[width=0.5\textwidth,height=\textheight]{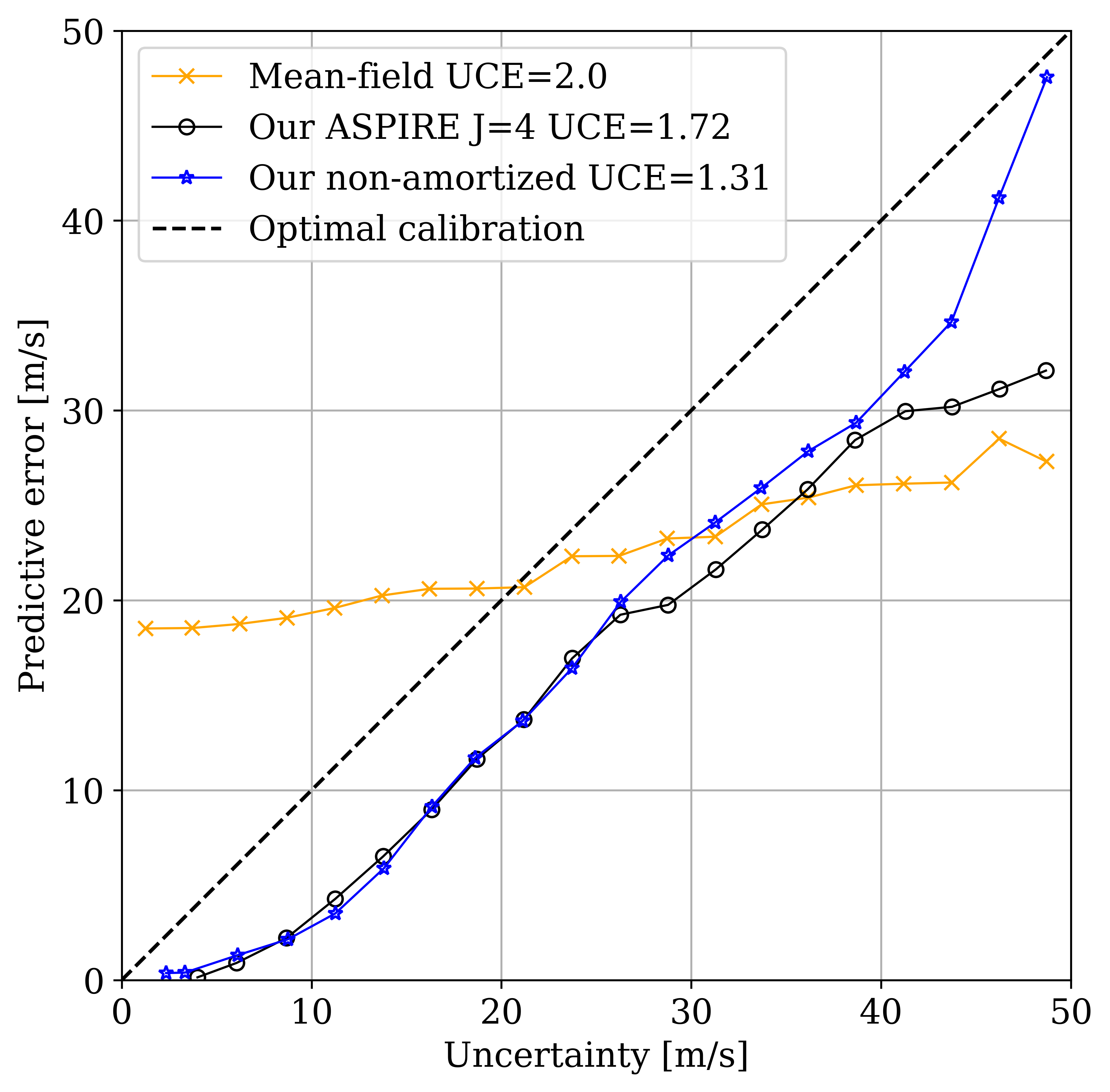}

}

\caption{\label{fig-calibration-compare-all}The mean-field approximation
shows poor calibration and the gold standard non-amortized method has
the best calibration as expected. Our amortized method is close to the
gold standard while using a small fraction of the compute cost.}

\end{figure}

Following the method in Section~\ref{sec-calibration}, we compare the
calibration curves of the three VI methods under consideration. Since
the two non-amortized methods are compute intensive, we are only able to
compute the calibration curve for a single test example
Figure~\ref{fig-compare-mean}. This contrasts with the more extensive
evaluation carried out in Figure~\ref{fig-improvement-uq}, which
encompasses a range of test cases thanks to the cheap online cost of our
method. The calibration of the three methods is shown in
Figure~\ref{fig-calibration-compare-all}. We observe that the mean-field
method shows poor calibration while our method achieves better
calibration that is close to the one of the non-amortized method. This
final observation is significant since it aligns with our thesis that we
can achieve approximation quality similar to an expensive non-amortized
method at a fraction of the cost. This finding highlights its
computational efficiency, making it a compelling choice in scenarios
where time-to-solution is a limiting factor.

\subsection{Benchmark III: Computational cost}\label{sec-computation}

The primary computational cost in the above methods lies in executing
the physics-based forward operator \(\mathcal{F}\) and its gradient,
especially in wave-based imaging where \(\mathcal{F}\) requires solving
PDEs. This step is much more demanding than posterior sampling, which
involves just a single neural network pass for normalizing flows. Thus,
we measure computational costs in terms of PDE solves.

\subsubsection{Offline phase:}\label{offline-phase}

In amortized VI, the bulk of computational expenses occur during the
offline phase. Similar to other simulation-based inference methods,
synthetic observations are generated by evaluating the forward operator.
Our method also requires the computation of the gradient for each
training sample, which equates to two more forward operators.
Additionally, each refinement requires recalculating the gradient.
However, a few iterations (3-4) are generally sufficient for
satisfactory results. Although the initial training phase is
resource-intensive, the amortized model, once trained, becomes
cost-effective with repeated use across various datasets.

\subsubsection{Online phase:}\label{online-phase}

The main cost during the online phase is also PDE solves. Each
refinement iteration, requiring gradient calculation, incurs a cost of
\(2\) PDE solves Section~\ref{sec-jacobian}. Compared to earlier
Bayesian methods that required \(1000 - 10000\) online PDE solves
\citep{zhao2022bayesian}, our method significantly reduces this to less
than \(10\) PDE online solves, nearing real-time imaging. In medical
applications where timely results are crucial \citep{bauer2013real},
fast online inference is essential.

\subsubsection{Compute break-even:}\label{compute-break-even}

Despite the high cost of the offline phase, our method becomes
cost-effective after a certain number of evaluations. The break-even
point can be estimated from Table~\ref{tbl-cost}. For instance, for
TUCT, compared to a mean-field solution requiring \(600\) PDE solves,
our method---with \(\mathbf{N}=1000\) and \(J=4\)---incurs \(9000\)
offline and \(8\) online PDE solves for a total of \(9008\) PDE solves.
Thus becomes cost-effective after about \(15\) test cases, not
accounting for improved estimates and uncertainty.

When juxtaposed with our proposed gold-standard non-amortized method,
our amortized approach breaks-even more rapidly. For the non-amortized
method, we use a pretrained ASPIRE 1 network, which requires
\(1000 + 2\times 1000\times 1=3000\) offline PDEs solves, followed by
\(2\times 1=2\) online solves. The optimization of
\eqref{eq-train-composition-weak} uses \(L=400\) outer loop iterations
leading to online \(800\) PDE solves. The total cost is \(3802\) PDE
solves, which means the amortized method pays itself back when used on
\(\frac{9008}{3802} \approx 3\) test cases.

\begin{longtable}[]{@{}
  >{\raggedright\arraybackslash}p{(\columnwidth - 4\tabcolsep) * \real{0.4624}}
  >{\raggedright\arraybackslash}p{(\columnwidth - 4\tabcolsep) * \real{0.3118}}
  >{\raggedright\arraybackslash}p{(\columnwidth - 4\tabcolsep) * \real{0.2258}}@{}}
\caption{Costs measured by evaluations of forward/adjoint
operator.}\label{tbl-cost}\tabularnewline
\toprule\noalign{}
\begin{minipage}[b]{\linewidth}\raggedright
Method
\end{minipage} & \begin{minipage}[b]{\linewidth}\raggedright
Offline cost
\end{minipage} & \begin{minipage}[b]{\linewidth}\raggedright
Online cost
\end{minipage} \\
\midrule\noalign{}
\endfirsthead
\toprule\noalign{}
\begin{minipage}[b]{\linewidth}\raggedright
Method
\end{minipage} & \begin{minipage}[b]{\linewidth}\raggedright
Offline cost
\end{minipage} & \begin{minipage}[b]{\linewidth}\raggedright
Online cost
\end{minipage} \\
\midrule\noalign{}
\endhead
\bottomrule\noalign{}
\endlastfoot
ASPIRE & \(N + 2 \times N \times J\) & \(2 \times J\) \\
Our non-amortized method & \(N + 2 \times N \times J\) &
\(2 \times J + 2 \times L\) \\
Mean-field approximation & \(\textrm{None}\) & \(2 \times L\) \\
\end{longtable}

\section{Related work}\label{related-work}

A precursor to our work, \citep{radev2020bayesflow} introduced the
concept of the summary network, which acts on the condition and is
optimized under the same objective as the normalizing flow. Our approach
extends this concept into what we term a ``physical summary network,''
where each refinement iteration enhances the summary statistic by
improving the fiducial quality. Although not iterative and for different
modalities \citep{denker2021conditional} demonstrated one of the first
uses of CNFs for medical imaging.

During the preparation of this manuscript, we identified closely related
research \citep{barbano2021quantifying}, which proposes a method for
solving Bayesian inverse problems that resembles loop-unrolling
augmented with Bayesian network layers. We acknowledged their
contributions. However, we note key differences:
\citep{barbano2021quantifying} employ Bayesian networks that model
distributions on the network weights, which can impose restrictive
assumptions on the distribution families that can be learned akin to
mean-field approximations. In contrast, our use of CNFs aims to directly
learn the Bayesian posterior, and theoretically, as universal
approximators \citep{teshima2020coupling, draxler2023convergence}, offer
greater flexibility. We believe our method generalizes to other
conditional density estimators whereas the work in
\citep{barbano2021quantifying} is specific to Bayesian networks.

Our work also shares similarities with DEEPGEM by
\citep{gao2021deepgem}, which utilizes Expectation Maximization to
solving inverse problems. Their process involves optimizing a
non-amortized normalizing flow to sample from a posterior based on
current nuisance parameter estimates, followed by MAP optimization.
Crucially, our method is different as it is amortized, eliminating
network retraining or costly MAP optimization at inference time.
Additionally, our method requires few online gradients (3-4), compared
to the numerous ones needed by DEEPGEM, thereby significantly reducing
compute.

\section{Future work}\label{future-work}

The superior posterior mean achieved by the non-amortized method
compared to our solution indicates that further information could be
extracted from \(\mathbf{y}^{\mathrm{obs}}\) through additional
refinement iterations. Although we proposed some heuristics, determining
the optimal number of refinement iterations to maximize performance
remains an area of research. Our technique is compatible with any
conditional density estimator. While we have utilized Normalizing Flows
in our implementation, the framework can easily adapt to other
conditional density estimators such as Variational Autoencoders (VAEs)
\citep{sohn2015learning}, GANs \citep{mirza2014conditional} diffusion
models \citep{song2020score}.

\begin{figure}[H]

\begin{minipage}{0.50\linewidth}

\centering{

\includegraphics[width=0.75\textwidth,height=\textheight]{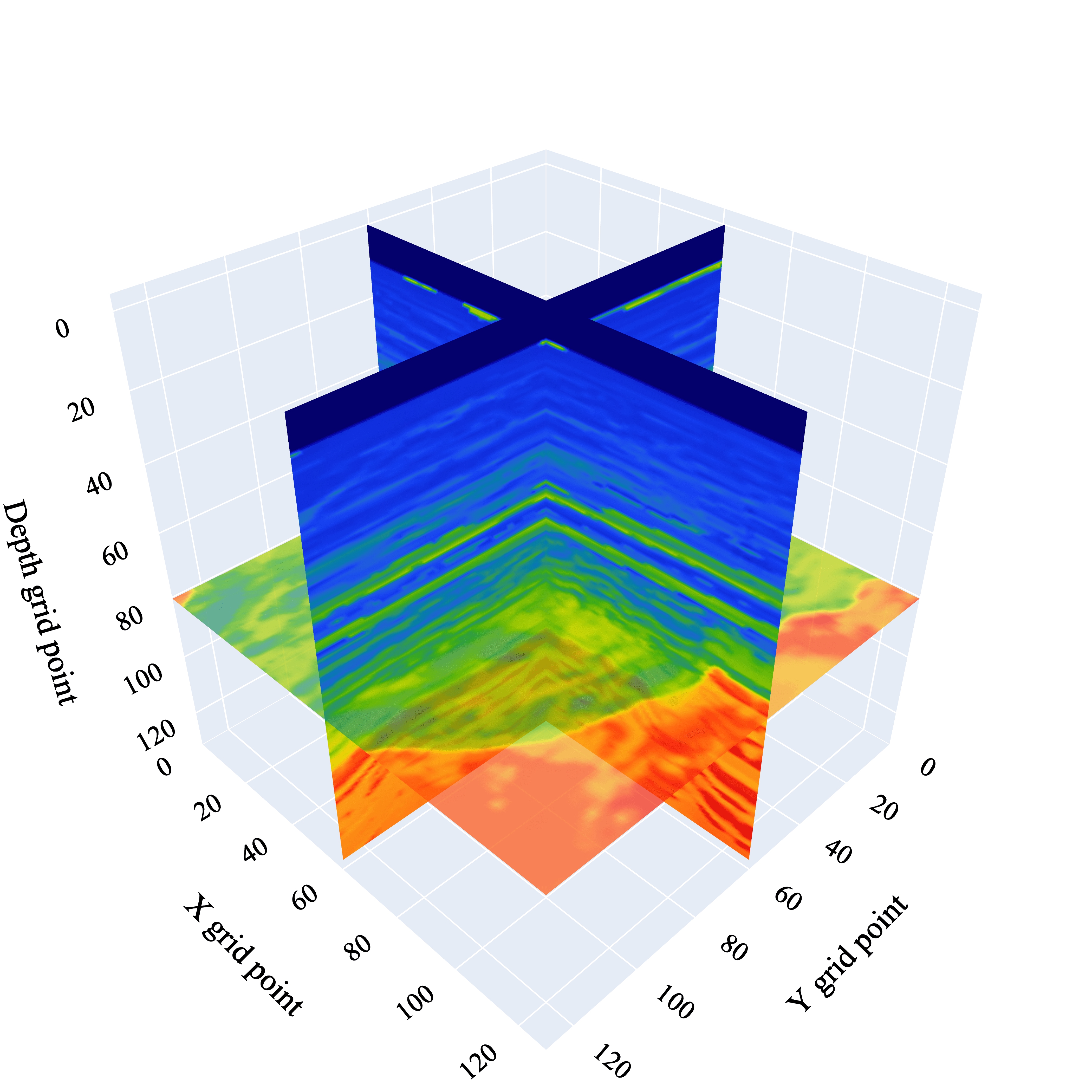}

}

\subcaption{\label{fig-3d-gt}Ground truth}

\end{minipage}%
\begin{minipage}{0.50\linewidth}

\centering{

\includegraphics[width=0.75\textwidth,height=\textheight]{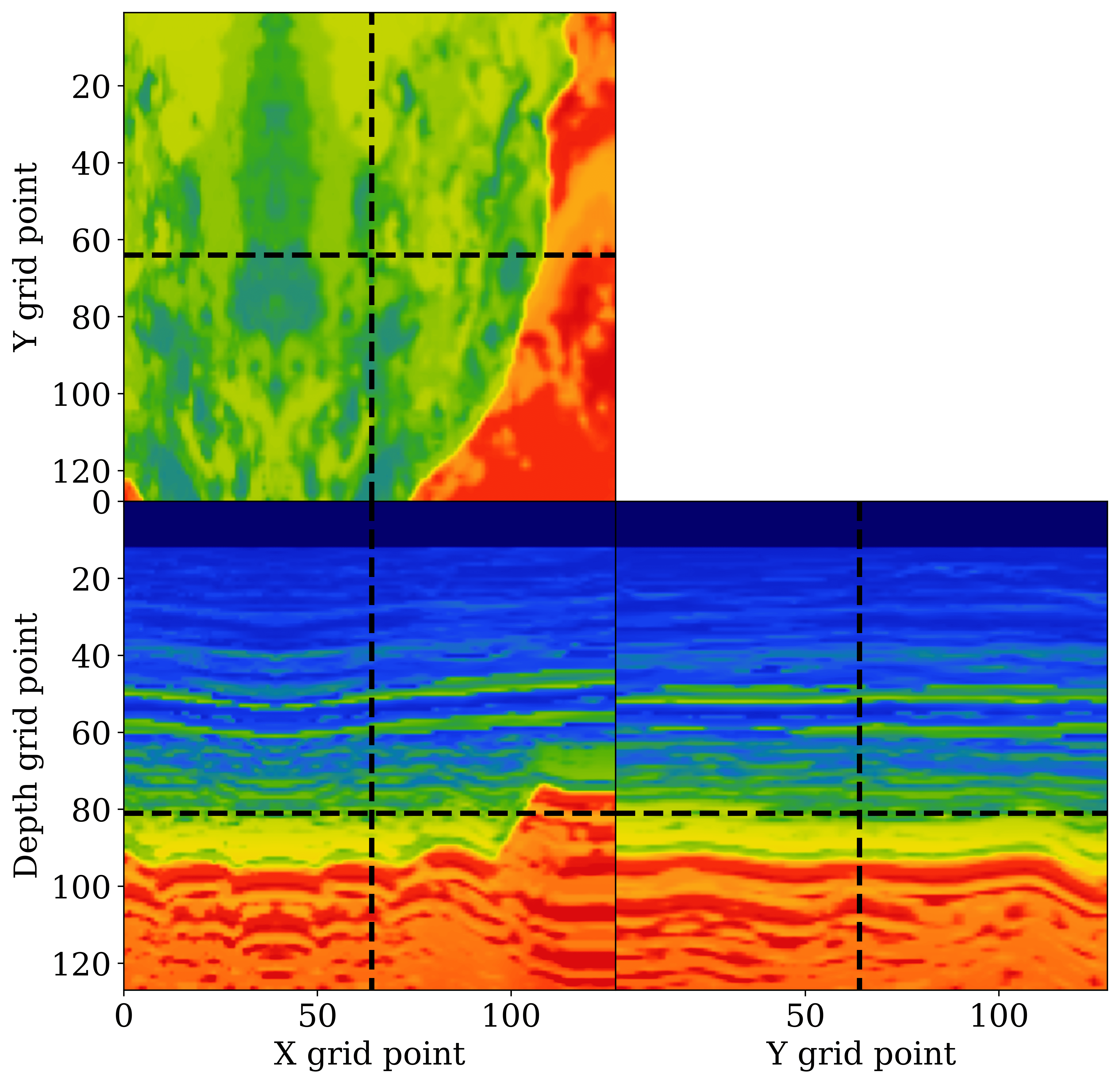}

}

\subcaption{\label{fig-3d-gt-slices}Ground truth slices}

\end{minipage}%
\newline
\begin{minipage}{0.50\linewidth}

\centering{

\includegraphics[width=0.75\textwidth,height=\textheight]{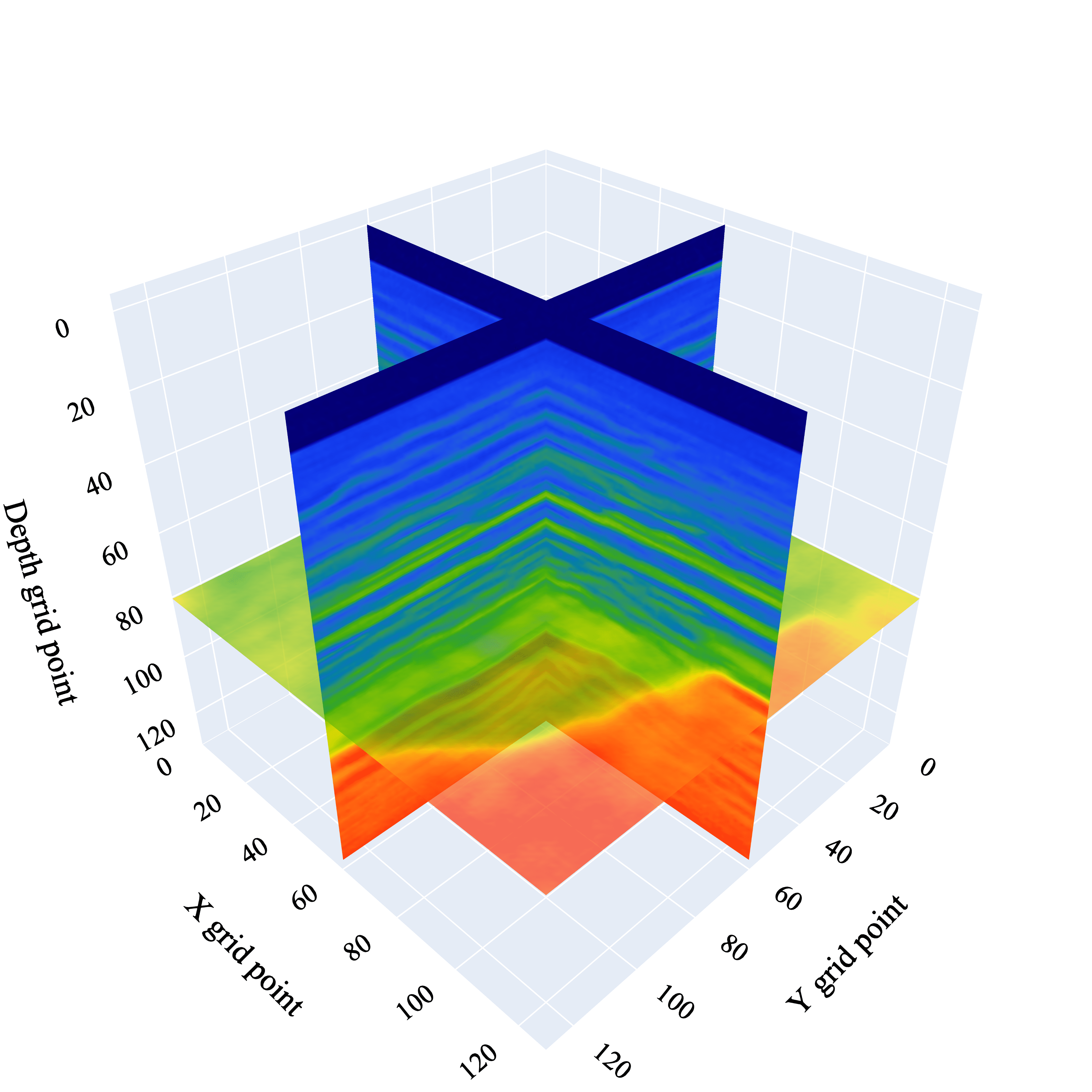}

}

\subcaption{\label{fig-3d-mean}Our posterior mean}

\end{minipage}%
\begin{minipage}{0.50\linewidth}

\centering{

\includegraphics[width=0.75\textwidth,height=\textheight]{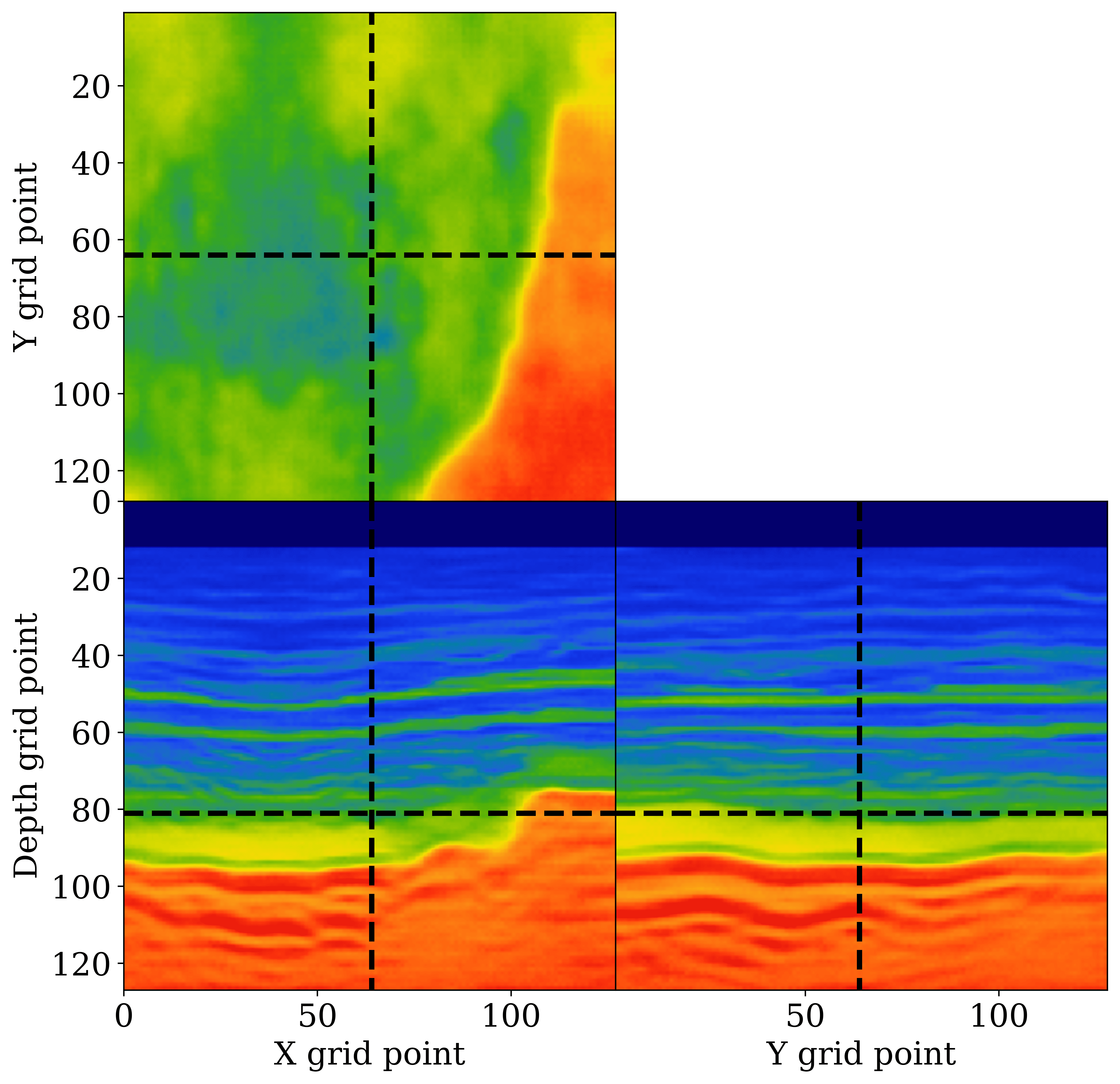}

}

\subcaption{\label{fig-3d-mean-slices}Our posterior mean slices}

\end{minipage}%
\newline
\begin{minipage}{0.50\linewidth}

\centering{

\includegraphics[width=0.75\textwidth,height=\textheight]{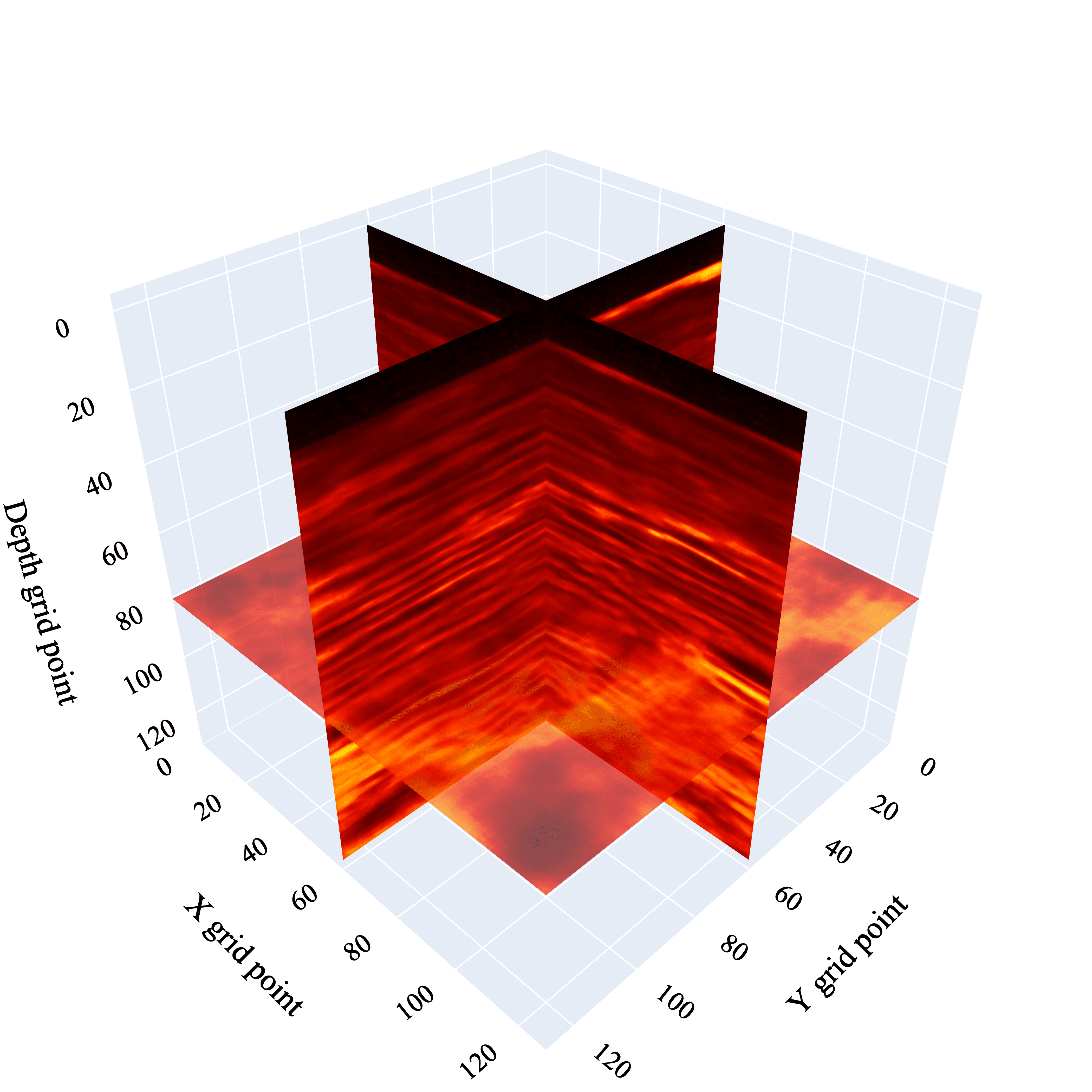}

}

\subcaption{\label{fig-3d-std}Our posterior deviation}

\end{minipage}%
\begin{minipage}{0.50\linewidth}

\centering{

\includegraphics[width=0.75\textwidth,height=\textheight]{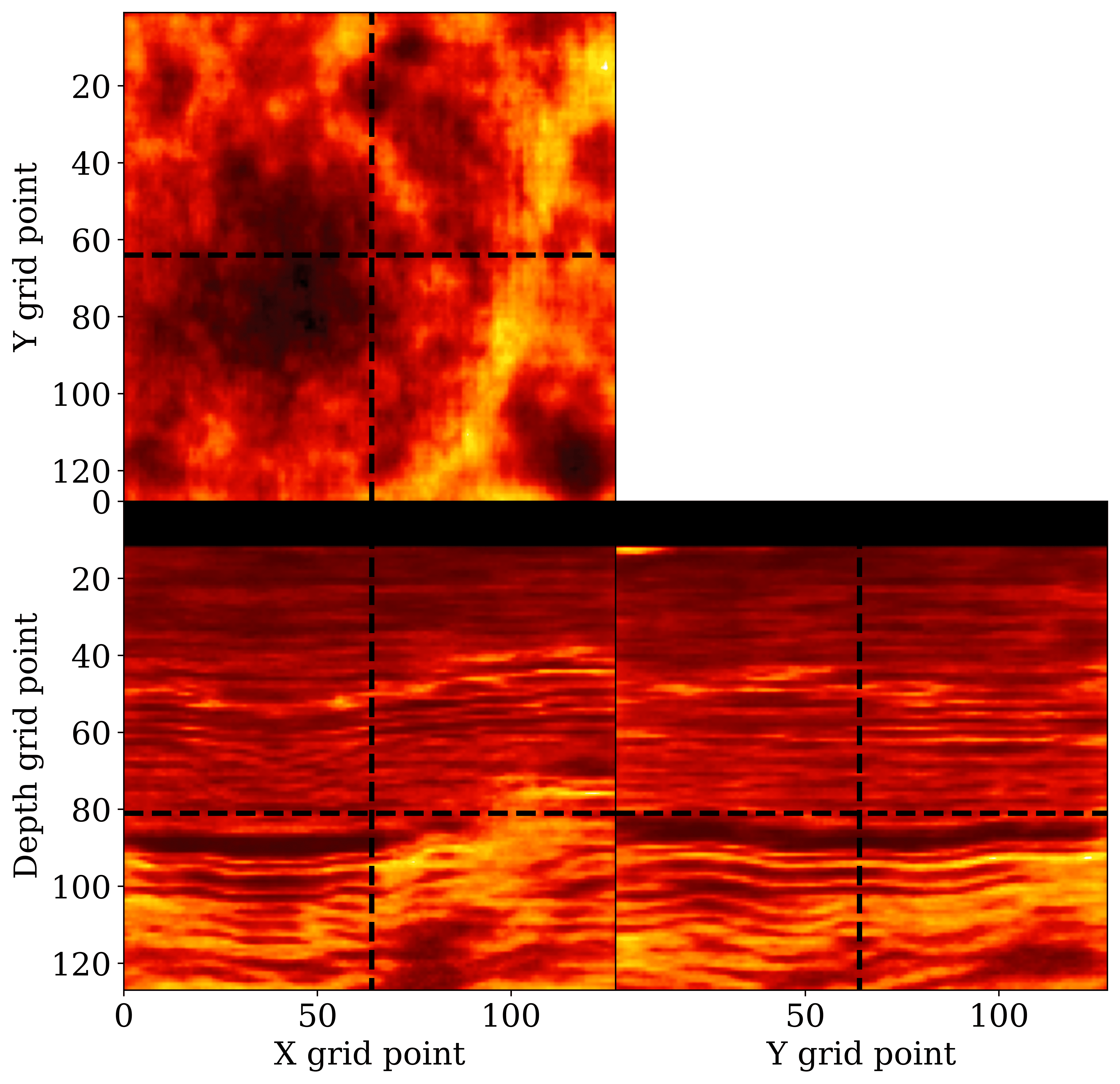}

}

\subcaption{\label{fig-3d-std-slices}Our posterior deviation slices}

\end{minipage}%

\caption{\label{fig-3d-compass}ASPIRE for 3D inverse problems. (a)
Ground truth 3D render. (b) Ground truth folded out slices. (c) Our
posterior mean 3D render. (d) Posterior mean slices. (e) Our posterior
deviation 3D render. (f) Posterior deviation slices. The ASPIRE 2 shows
physically viable probabilistic estimates for a
\(128\times128\times128\) FWI problem.}

\end{figure}

Due to out-of-plane effects of acoustic modeling, the TUCT problem is
best treated in 3D \citep{guasch2020full}. Although TUCT was
demonstrated here in 2D, ASPIRE is not limited to 2D problems.
Particularly, when empowered by memory-frugal normalizing flows
\citep{orozco2023invertiblenetworks} ASPIRE can achieve full volume
Bayesian inference for 3D inverse problems. For our TUCT example, the
limiting factor was the absence of a 3D training dataset but in seismic
imaging (a field that appreciates the importance of 3D modeling and
inference) we used the 3D Compass dataset \citep{jones2012building} and
share the results of solving 3D FWI in Figure~\ref{fig-3d-compass}.
Setup details are similar to the TUCT problem. A detailed study of the
3D capabilities of ASPIRE is being prepared for future work but here we
succinctly report that for a \(128 \times 128 \times128\) inference
problem, offline training took \(1\) day on a single GPU and that the
uncertainties shown in Figure~\ref{fig-3d-std} were pleasingly
correlated with structures that are known to be difficult to image
i.e.~structures that are: deeper, vertical or close to the edge.

\section{Conclusions}\label{conclusions}

We introduced a method that iteratively improves on approximations to
Bayesian posteriors in the context of inverse problems. Our method
brings together concepts from generative modeling, physics-hybrid
methods, and statistics. Practically our algorithm achieves higher
performance by iteratively extracting more information from the observed
data. The mathematical interpretation of our method is to make a
score-based summary statistic more informative by moving the fiducial
points closer to the maximum likelihood estimates. Our method forms an
interesting middle ground between amortized VI and non-amortized VI.
Importantly, the offline training phase makes it such that the online
costs are small rendering our approach suitable for applications that
demand fast online turn-around times. Our experiments demonstrate
improvements in estimated posteriors on a stylized example where the
posterior is known analytically. In a realistic medical transcranial
ultrasound imaging application, the online cost is many times cheaper
than non-amortized methods while demonstrating high-quality amortized
inference. We believe that this approach represents a step forward in
the field, offering a computationally efficient solution for Bayesian
inference in high-dimensional inverse problems with expensive to
evaluate forward operators.

\section{Availability of data and
materials}\label{availability-of-data-and-materials}

The scripts and datasets to reproduce the experiments are available on
the GitHub page
\href{https://github.com/slimgroup/ASPIRE.jl}{ASPIRE.jl}.

\section{Acknowledgements}\label{acknowledgements}

This research was carried out with the support of Georgia Research
Alliance and partners of the ML4Seismic Center. The authors gratefully
acknowledge the contribution of OpenAI's ChatGPT for refining sentence
structure and enhancing the overall readability of this manuscript.
After using this service, the authors reviewed and edited the content as
needed and take full responsibility for the content of the publication.

\appendix

\section{FASTMRI acoustic dataset creation}\label{sec-prior-generation}

\begin{figure}

\begin{minipage}{0.33\linewidth}
\includegraphics{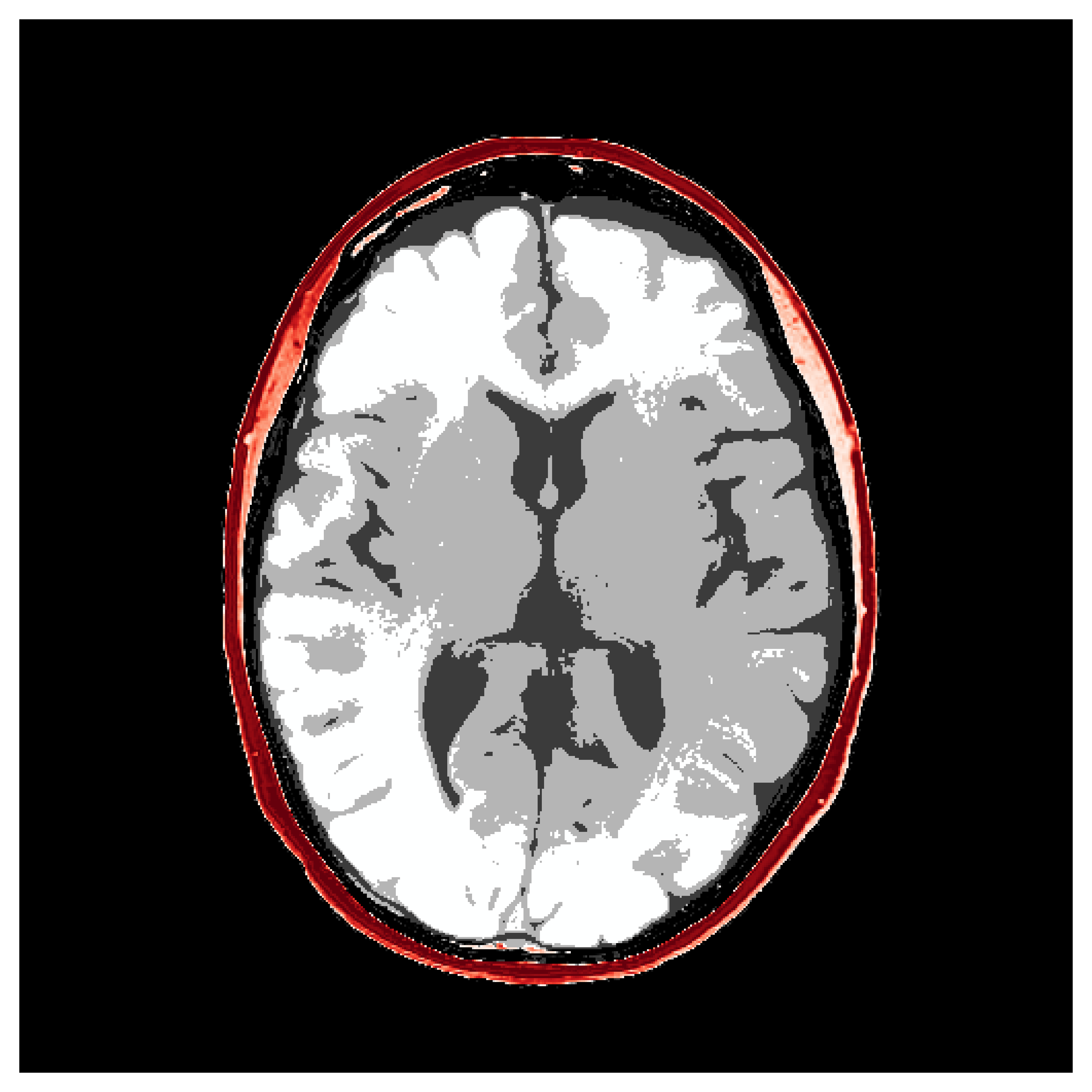}\end{minipage}%
\begin{minipage}{0.33\linewidth}
\includegraphics{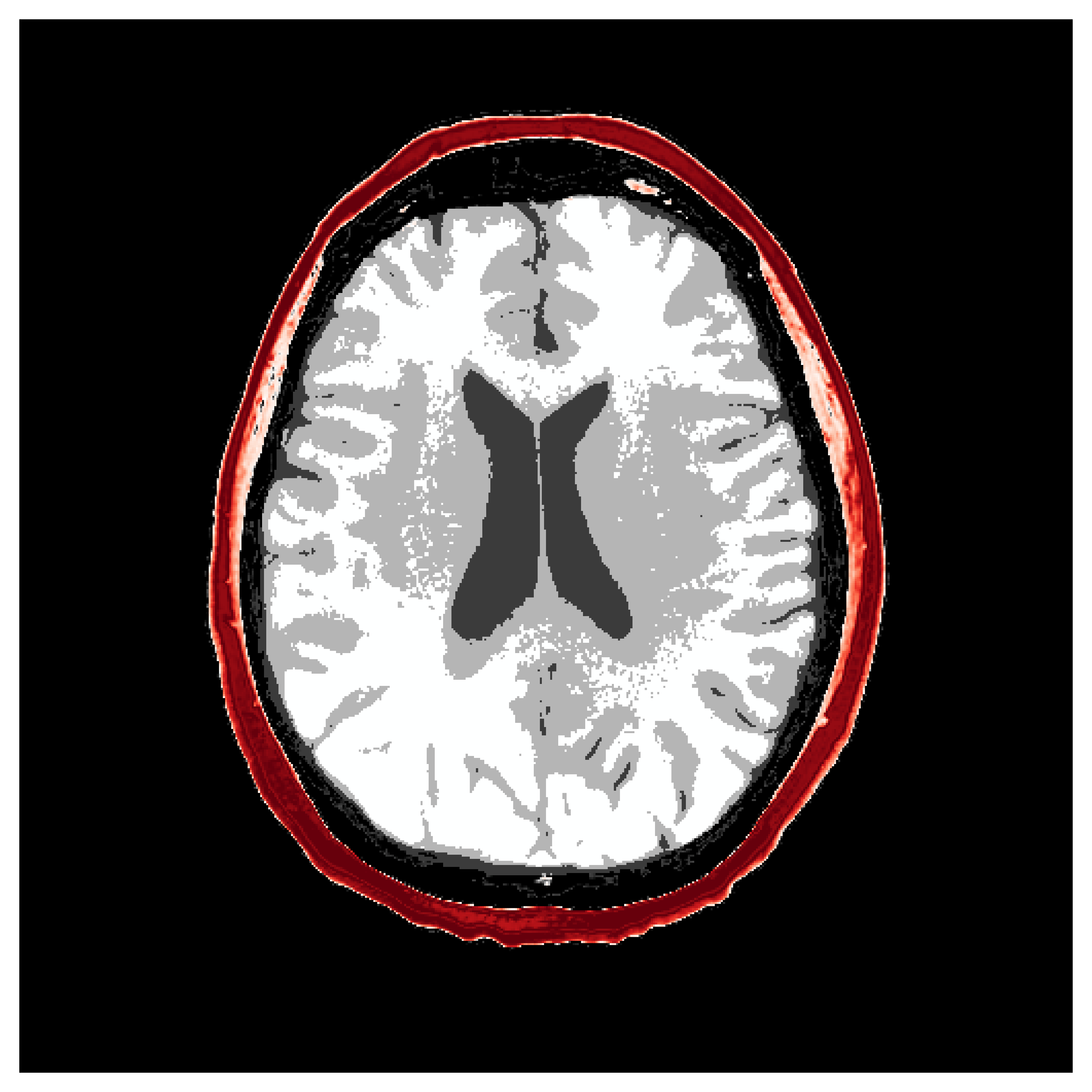}\end{minipage}%
\begin{minipage}{0.33\linewidth}
\includegraphics{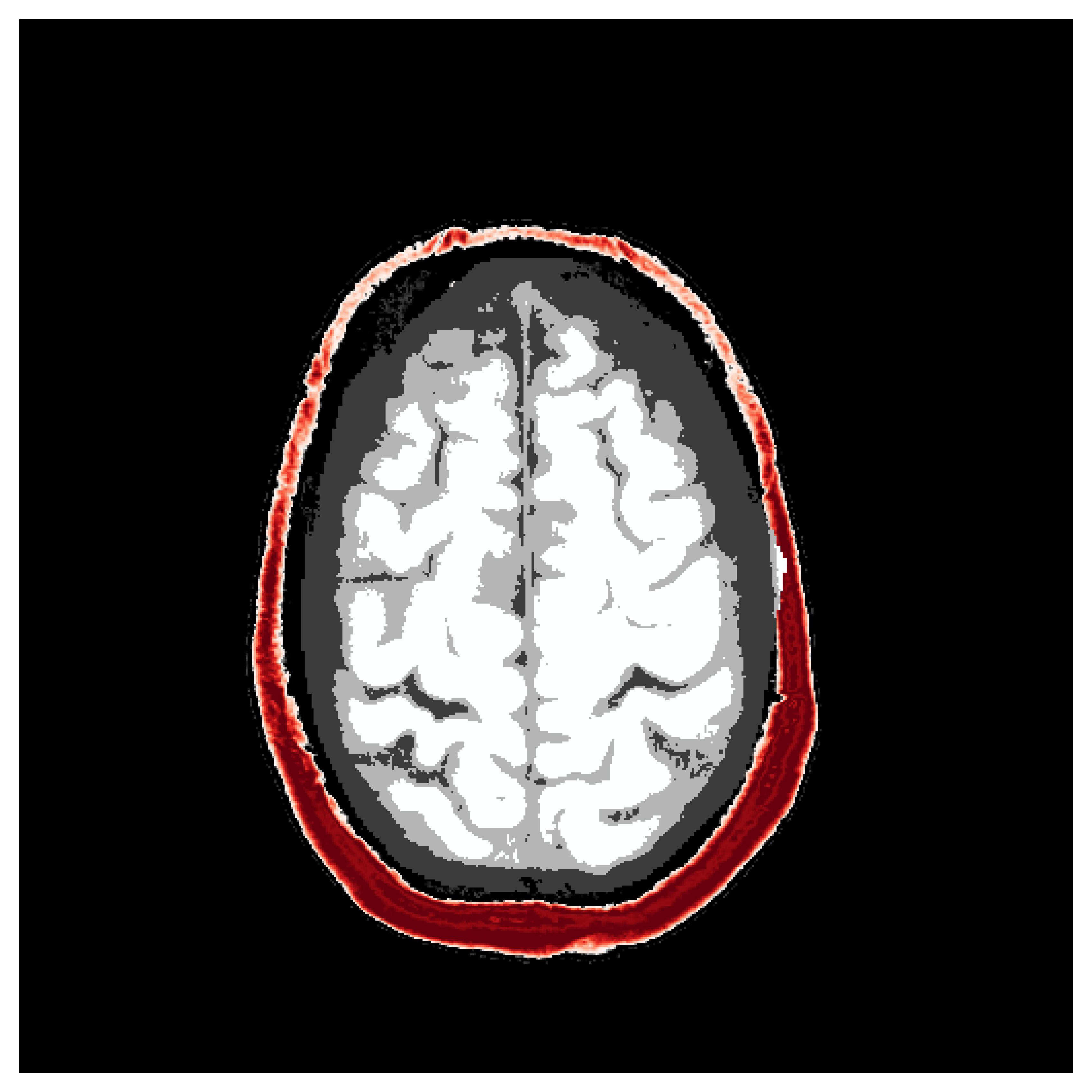}\end{minipage}%
\newline
\begin{minipage}{0.33\linewidth}
\includegraphics{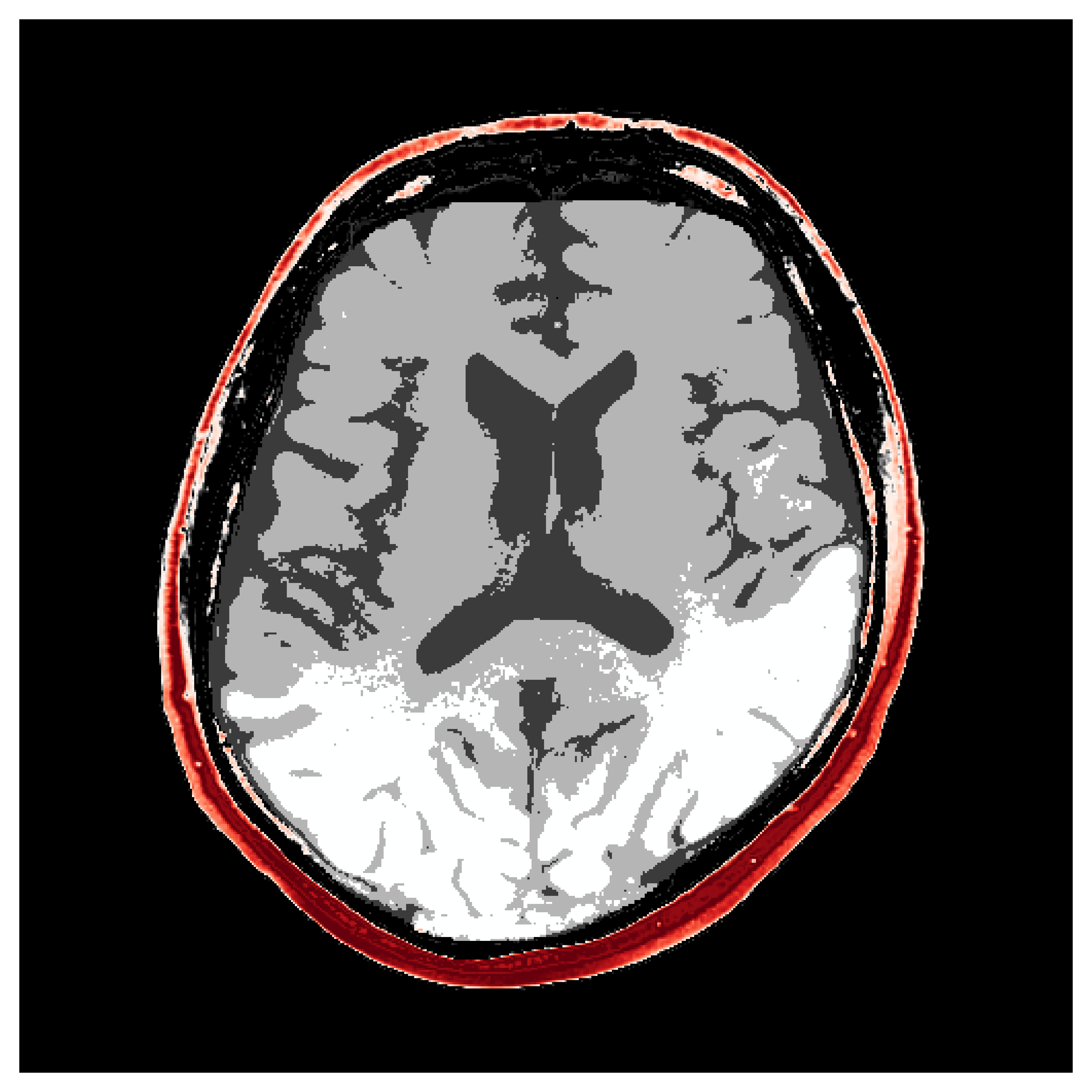}\end{minipage}%
\begin{minipage}{0.33\linewidth}
\includegraphics{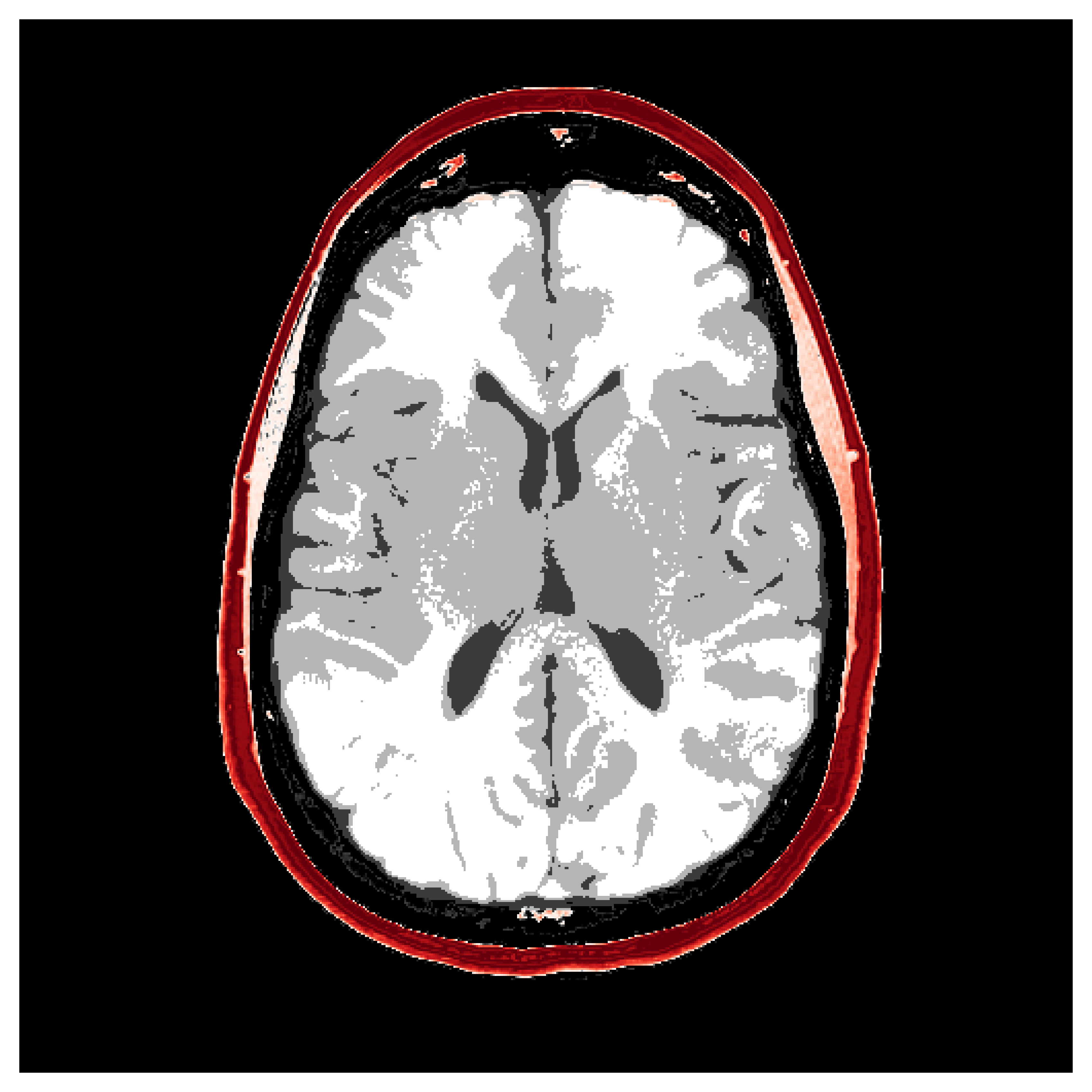}\end{minipage}%
\begin{minipage}{0.33\linewidth}
\includegraphics{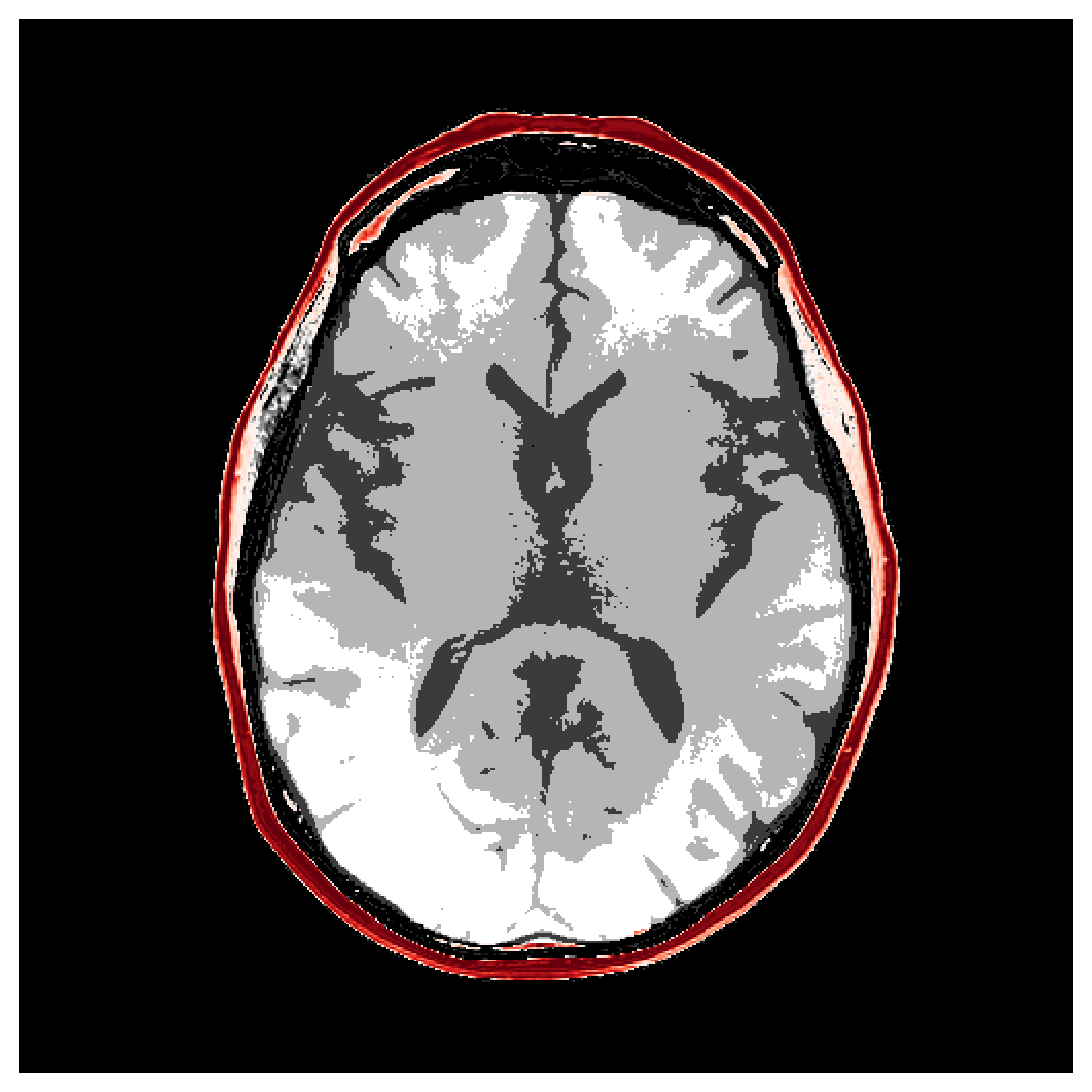}\end{minipage}%

\caption{\label{fig-train-dataset}Examples of training examples used to
train our method \(\mathbf{x}^{(n)} \sim p(\mathbf{x})\).}

\end{figure}

\begin{figure}

\begin{minipage}{0.50\linewidth}

\centering{

\includegraphics{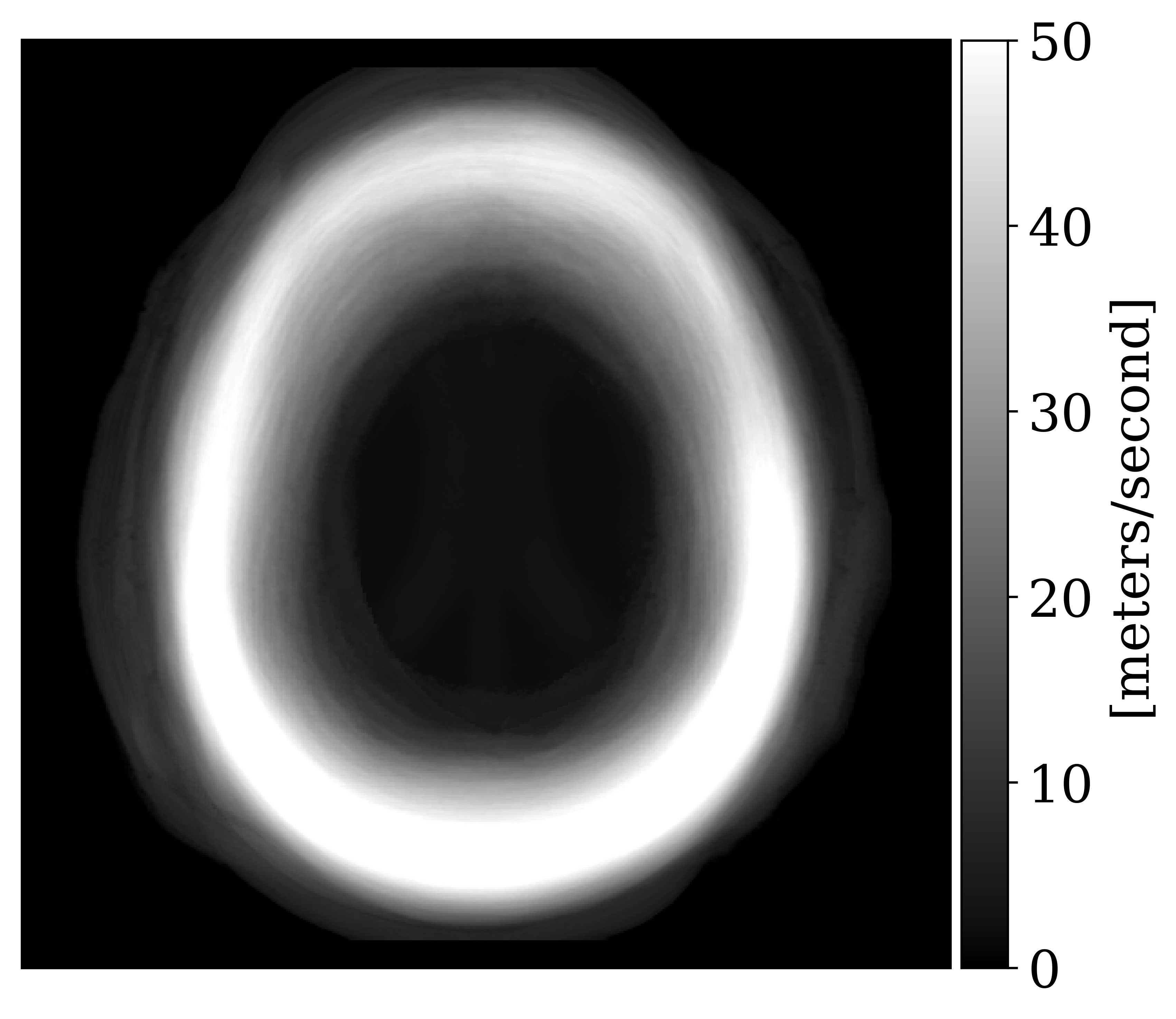}

}

\subcaption{\label{fig-var-iter1}\(\sqrt{\mathbb{V}} \, \, p(\mathbf{x})\)}

\end{minipage}%
\begin{minipage}{0.50\linewidth}

\centering{

\includegraphics{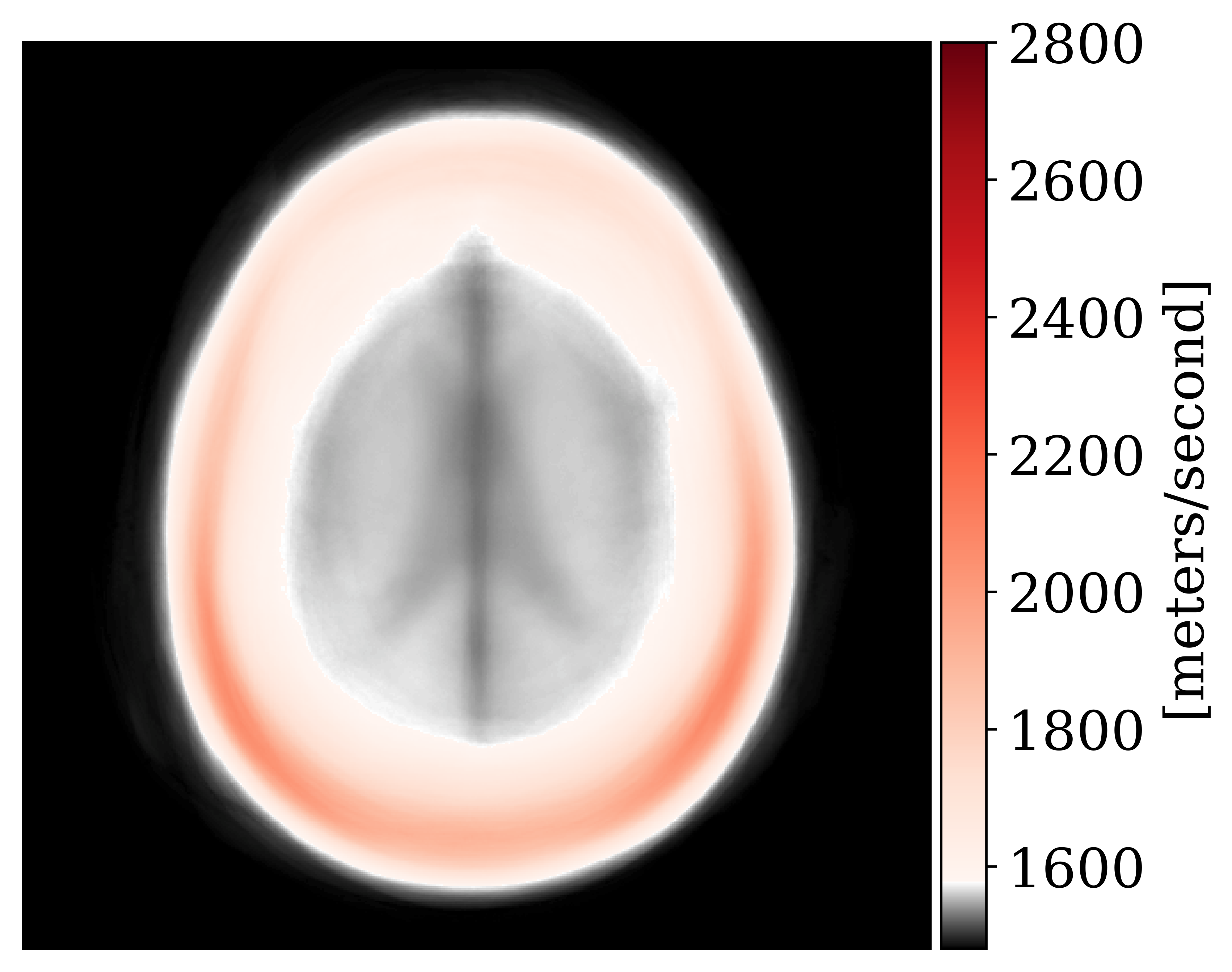}

}

\subcaption{\label{fig-var-iter1}\(\mathbb{E} \, \, p(\mathbf{x})\)}

\end{minipage}%

\caption{\label{fig-variance-training}Training dataset used to train
models. (a) Standard deviation of samples. (b) Mean of samples.}

\end{figure}

Based off of the MRI dataset \citep{zbontar2018fastmri}, we manually
assigned acoustic values to MRI intensities by following the table of
acoustic brain tissue properties in the supplemental section of
\citep{guasch2020full}. Although MRI intensities are not necessarily
related to acoustic tissue properties, we found that we could produce
reasonably realistic acoustic parameters as compared to the acoustic
parameters from the MIDA volume. In Figure~\ref{fig-train-dataset}, we
show some example training acoustic parameters. We also plot the average
and standard variation between all 1000 training samples in
Figure~\ref{fig-variance-training}. From these plots, we note that there
are few similarities between training examples apart from the
biologically consistent human brain structures.

\section{Wave modeling and FWI implementation}\label{sec-fwi}

To mask source-receiver artifacts, the gradients used for the
traditional FWI optimizations are masked by a binary matrix where the
mask was made large enough to include the skull but otherwise assumed no
knowledge of the skull. We also avoid the inverse crime by generating
any ``observed'' acoustic data with a spatial finite-difference kernel
of size \(16\) gridpoints and with computational time discretization of
\(0.025\) microseconds while the physical operator used for gradient
calculation corresponds to a simulation with spatial finite-difference
kernel of size \(8\) gridpoints and computational time discretization of
\(0.5\) microseconds.

\section{Conditional normalizing flow training}\label{sec-training}

We implement a conditional normalizing flow with an architecture that
consists of \(3\) multiscale levels as in RealNVP
\citep{dinh2016density} and each level contains \(9\) conditional
coupling layers similar to C-INN layers \citep{ardizzone2019conditional}
and also GLOW learned channel permutations \citep{kingma2014adam}. The
learned network in the coupling layers is a \(3\) layer convolutional
resnet where each convolutional block has \(64\) hidden channels. As a
summary network, we used a \(4\) level UNet. In total, the network is
made out of \(6,913,326\) parameters. The optimization is stochastic
gradient descent with batch size of \(8\) and an ADAM
\citep{kingma2014adam} optimizer with learning rate of \(8e-4\).
Following the work of \citep{ardizzone2019conditional, tran2023one}, we
added Gaussian noise to the target images with standard deviation of
\(0.01\). We trained the network until the objective did not improve on
a leave-out validation set. The network architecture implementation and
training code can found in the github
\href{https://github.com/slimgroup/ASPIRE.jl}{ASPIRE.jl}.

\section{More posterior samples}\label{sec-more-post-samples}

\begin{figure}[H]

\begin{minipage}{\linewidth}
\begin{center}
\includegraphics[width=0.8\textwidth,height=\textheight]{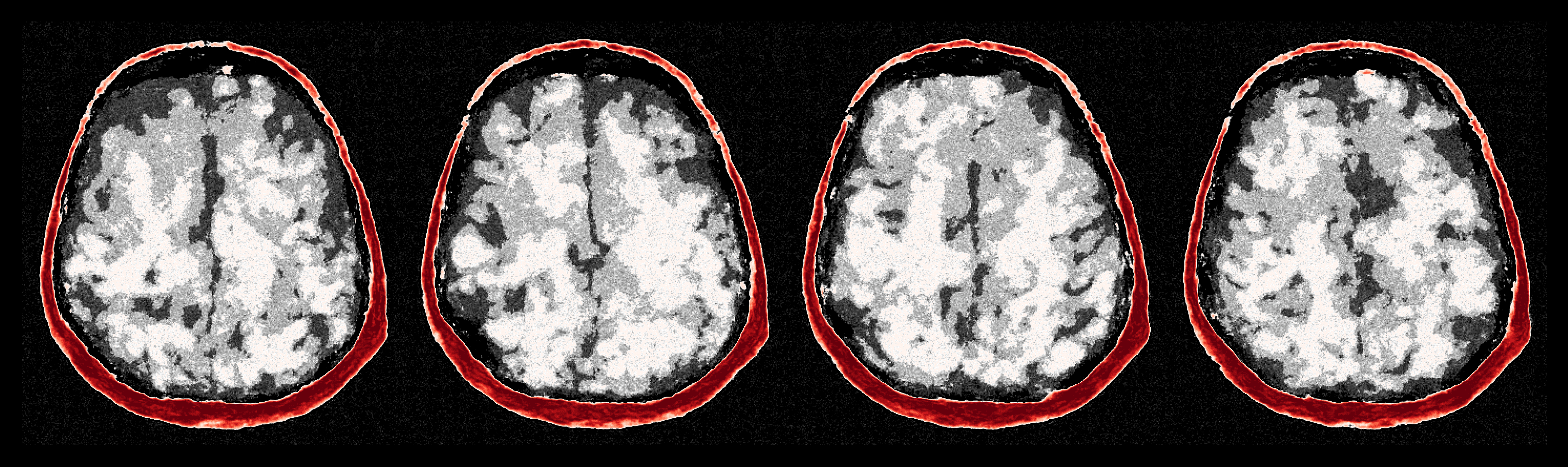}
\end{center}
\end{minipage}%
\newline
\begin{minipage}{\linewidth}
\begin{center}
\includegraphics[width=0.8\textwidth,height=\textheight]{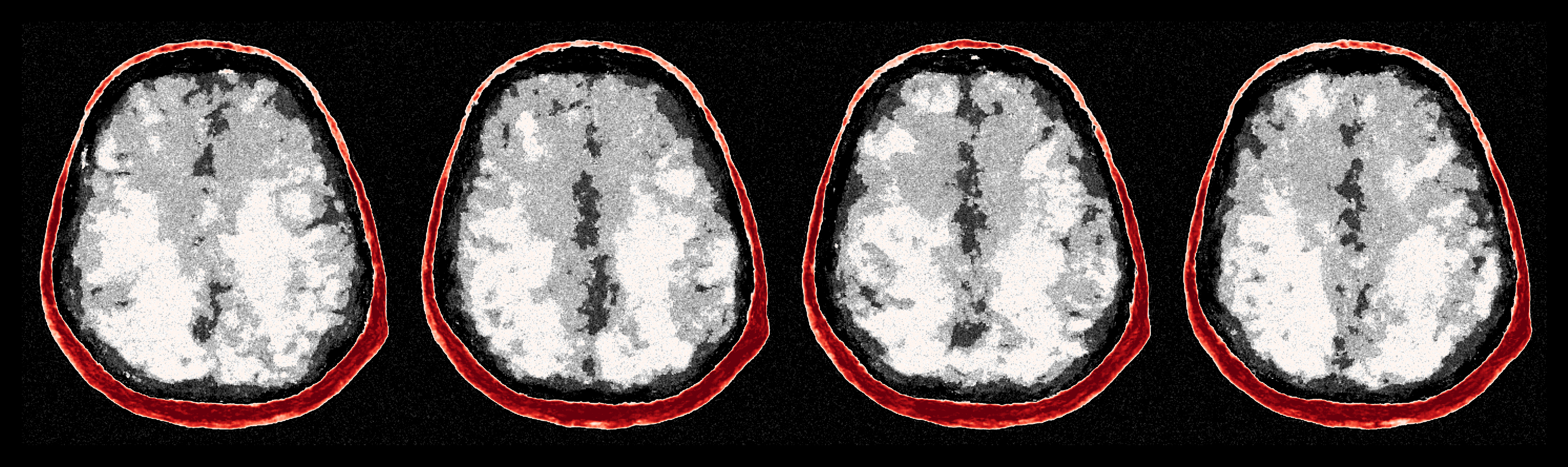}
\end{center}
\end{minipage}%
\newline
\begin{minipage}{\linewidth}
\begin{center}
\includegraphics[width=0.8\textwidth,height=\textheight]{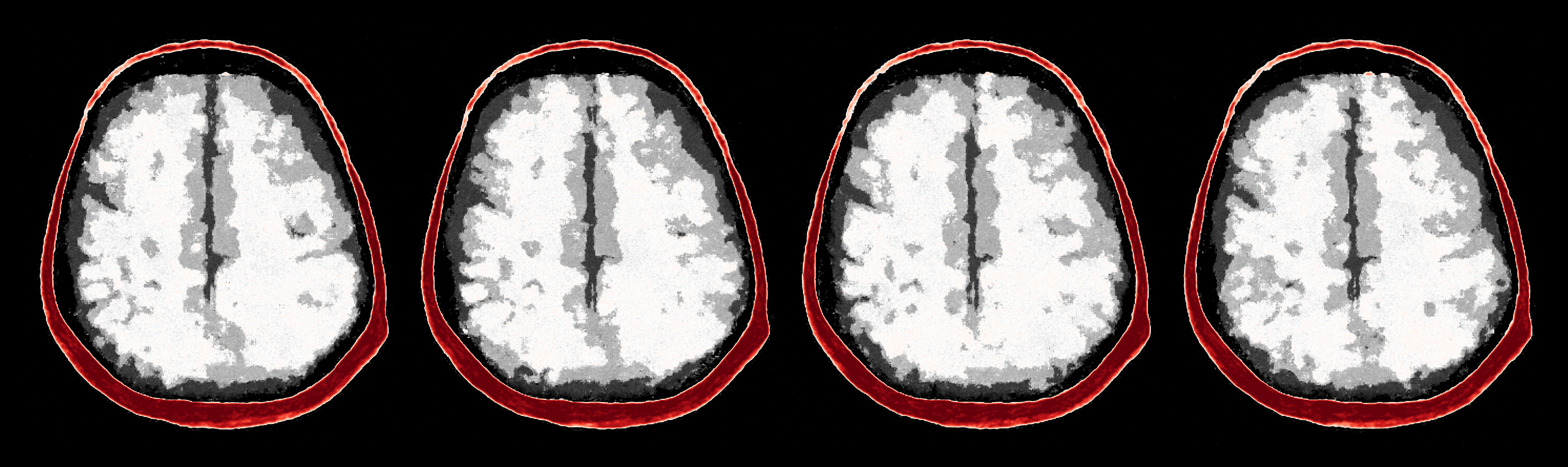}
\end{center}
\end{minipage}%
\newline
\begin{minipage}{\linewidth}
\begin{center}
\includegraphics[width=0.8\textwidth,height=\textheight]{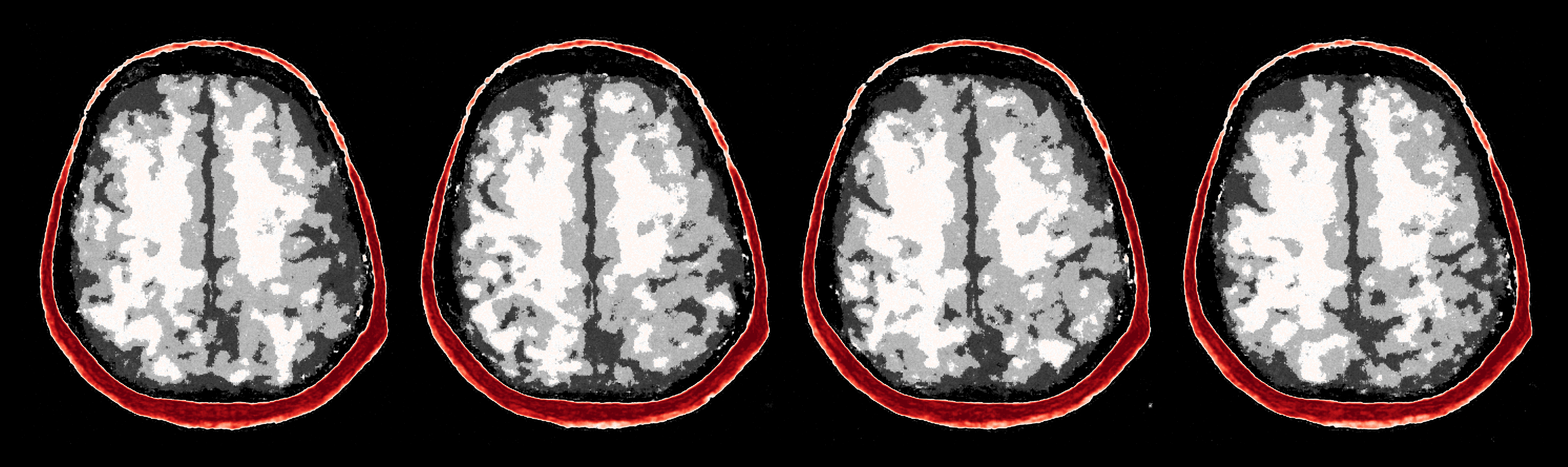}
\end{center}
\end{minipage}%

\caption{\label{fig-more-post}Posterior sampling for four iterations of
ASPIRE. First row is ASPIRE 1, second row is ASPIRE 2 and so on until
the last row showing results from ASPIRE 4. Each refinement visually
improves the quality and agreement of the posterior samples.}

\end{figure}%

\section{Selecting number of posterior
samples}\label{sec-num-post-samples}

From posterior samples, Monte Carlo estimates of the posterior
statistics can be calculated. As short hand, statistic estimates from
the distribution \(p\) are:

\begin{equation}\phantomsection\label{eq-pm}{
\mathbb{E} \, p := \mathbb{E}_{\mathbf{x}\sim p(\mathbf{x}|\mathbf{y})} \left[ \mathbf{x}\right]
}\end{equation}

\begin{equation}\phantomsection\label{eq-stdev}{ 
\sqrt{\mathbb{V}} \, p := \sqrt{  \mathbb{E}_{\mathbf{x} \sim p(\mathbf{x}|\mathbf{y})} \Bigl[\bigl(\mathbf{x} - \mathbb{E} \, p  \bigr )^2\Bigr]} .
}\end{equation}

For example, the posterior mean calculated from samples from the above
trained approximate posterior \(p_{\widehat{\boldsymbol{\theta}}}\) is
referred to as \(\mathbb{E} \, p_{\widehat{\boldsymbol{\theta}}}\).

\begin{figure}

\begin{minipage}{0.33\linewidth}

\centering{

\includegraphics{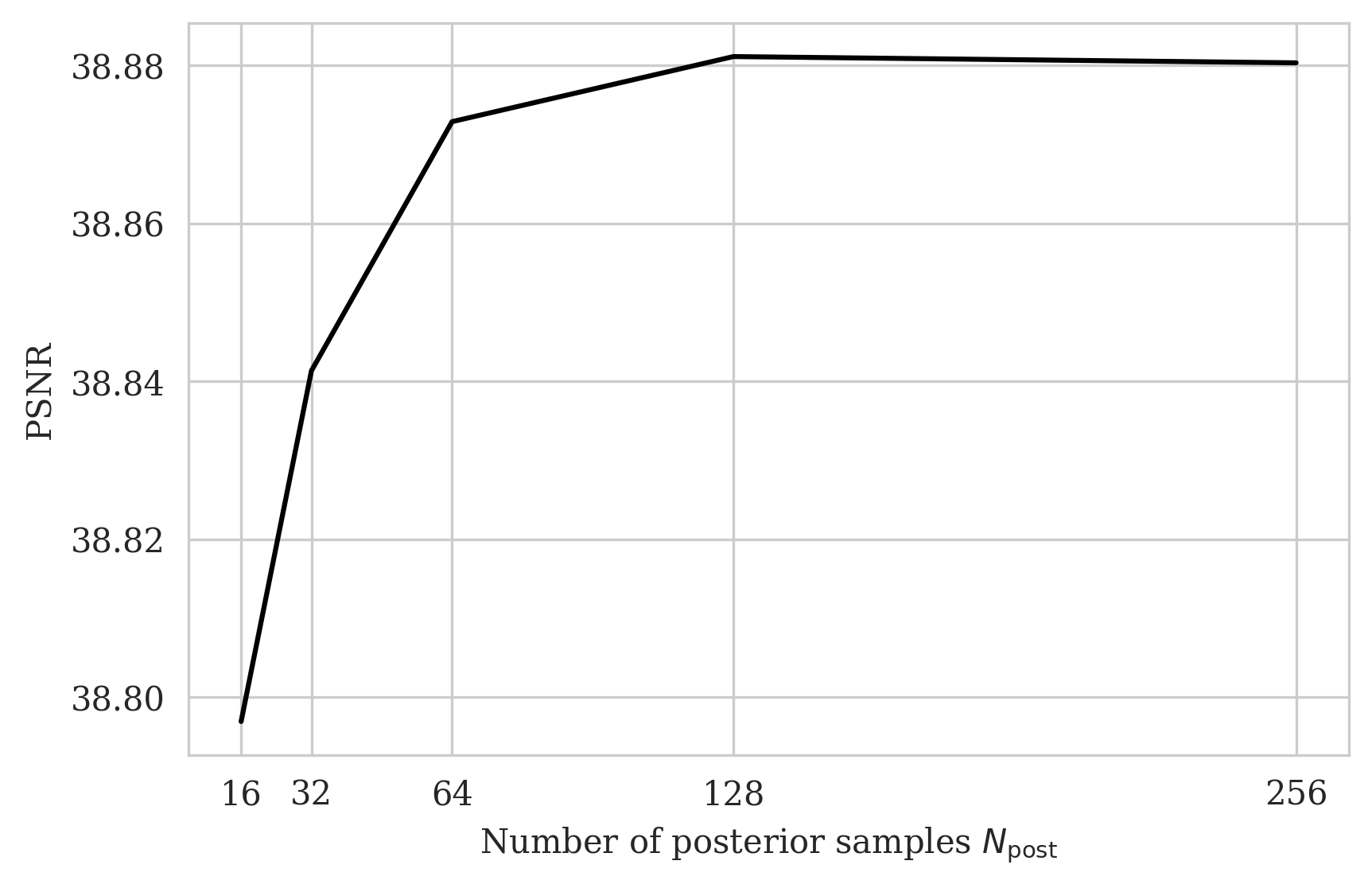}

}

\subcaption{\label{fig-log-}PSNR}

\end{minipage}%
\begin{minipage}{0.33\linewidth}

\centering{

\includegraphics{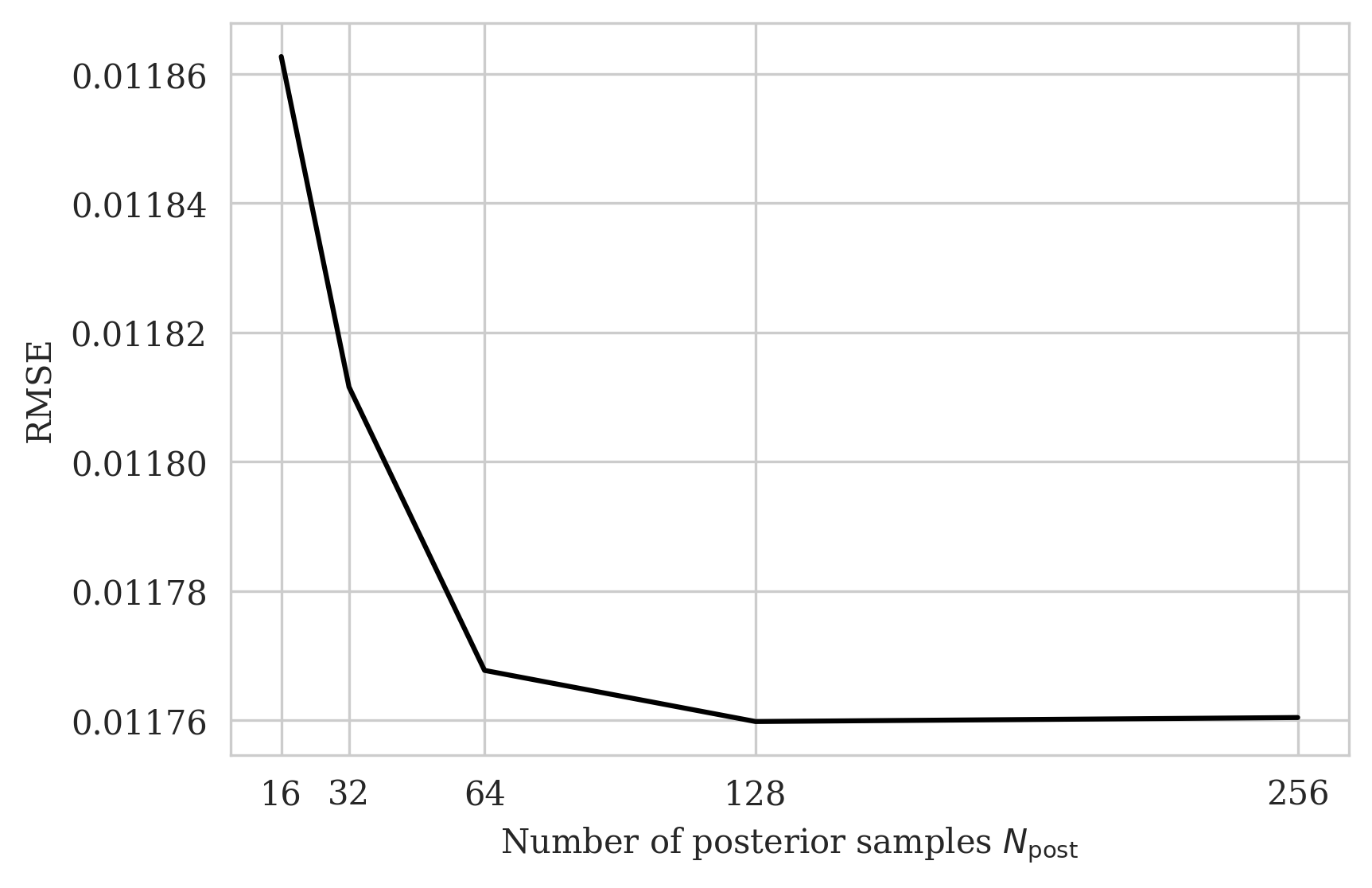}

}

\subcaption{\label{fig-log-}RMSE}

\end{minipage}%
\begin{minipage}{0.33\linewidth}

\centering{

\includegraphics{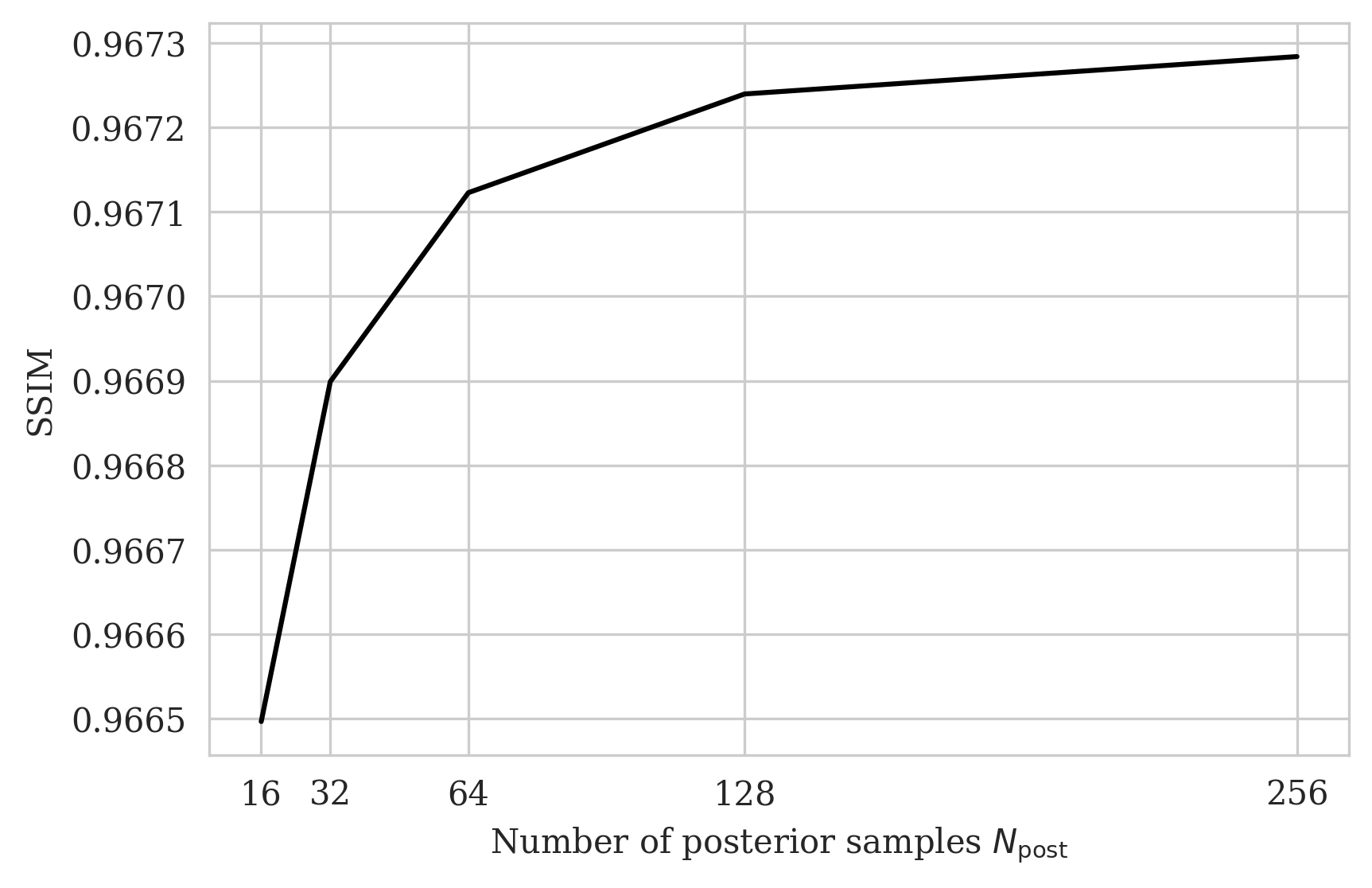}

}

\subcaption{\label{fig-log-}SSIM}

\end{minipage}%

\caption{\label{fig-convergence-posterior-samples}Quality metrics of the
posterior mean converge between \(256\) and \(512\) posterior samples.}

\end{figure}

We calculated empirical means and standard deviations using \(512\)
samples, but due to the efficiency of our method (10ms per sample), a
larger number of samples could be rapidly generated. We chose the
quantity \(512\) due to the convergence of quality metrics over number
of posterior samples as shown in
Figure~\ref{fig-convergence-posterior-samples}.

\renewcommand\refname{References}
  \bibliography{paper.bib}

\end{document}